\documentclass{article}

\usepackage[numbers]{natbib}
\usepackage{amsmath}

\usepackage{graphicx}
\usepackage{subfigure}
\usepackage{hyperref}
\usepackage{lineno}

\usepackage{amsfonts, amsthm, amssymb}
\newtheorem{theorem}{Theorem}[section]
\newtheorem{lemma}[theorem]{Lemma}

\newcommand{\Rho}{\mathrm{P}}

\begin{document}

\title{Training on the Edge of Stability Is Caused by Layerwise Jacobian Alignment}
\author{Mark Lowell \\
	National Geospatial-Intelligence Agency \\
	\texttt{Mark.C.Lowell@nga.mil} \\
	\and
	Catharine Kastner \\
	National Geospatial-Intelligence Agency
}

\maketitle

\begin{abstract}

During neural network training, the sharpness of the Hessian matrix of the training loss rises until training is on the edge of stability. As a result, even nonstochastic gradient descent does not accurately model the underlying dynamical system defined by the gradient flow of the training loss. We use an exponential Euler solver to train the network without entering the edge of stability, so that we accurately approximate the true gradient descent dynamics. We demonstrate experimentally that the increase in the sharpness of the Hessian matrix is caused by the layerwise Jacobian matrices of the network becoming aligned, so that a small change in the network preactivations near the inputs of the network can cause a large change in the outputs of the network. We further demonstrate that the degree of alignment scales with the size of the dataset by a power law with a coefficient of determination between $0.74$ and $0.98$.

\end{abstract}

Neural networks have become the \emph{de facto} standard in computer vision, natural language processing, and reinforcement learning. However, despite their empirical success, we still do not understand why neural networks work. Experiments have shown that ``bad'' neural network parameters do exist which perform very well on their training data but very poorly on test data, but networks do not arrive at those parameters when trained in practice \citep{rethinking}. Therefore, if we are going to understand why neural networks work, we need to understand their training process.

Neural networks are almost universally trained using algorithms based on gradient descent. The network parameters are obtained by solving a system of ordinary differential equations, where the flow is the gradient of a function measuring the network's error over the training data, commonly referred to as the training loss. Descending along this gradient reduces the training loss until the network reaches a minimum.

Since the flow does not have an explicit solution, in practice we discretize gradient descent into steps. However, \cite{edge-of-stability} showed that even in full-batch nonstochastic gradient descent, this discretization does \emph{not} accurately model the gradient flow. The maximum learning rate at which the training loss decreases monotonically is inversely proportional to the largest eigenvalue, or \textbf{sharpness}, of the Hessian matrix of the training loss. If the learning rate exceeds the maximum learning rate, the system is called unstable, and the training loss ordinarily diverges. \cite{edge-of-stability} showed experimentally that during training, the sharpness rises until the maximum learning rate approximately \emph{equals} the actual learning rate, \emph{for all reasonable choices of learning rate}. They showed that this occurs for a wide variety of network architectures, loss functions, and training datasets, and it appears to be a universal property of neural networks. Despite this, the training loss still decreases over time, although not monotonically. This phenomenon is called \textbf{training on the edge of stability} \citep{edge-of-stability}. % TODO per SI: Add sentence to the effect of, "Avoiding the edge of stability requires an infinitesmal learning rate"

In this work, we present a new mathematical analysis that separates the sharpness of the Hessian matrix into components. We then train a series of neural networks using a variant of the exponential Euler method \citep{exponential-integration-origin} so that training approximately follows the actual gradient flow, without entering the edge of stability. We use this method to show experimentally that training on the edge of stability is caused by the network's layerwise Jacobian matrices, which measure how a change in the pre-activations of one layer affects the pre-activations of the next layer. As the network is trained, the singular vectors of the layerwise Jacobians rotate so that a change along a narrow subspace at the network inputs can cause a large change in the network outputs. Further, the peak alignment increases with dataset size, following a power law with a coefficient of determination between $0.74$ and $0.98$.

Section~\ref{sec:related-work} discusses previous work on the eigenspectrum of the Hessian matrix, the edge of stability, and related phenomena. Section~\ref{sec:methodology} defines our terminology and presents our decomposition of the sharpness of the Hessian matrix. Section~\ref{sec:experiments} describes the experiments we conducted. Section~\ref{sec:results} states the results. Section~\ref{sec:discussion} discusses the results and analyzes their significance. Section~\ref{sec:conclusion} concludes.

\section{Related Work}
\label{sec:related-work}

The Hessian matrix of the training loss has been studied for some time owing to its role in neural network optimization. Early work showed that the eigenvectors of the Hessian matrix separate into a small top subspace with large eigenvalues and a larger bulk subspace with small eigenvalues \citep{bottou,eigenvalues,eigenvalues-2,eigenvalues-scale,eigenvalues-scale-2,eigenvalue-spectra}. At the same time, the gradient of the training loss lies primarily in the top subspace \citep{subspace}. It has recently been discovered that this top subspace is so highly curved that training occurs on the edge of stability \citep{edge-of-stability}. Training using sharpness-aware minimization \citep{edge-of-stability-sam} and adaptive gradient methods \citep{edge-of-stability-adam} has also been shown to occur on the edge of stability, so the phenomenon is not limited to ordinary gradient descent. % TODO per SI: why is it important that the edge of stability occurs for non-gradient-descent methods?

The discovery of training on the edge of stability has set off a substantial body of follow-up work \citep{eos-misc-1,eos-misc-3,eos-misc-4,eos-misc-5,eos-misc-6}. In particular, there has been considerable effort to explain why the training loss still decreases in expectation \citep{eos-misc-2,eos-misc-7,eos-misc-8}. Recent work has shown that when training on the edge of stability, the component of the gradient lying in the top subspace tends to cancel itself out, so that training is approximately equivalent to gradient descent using the projection of the gradient onto the bulk subspace \citep{self-stabilization}. There is even some evidence that training on the edge of stability may speed up convergence \citep{eos-misc-10} or improve generalization \citep{layerwise-sharpness}. However, there has been little progress in elucidating the cause of the edge of stability phenomenon. \cite{eos-misc-2} speculated that the increase in sharpness may be caused by the interaction of weight decay with the positive homogeneity of the network as a function, but training on the edge of stability occurs with networks that are trained without weight decay. \cite{eos-misc-8} observed that training on the edge of stability is connected to the subquadratic scaling of the loss function, but we do not yet know what causes this subquadratic scaling. \cite{edge-of-stability} observed that the increase in sharpness is primarily derived from the $G$ component of the Hessian matrix. \cite{layerwise-sharpness} further observed that if $G$ is separated into components corresponding to the layers of the network, the sharpness is highest for layers at the beginning of the network and lowest for layers at the end of the network.

\section{Methodology}
\label{sec:methodology}

\subsection{Terminology and Setup}
\label{sec:terminology-and-setup}

Let $f(x, \theta):X\times\Theta\to Z$ be a neural network with input $x$ and parameters $\theta$. The prototypical neural network is a multi-layer perceptron (MLP), where $\theta = (A^1|b^1|A^2|b^2|...)$ and:
$$
f(x, \theta) = A^{L} x^{L} + b^{L}
$$$$
\hat{x}^i = A^{i-1}x^{i-1} + b^{i-1}
$$$$
x^i = \sigma(\hat{x}^i)
$$
where $L$ is the number of layers in the network, $i$ indexes the layer, $A^i \in \mathbb{R}^{n_i\times n_{i+1}}, b^i \in \mathbb{R}^{n_{i+1}}$, $x^0=x$ are the network inputs, $n_0$ is the dimension of the inputs, $n_i$ is the width of layer $i$, $n_{L}$ is the dimension of the network outputs, and $\sigma$ is an elementwise activation function such as the rectified linear unit. We define $x^i$ to be the \textbf{activation} of layer $i$ and $\hat{x}^i$ to be the \textbf{pre-activation}.

Let $p$ be a probability distribution generating pairs $(x, y) \in X\times Y$, where $y$ is a label of the input $x$. The goal of supervised learning is to find parameters $\theta$ such that $f(x, \theta)$ approximates $y$ for all $(x, y) \sim p$. Note that it is possible $Z \neq Y$; for example, $y$ may be an integer category while $z$ is a vector of probabilities.

We define the \textbf{criterion} $l(z, y)$ to be a convex function measuring the difference between a network output $z\in Z$ and the label $y\in Y$. Commonly used criteria include the cross-entropy loss and the mean squared error (MSE). We define the \textbf{training loss} $\widetilde{\mathcal{L}}(\theta)$ to be the expectation of the criterion applied to the network outputs over a training dataset $\mathcal{T}$ sampled from $p$:
$$
\widetilde{\mathcal{L}}(\theta) = \underset{(x,y)\sim\mathcal{T}}{\mathbb{E}} l(f(x, \theta), y)
$$

% Note that we distinguish between the criterion, which is a function of the network outputs $f(x, \theta)$, and the loss function, which is a function of the network parameters $\theta$. We require that the criterion be convex, while the loss function famously is not.

During training, $\theta_0$ is initialized via some random distribution and is then optimized via gradient descent on the training loss:
$$
\theta_{u + 1} = \theta_u - \eta \nabla_\theta\widetilde{\mathcal{L}}(\theta_u)
$$
where $u$ is the index of the step and $\eta$ is a learning rate. This procedure is intended to approximate the dynamical system $\frac{d\theta}{dt}(t) = -\nabla_\theta\widetilde{\mathcal{L}}(\theta(t))$. The Taylor series expansion implies that:
$$
	\widetilde{\mathcal{L}}(\theta + \Delta\theta) = 
	\widetilde{\mathcal{L}}(\theta) + \Delta\theta^T\nabla_\theta\widetilde{\mathcal{L}}(\theta)
	+ \frac{1}{2}\Delta\theta^T(\mathcal{H}_\theta\widetilde{\mathcal{L}}(\theta))\Delta\theta + \mathcal{O}(||\Delta\theta||^3)
$$\begin{equation}\label{eqn:1}\begin{split}
	\widetilde{\mathcal{L}}(\theta_{u+1}) = &
	\widetilde{\mathcal{L}}(\theta_u) - \eta||\nabla_\theta\widetilde{\mathcal{L}}(\theta_u)||^2 \\
	& + \frac{1}{2}\eta^2 \nabla_\theta\widetilde{\mathcal{L}}(\theta_u)^T(\mathcal{H}_\theta\widetilde{\mathcal{L}}(\theta_u))\nabla_\theta\widetilde{\mathcal{L}}(\theta_u)
	+ \mathcal{O}(\eta^3)
\end{split}\end{equation}
where $\mathcal{H}_\theta\widetilde{\mathcal{L}}(\theta)$ denotes the Hessian matrix of the training loss:
$$
	(\mathcal{H}_\theta\widetilde{\mathcal{L}}(\theta))_{i,j} = \frac{\partial^2\widetilde{\mathcal{L}}(\theta)}{\partial \theta_i \partial \theta_j}
$$
Since $\mathcal{H}_\theta\widetilde{\mathcal{L}}(\theta)$ is symmetric by construction, by the finite-dimensional spectral theorem its eigenvalues are all real and it has an orthonormal real eigenbasis. Let $\lambda_m(M)$ be the $m$th eigenvalues of a symmetric matrix $M$, ordered from greatest to least, and let $v_m(M)$ be the corresponding eigenvector. If we define $\nabla_\theta\widetilde{\mathcal{L}}(\theta) = \sum c_m(\theta)v_m(\mathcal{H}_\theta\widetilde{\mathcal{L}}(\theta))$ for some coordinates $c_m(\theta) = \nabla_\theta\widetilde{\mathcal{L}}(\theta) \cdot v_m(\mathcal{H}_\theta\widetilde{\mathcal{L}}(\theta)) \in\mathbb{R}$, then Equation~\ref{eqn:1} becomes:

$$
	\widetilde{\mathcal{L}}(\theta_{u+1}) = 
	\widetilde{\mathcal{L}}(\theta_u) + \sum_m \eta c_m^2(\theta_u)\left(-1 + \frac{1}{2}\eta\lambda_m(\mathcal{H}_\theta\widetilde{\mathcal{L}}(\theta_u))\right)
	+ \mathcal{O}(\eta^3)
$$
We generally assume that $\eta$ is small enough that we can neglect the error term $\mathcal{O}(\eta^3)$. Under this assumption, the training loss will decrease monotonically if $\eta < 2 / \lambda_1(\mathcal{H}_\theta\widetilde{\mathcal{L}}(\theta_u))$. If $\eta > 2 / \lambda_1(\mathcal{H}_\theta\widetilde{\mathcal{L}}(\theta_u))$, the system is unstable and we expect the loss to diverge. If $\eta \approx 2 / \lambda_1(\mathcal{H}_\theta\widetilde{\mathcal{L}}(\theta_u))$, the system is on the edge of stability \citep{edge-of-stability}.

\subsection{Decomposition of the Hessian Matrix}
\label{sec:decomposition}

Identifying the cause of training on the edge of stability requires decomposing the sharpness of $\mathcal{H}_\theta\widetilde{\mathcal{L}}(\theta)$ into components. For notational convenience, we will suppress the inputs to our matrices unless they are needed for clarity.

First, $\mathcal{H}_\theta\widetilde{\mathcal{L}}$ can be decomposed into the sum of two matrices \citep{gauss-newton}:
\begin{equation}\label{eqn:g-plus-h}
	\mathcal{H}_\theta \widetilde{\mathcal{L}} = G + H
\end{equation}$$
	G_{j,k} = 
	\underset{x,y\sim\mathcal{T}}{\mathbb{E}} \left(
	\sum_{q,r} \frac{\partial^2 l}{\partial z_q\partial z_r}
	\frac{\partial f_q}{\partial \theta_j}
	\frac{\partial f_r}{\partial \theta_k}
	\right)
$$$$
	H_{j,k} = 
	\underset{x,y\sim\mathcal{T}}{\mathbb{E}} \left(
	\sum_{q} \frac{\partial l}{\partial z_q}
	\frac{\partial^2 f_q}{\partial \theta_j\partial \theta_k}
	\right)
$$
The $G$ matrix can be rewritten in matrix notation as:
$$
	G = 
	\underset{x,y\sim\mathcal{T}}{\mathbb{E}} \left(
	\left(\frac{\partial f}{\partial \theta}\right)^T
	(\mathcal{H}_zl) \left(\frac{\partial f}{\partial\theta}\right)	\right)
$$
where, by an abuse of notation, $\frac{\partial f}{\partial \theta}$ denotes the Jacobian matrix of the network outputs in terms of its parameters, and $\mathcal{H}_zl$ denotes the Hessian matrix of the criterion $l$ in terms of the network outputs $z$. \cite{edge-of-stability} found that the sharpness of $H$ is generally low, and the top eigenvalues of $\mathcal{H}_\theta\widetilde{\mathcal{L}}$ are dominated by the top eigenvalues of $G$.

Since we have assumed that the criterion $l$ is convex, $\mathcal{H}_zl$ will be positive semi-definite. It therefore has a unique positive semi-definite symmetric square root $R(z, y)$ \citep{matrix-analysis}, such that $RR = \mathcal{H}_zl$. We define $K$ as:
$$
	K = R \frac{\partial f}{\partial\theta}
$$
We can then rewrite $G$ as the second moment matrix of $K$:
$$
	G = \mathbb{E} \left(K^TK\right)
$$
Recall $\lambda_{1}(M)$ denotes the top eigenvalue of a matrix $M$. We define the \textbf{overlap ratio} $\rho(M)$ of a random matrix $M$ to be:
$$
\rho(M) = \frac{\lambda_{1}(\mathbb{E} (M^TM))}{\mathbb{E} (\lambda_{1}(M^TM))} = \frac{\lambda_{1}(\mathbb{E} (M^TM))}{\mathbb{E} ||M||_{\max}^2}
$$
where $||\cdot||_{\max}$ is the operator norm. We can then decompose the top eigenvalue of $G$ into two components:
\begin{equation}\label{eqn:rho}
\lambda_{1}(G) = \rho(K) \cdot \mathbb{E} ||K||_{\max}^2
\end{equation}
Further, if $\lambda, v$ is a nonzero eigenvalue, eigenvector of $M^TM$, then $\lambda, Mv / ||Mv||$ is a nonzero eigenvalue, eigenvector of $MM^T$ (see Lemma~\ref{lemma:mmt} in the Supplemental Material section). $K$ is an $O \times P$ matrix, where $O$ is the number of network outputs -- typically quite small -- and $P$ is the number of network parameters, which is typically quite large. The nonzero eigenvalues of $K^TK$, which is a $P \times P$ matrix, are the same as the nonzero eigenvalues of $KK^T$, which is an $O \times O$ matrix.

Recall that $A^i, b^i$ are the weight and bias of layer $i$, respectively. We define $\frac{\partial f}{\partial A^i}, \frac{\partial f}{\partial b^i}$ to be the Jacobian matrices of the network outputs in terms of these parameters. $\frac{\partial f}{\partial \theta}$ then has a block structure corresponding to the parameters in $\theta$:
$$
\frac{\partial f}{\partial \theta} = \left(
\left.\frac{\partial f}{\partial A^1} \right |
\left.\frac{\partial f}{\partial b^1} \right |
\left.\frac{\partial f}{\partial A^2} \right | ... \right)
$$
$KK^T$ can therefore be decomposed into a sum of matrices, each corresponding to one of the layers of the network:
\begin{equation}\label{eqn:kkt}
	KK^T = K^1(K^1)^T + K^2(K^2)^T + ...
\end{equation}$$
K^i = R\left(\left. \frac{\partial f}{\partial A^i} \right| \frac{\partial f}{\partial b^i}\right)
$$
\cite{layerwise-sharpness} observed that the component of $KK^T$ corresponding to $K^1(K^1)^T$ tends to have higher sharpness than later layers. The chain rule applied to the definition of the activations and pre-activations implies that:
%\begin{equation}\label{eqn:jac_a}
$$
\frac{\partial f_q}{\partial A^i_{j,k}} = 
\frac{\partial f_q}{\partial \hat{x}^i_j} 
\frac{\partial \hat{x}^i_j}{\partial A^i_{j,k}} =
\frac{\partial f_q}{\partial \hat{x}^i_j} x^{i-1}_k
$$$$
%\end{equation}\begin{equation}\label{eqn:jac_b}
\frac{\partial f_q}{\partial b^i_{j}} = 
\frac{\partial f_q}{\partial \hat{x}^i_j} 
\frac{\partial \hat{x}^i_j}{\partial b^i_{j}} =
\frac{\partial f_q}{\partial \hat{x}^i_j} 
$$$$
%\end{equation}\begin{equation}\label{eqn:jac_hat_x}
\frac{\partial f_q}{\partial \hat{x}^i_j} = 
\frac{\partial f_q}{\partial {x}^i_j} 
\frac{\partial x^i_j}{\partial \hat{x}^i_j} = 
\frac{\partial f_q}{\partial {x}^i_j}
\sigma'(\hat{x}^i_j)
$$$$
%\end{equation}\begin{equation}\label{eqn:jac_x}
\frac{\partial f_q}{\partial {x}^i_j} =
\sum_k
\frac{\partial f_q}{\partial \hat{x}^{i+1}_k}
\frac{\partial \hat{x}^{i+1}_k}{\partial x^i_j} =
\sum_k
\frac{\partial f_q}{\partial \hat{x}^{i+1}_k}
A^{i+1}_{k,j}
$$
%\end{equation}
If we apply these rules to Equation~\ref{eqn:kkt}, we obtain:
$$
\left(\frac{\partial f}{\partial A^i} \left(\frac{\partial f}{\partial A^i}\right)^T\right)_{q,r} =
\sum_{j,k} \frac{\partial f_q}{\partial A^i_{j,k}}\frac{\partial f_r}{\partial A^i_{j,k}} =
\sum_{j,k} (x^{i-1}_k)^2 \frac{\partial f_q}{\partial \hat{x}^i_{j}}\frac{\partial f_r}{\partial \hat{x}^i_{j}} =
||x^{i-1}||^2 \left(\frac{\partial f}{\partial \hat{x}^i} \left(\frac{\partial f}{\partial \hat{x}^i}\right)^T\right)_{q,r}
$$$$
\left(\frac{\partial f}{\partial b^i} \left(\frac{\partial f}{\partial b^i}\right)^T\right)_{q,r} =
\sum_{j} \frac{\partial f_q}{\partial b^i_{j}}\frac{\partial f_r}{\partial b^i_{j}} =
\sum_{j} \frac{\partial f_q}{\partial \hat{x}^i_{j}}\frac{\partial f_r}{\partial \hat{x}^i_{j}} =
\left(\frac{\partial f}{\partial \hat{x}^i} \left(\frac{\partial f}{\partial \hat{x}^i}\right)^T\right)_{q,r}
$$
where we define $\frac{\partial f}{\partial \hat{x}^i}$ to be the \textbf{pre-activation Jacobians}, or the Jacobian of the network outputs in terms of the pre-activations of layer $i$. Therefore, $K^i(K^i)^T = \Delta^i(\Delta^i)^T$, where:
$$
\Delta^i = 
\sqrt{1 + ||x^{i-1}||^2} R \left(\frac{\partial f}{\partial\hat{x}^i}\right)
$$
We define the \textbf{layerwise Jacobians} $\frac{\partial \hat{x}^{i+1}}{\partial \hat{x}^i}$ analogously to the pre-activation Jacobians:
$$
\left(\frac{\partial \hat{x}^{i+1}}{\partial \hat{x}^i}\right)_{jk} =
\frac{\partial \hat{x}^{i+1}_j}{\partial \hat{x}^i_k} =
 \frac{\partial \hat{x}^{i+1}_j}{\partial {x}^i_k} 
 \frac{\partial x^{i}_k}{\partial \hat{x}^i_k} = 
\sigma'(\hat{x}^i_k) A^{i+1}_{j,k}
$$
For most common activation functions, such as ReLU and tanh, $\sigma'(x) \approx 0$ for at least half of its domain. This implies that the layerwise Jacobians will generally be ill-conditioned.

We can then relate the pre-activation Jacobians and layerwise Jacobians by:
$$
\frac{\partial f}{\partial \hat{x}^i} 
= 
\frac{\partial f}{\partial \hat{x}^{i+1}}
\frac{\partial \hat{x}^{i+1}}{\partial \hat{x}^i}
$$
and define $\Delta^i$ iteratively as:
$$
\Delta^i = \chi^i \Delta^{i+1} 
\frac{\partial \hat{x}^{i+1}}{\partial \hat{x}^i}
$$\begin{equation}\label{eqn:delta-last}
\Delta^{L} = \sqrt{1 + ||x^{L-1}||^2} R
\end{equation}
where $L$ is the number of layers, and:
$$
\chi^i = \sqrt{\frac{1 + ||x^{i-1}||^2}{1 + ||x^i||^2}}
$$
For random matrices $M_1, M_2$, we define $r(M_1, M_2)$ to be the \textbf{alignment ratio}:
$$
r(M_1, M_2) = \frac{ \mathbb{E} ||M_1M_2||^2_{\max}}{\mathbb{E} ||M_1||^2_{\max} \cdot \mathbb{E} ||M_2||^2_{\max} }
$$
This allows us to break $\mathbb{E} ||\Delta^i||^2_{\max}$ into the product:
\begin{equation}\label{eqn:delta-mid}
\mathbb{E} ||\Delta^i||^2_{\max} =
	\mathbb{E} (\chi^i)^2 \cdot
\mathbb{E} ||\Delta^{i+1}||^2_{\max} \cdot
\mathbb{E} \left|\left|\frac{\partial \hat{x}^{i+1}}{\partial \hat{x}^i}\right|\right|_{\max}^2 \cdot
r\left(\Delta^{i+1}, \frac{\partial \hat{x}^{i+1}}{\partial \hat{x}^i}\right) \cdot
	r\left(\chi^i I, \frac{\Delta^i}{\chi^i} \right)
\end{equation}
Combining Equations~\ref{eqn:delta-last} and \ref{eqn:delta-mid}, we obtain, for any layer $k$:
\begin{equation}\label{eqn:main}
	\mathbb{E} ||\Delta^k||^2_{\max} = \Pi_\chi^k \cdot \Rho_{\chi,\Delta/\chi}^k \cdot \Pi_J^k \cdot \Rho_{\Delta, J}^k \cdot \mathbb{E} ||\Delta^L||^2_{\max}
\end{equation}
\begin{center}\begin{tabular}{cc}
$\Pi_\chi^k = \prod_{i=k}^L \mathbb{E} (\chi^i)^2$&$\Rho_{\chi,\Delta/\chi}^k = \prod_{i=k}^{L-1} r\left(\chi^i, \frac{\Delta^i}{\chi^i}\right)$\\
$\Pi_J^k = \prod_{i=k}^{L-1} \mathbb{E} \left|\left|\frac{\partial \hat{x}^{i+1}}{\partial \hat{x}^i}\right|\right|^2_{\max}$&$\Rho_{\Delta,J}^k = \prod_{i=k}^{L-1} r\left(\Delta^{i+1}, \frac{\partial \hat{x}^{i+1}}{\partial\hat{x}^i}\right)$
\end{tabular}\end{center}
Note that $\Pi_\chi^k, \Rho_{\chi,\Delta/\chi}^k, \Pi_J^k, \Rho_{\Delta,J}$ are all functions of time $t$, but we have suppressed the variable for notational convenience.

\section{Experiments}
\label{sec:experiments}

We trained a large number of multi-layer perceptrons (MLPs) using an exponential Euler solver \citep{exponential-integration-origin,exponential-integration-survey} to ensure training did not enter the edge of stability and instead accurately approximated the underlying gradient flow; see Section~\ref{sec:exponential-integration} in the Supplemental Material section for full details. We trained our networks on a variety of datasets and criteria. The full range of combinations used is shown in Table~\ref{tab:experiment-configurations}.

For datasets, we used CIFAR-10 \citep{cifar-10}, the UCR AtrialFibrillation time series classification dataset \citep{ucr}, the Stanford Sentiment Treebank (SST) sentence classification dataset \citep{sst}, and a synthetic image classification dataset inspired by dice (see Section~\ref{sec:dice} in the Supplemental Material section for details). We split the time series in AtrialFibrillation into sequences of length 128. We zero-padded or cropped all phrases in SST to a fixed length of 32 words and generated word embeddings using the RoBERTA base encoder \citep{roberta}. In all cases, we flattened the input data into vectors prior to inputting it into the network. We used multiple randomly selected subsets of each of the training datasets, varying the size by factors of 2.

For criteria, we used either cross-entropy or the mean squared error (MSE) between the network outputs and a one-hot encoded label vector.

For each combination of dataset, dataset size, and criterion, we trained sixteen networks using four different random seeds to initialize the network and four different random seeds to select the subset of the training data. For example, when training on CIFAR-10 using the cross-entropy criterion, we trained $5\times4\times4=80$ different networks.

In all experiments we used MLPs of depth 6 and width 512 as our network. We used the exponential linear unit (ELU) \citep{elu} activation function instead of ReLU to avoid numerical difficulties due to nonsmoothness. We initialized our networks using the Xavier initialization \citep{xavier}, but used the correction from the Kaiming initialization to scale by the gain \citep{kaiming}. We followed \cite{elu} in using $\sqrt{2}$ as the gain for the ELU function.

\begin{table*}\begin{center}
	\begin{tabular}{llll}
		Dataset&Subset Sizes&Criterion&Number of Experiments\\
		\hline
	                          Dice& 468, 937, 1875, 3750, 7500&Cross-Entropy& $4\times4\times5=80$ \\ % EoS-67
		              CIFAR-10& 390, 781, 1562, 3125, 6250&Cross-Entropy& $4\times4\times5=80$ \\ % EoS-68
		              CIFAR-10&        390, 781, 1562, 3125&          MSE& $4\times4\times4=64$ \\ % EoS-69
		    AtrialFibrillation&        60, 120, 240, 480, 960&Cross-Entropy& $4\times4\times5=80$ \\ % EoS-74
		                   SST&        461, 922, 1844, 3689&Cross-Entropy& $4\times4\times4=64$ \\ % EoS-75
%		AlexNet&              Dice&30,000 -- 468&Cross-Entropy&112 \\ % EoS-56: Ran 1 & 2.
	\end{tabular}
	\caption{\label{tab:experiment-configurations}\textbf{Configurations of Experiments}}\end{center}
\end{table*}

We trained each network using the exponential Euler method until the training loss reached 0.01 (for the cross-entropy criterion) or 0.02 (for MSE). We performed all experiments in double precision to eliminate potential issues with numerical overflow. At every 100 iterations we saved a snapshot of the network weights. We then went back and, for each snapshot, calculated the top eigenvalues of $G$ and $H$, $\rho(K)$, $\mathbb{E} ||K||^2_{\max}$, and $\mathbb{E} ||K^i||^2_{\max}$, $\Pi_\chi^i, \Rho_{\chi, \Delta / \chi}^i, \Pi_J^i$, and $\Rho_{\Delta, J}^i$ for all layers $i$. See Section~\ref{sec:power-iteration} in the Supplemental Material section for implementation details.

\section{Results}
\label{sec:results}

\begin{figure*}
	\begin{center}
		\includegraphics[width=\textwidth]{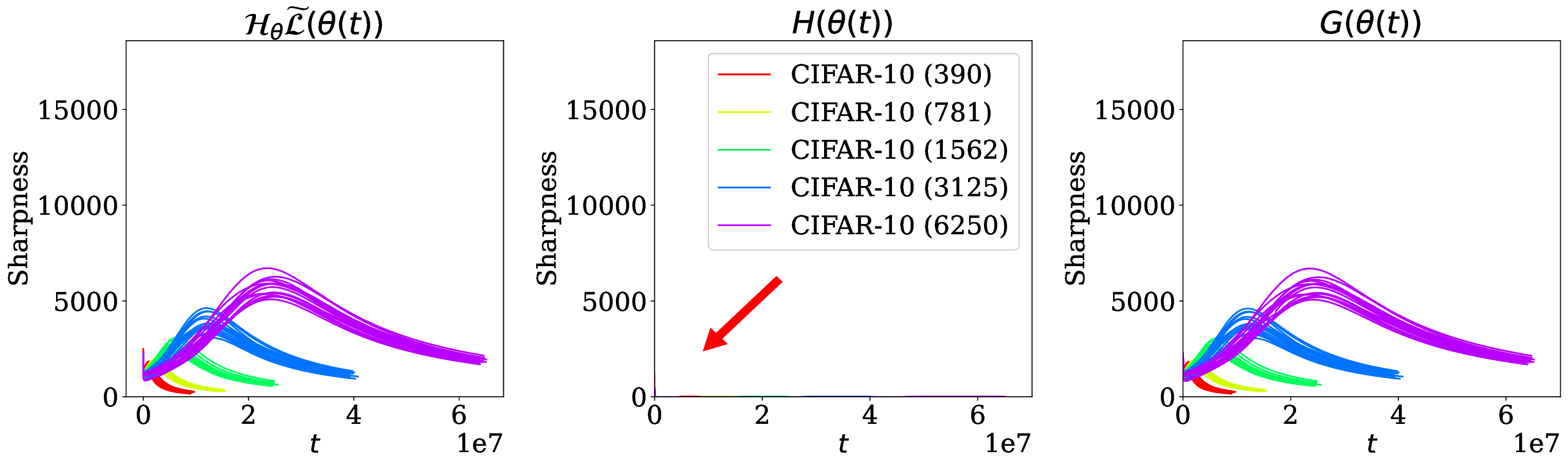}
		\caption{\label{fig:mlp-cifar10-cross-entropy-components-of-the-hessian}Sharpness of $\mathcal{H}_\theta\widetilde{\mathcal{L}}$ (left), $H$ (middle), and $G$ (right) when trained on CIFAR-10 with cross-entropy}
	\end{center}
\end{figure*}

\begin{figure*}
	\begin{center}
		\includegraphics[width=\textwidth]{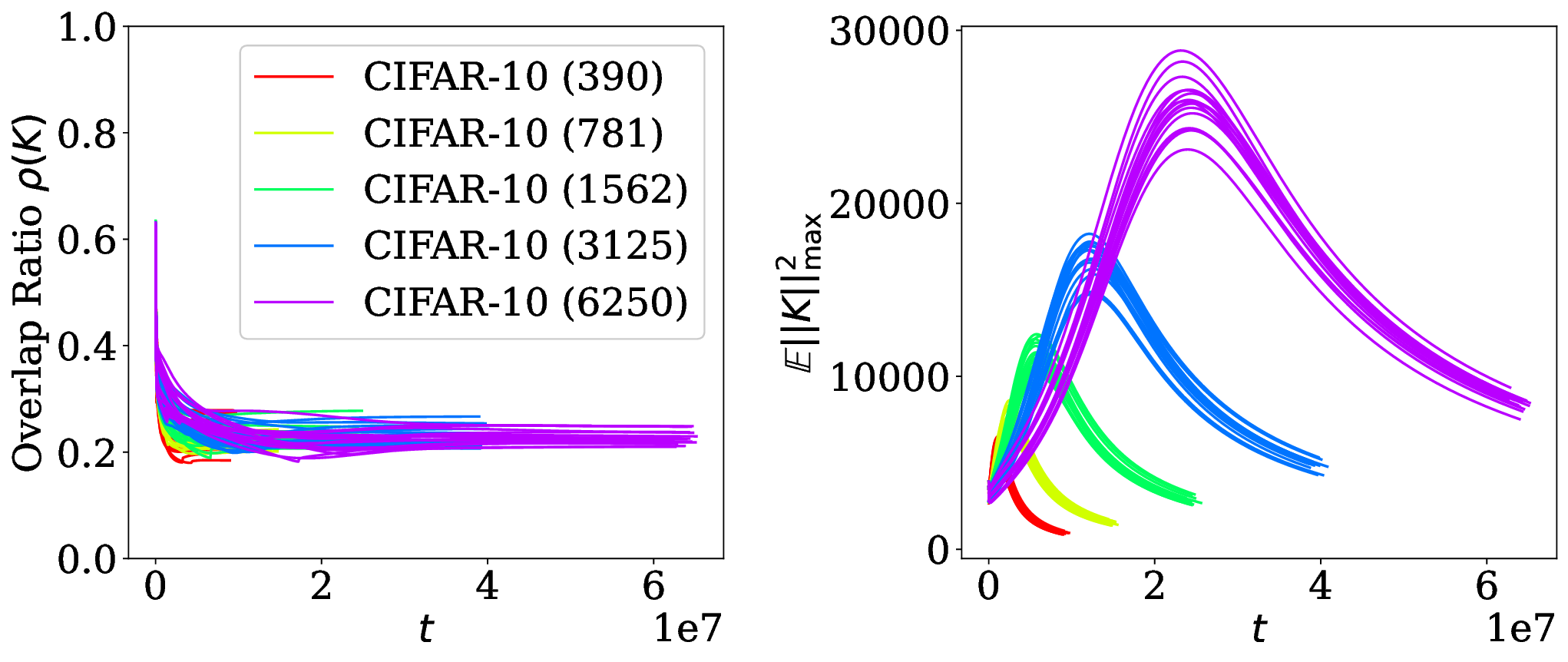}
		\caption{\label{fig:mlp-cifar10-cross-entropy-overlap-ratio}$\rho(K)$ (left) and $\mathbb{E}||K||^2_{\max}$ (right) when trained on CIFAR-10 with cross-entropy}
	\end{center}
\end{figure*}

\begin{figure}
	\begin{center}
		\includegraphics[width=\textwidth]{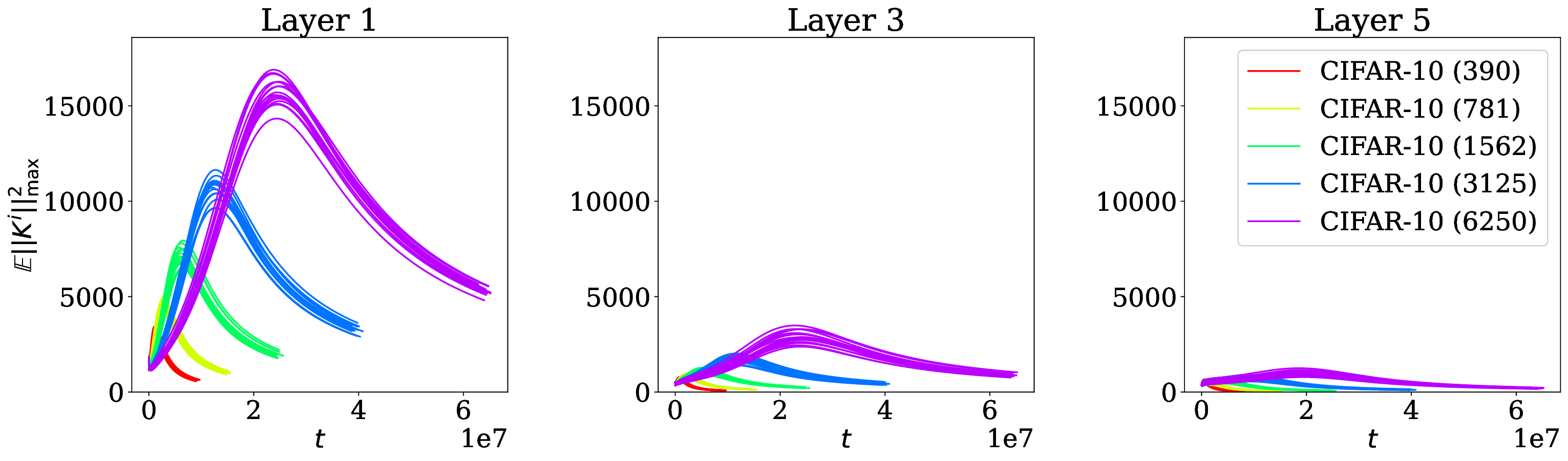}
		\caption{\label{fig:mlp-cifar10-cross-entropy-layerwise-sharpness-cutdown}$\mathbb{E} ||K_i||^2_{\max}$ for $i = 1$ (left), 3 (middle), and 5 (right) when trained on CIFAR-10 with cross-entropy}
	\end{center}
\end{figure}

\begin{figure}
	\begin{center}
		\includegraphics[width=\textwidth]{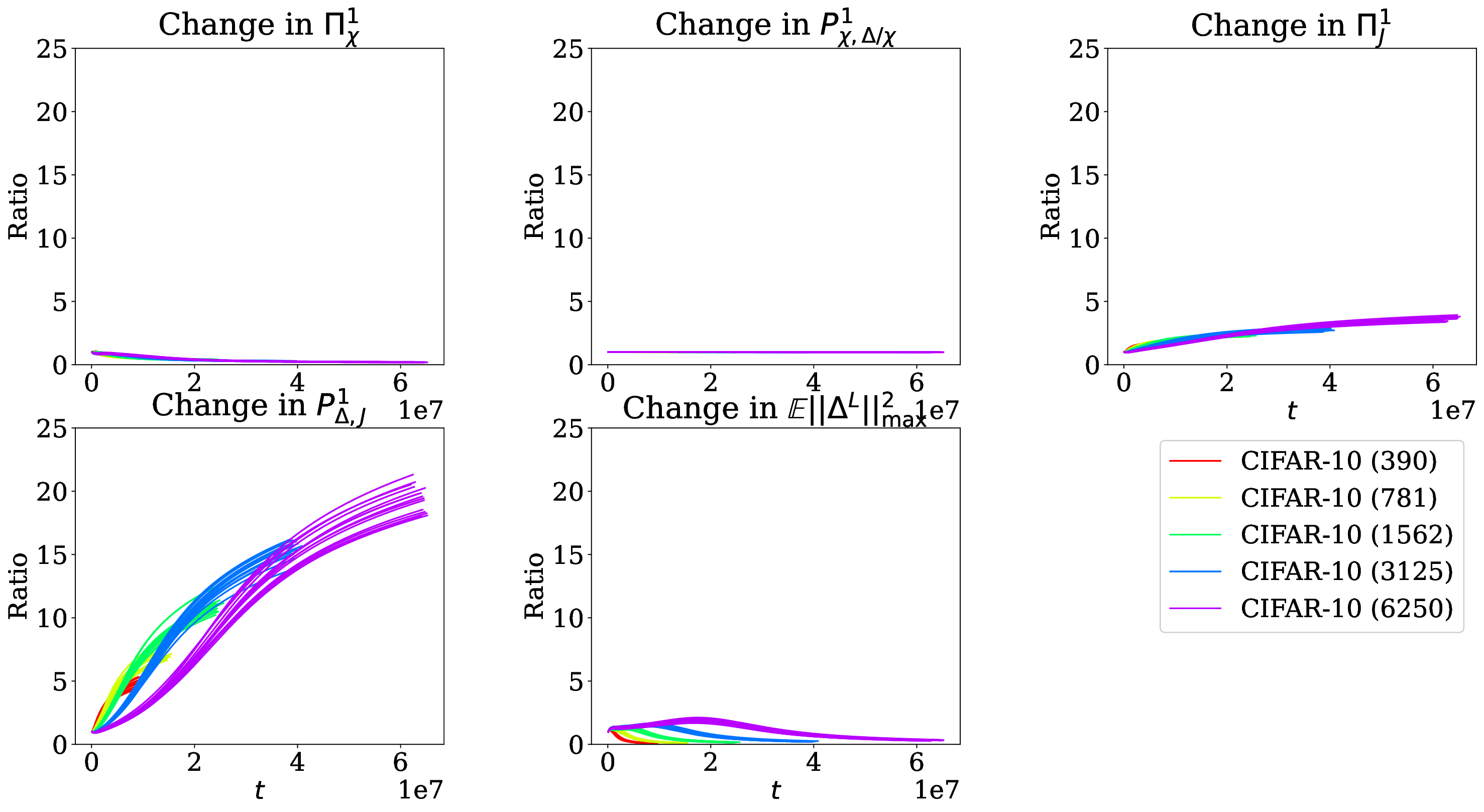}
		\caption{\label{fig:mlp-cifar10-cross-entropy-main}Change in components of $\mathbb{E}||\Delta^1||^2_{\max}$ when trained on CIFAR-10 with cross-entropy: $\Pi_\chi^1$ (top left), $\Rho_{\chi,\Delta/\chi}^1$ (top middle), $\Pi_J^1$ (top right), $\Rho_{\Delta,J}^1$ (bottom left), and $\mathbb{E}||\Delta^L||^2_{\max}$ (bottom middle).}
	\end{center}
\end{figure}

\begin{figure*}[h!]
	\begin{center}
		\includegraphics[width=\textwidth]{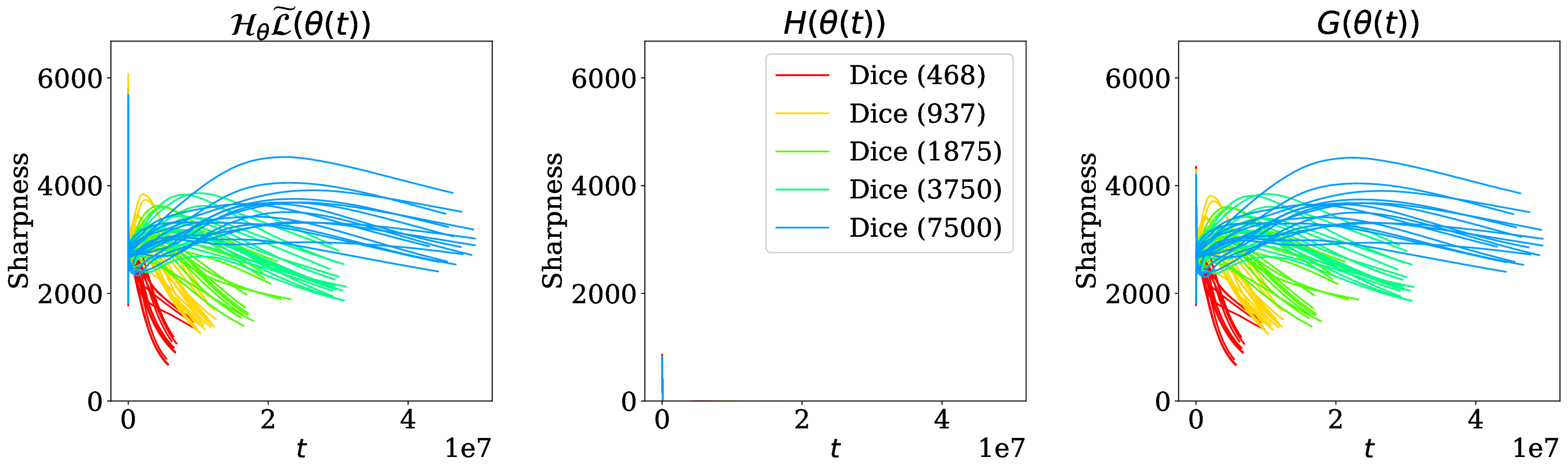}
		\caption{\label{fig:mlp-dice-cross-entropy-components-of-the-hessian}Sharpness of $\mathcal{H}_\theta\widetilde{\mathcal{L}}$ (left), $H$ (middle), and $G$ (right) when trained on dice dataset with cross-entropy}
	\end{center}
\end{figure*}

\begin{figure*}[h!]
	\begin{center}
		\includegraphics[width=\textwidth]{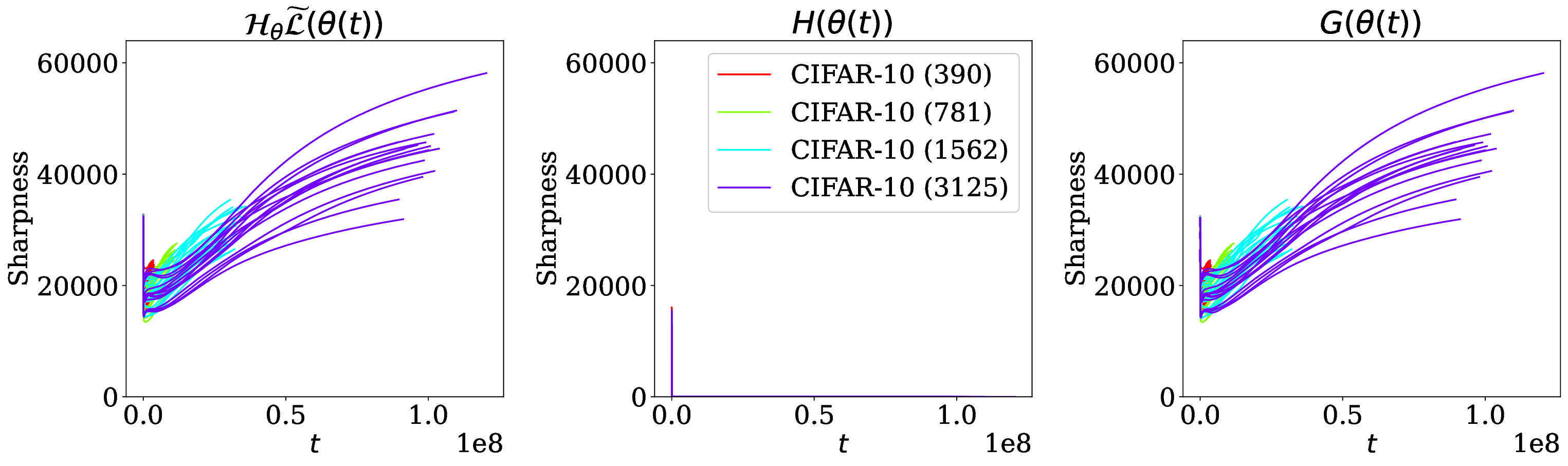}
		\caption{\label{fig:mlp-cifar10-mse-components-of-the-hessian}Sharpness of $\mathcal{H}_\theta\widetilde{\mathcal{L}}$ (left), $H$ (middle), and $G$ (right) when trained on dice dataset with MSE}
	\end{center}
\end{figure*}

\begin{figure*}[h!]
	\begin{center}
		\includegraphics[width=\textwidth]{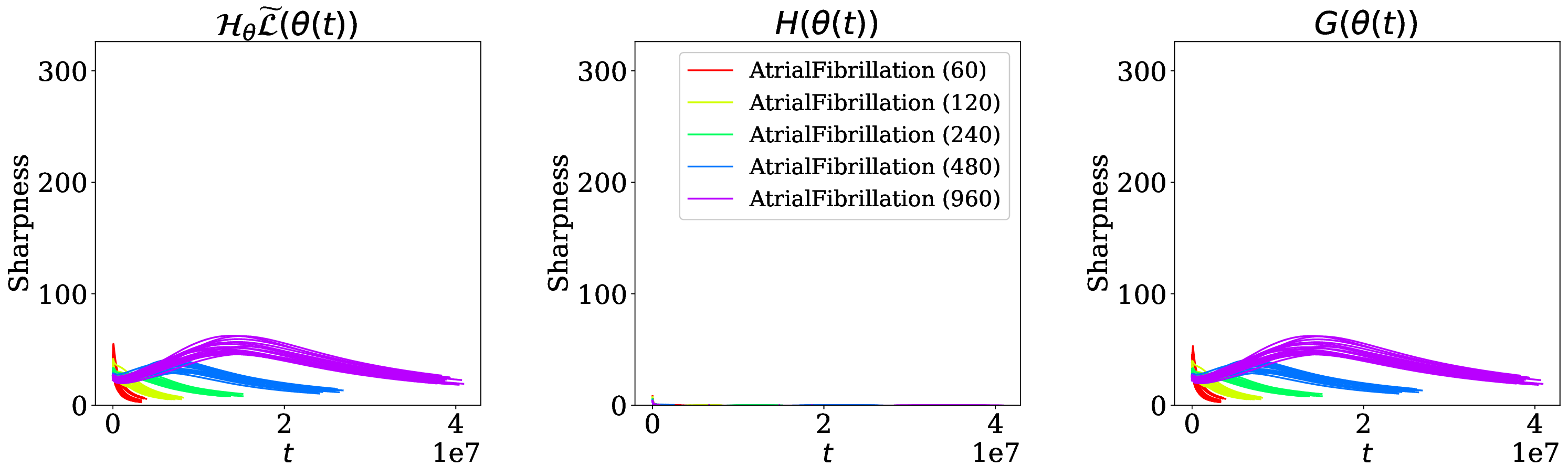}
		\caption{\label{fig:mlp-ucr-cross-entropy-components-of-the-hessian}Sharpness of $\mathcal{H}_\theta\widetilde{\mathcal{L}}$ (left), $H$ (middle), and $G$ (right) when trained on AtrialFibrillation with cross-entropy}
	\end{center}
\end{figure*}

\begin{figure*}[h!]
	\begin{center}
		\includegraphics[width=\textwidth]{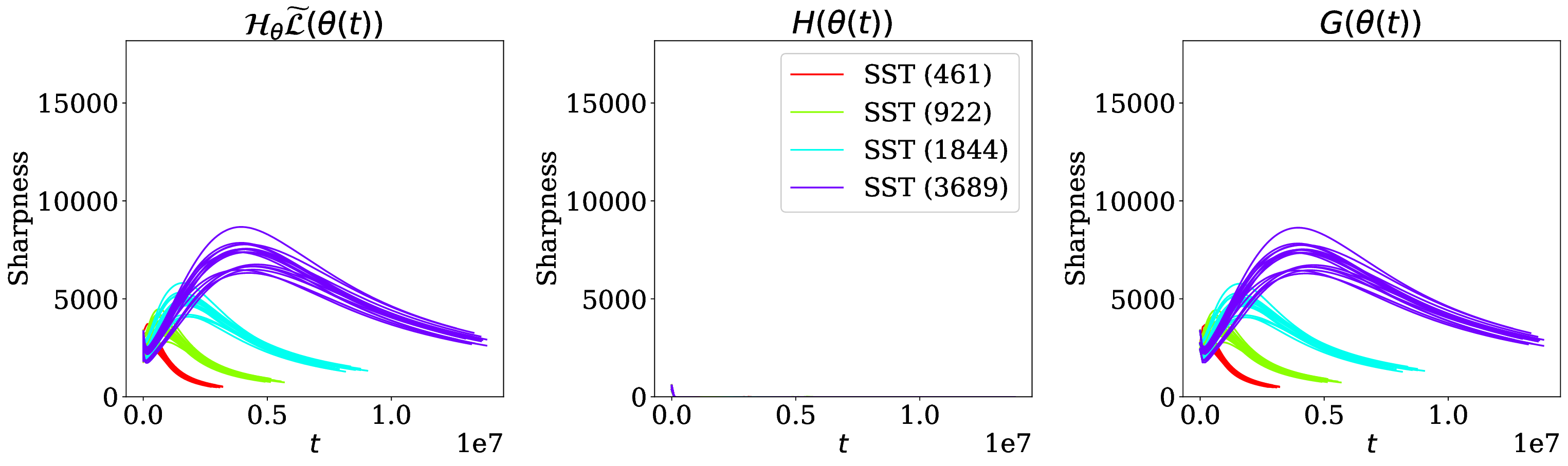}
		\caption{\label{fig:mlp-sst-cross-entropy-components-of-the-hessian}Sharpness of $\mathcal{H}_\theta\widetilde{\mathcal{L}}$ (left), $H$ (middle), and $G$ (right) when trained on SST with cross-entropy}
	\end{center}
\end{figure*}

\begin{figure}[h!]
	\begin{center}
		\includegraphics[width=\textwidth]{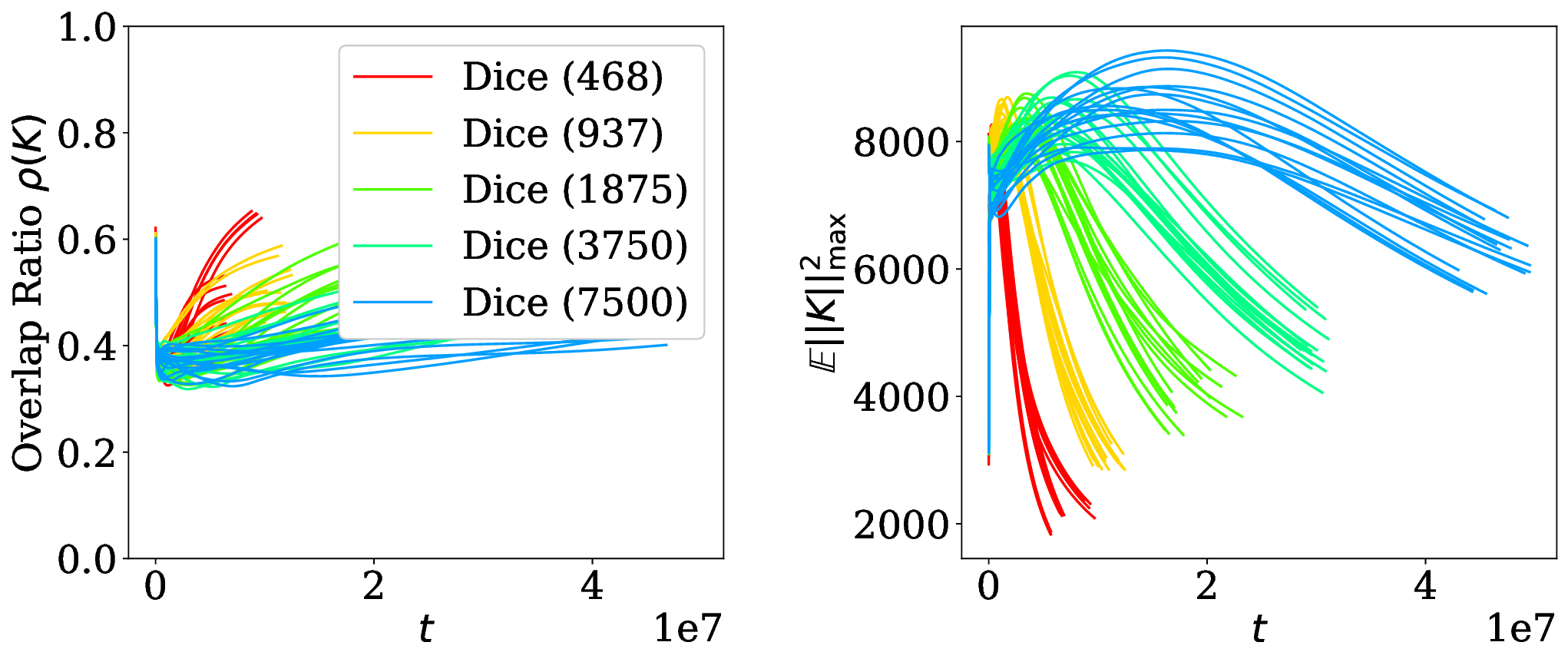}
		\caption{\label{fig:mlp-dice-cross-entropy-overlap-ratio}$\rho(K)$ (left) and $\mathbb{E}||K||^2_{\max}$ (right) when trained on dice dataset with cross-entropy}
	\end{center}
\end{figure}

\begin{figure}[h!]
	\begin{center}
		\includegraphics[width=\textwidth]{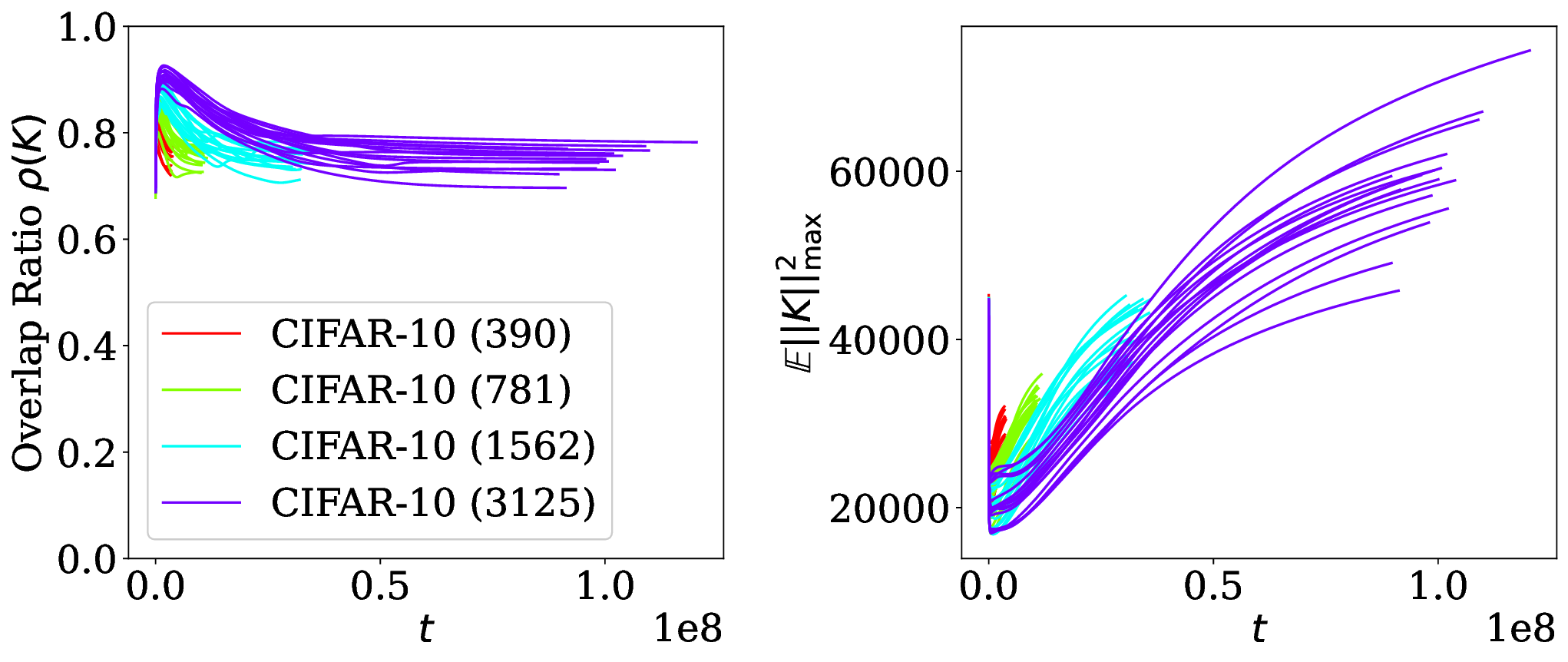}
		\caption{\label{fig:mlp-cifar10-mse-overlap-ratio}$\rho(K)$ (left) and $\mathbb{E}||K||^2_{\max}$ (right) when trained on CIFAR-10 with MSE}
	\end{center}
\end{figure}

\begin{figure}[h!]
	\begin{center}
		\includegraphics[width=\textwidth]{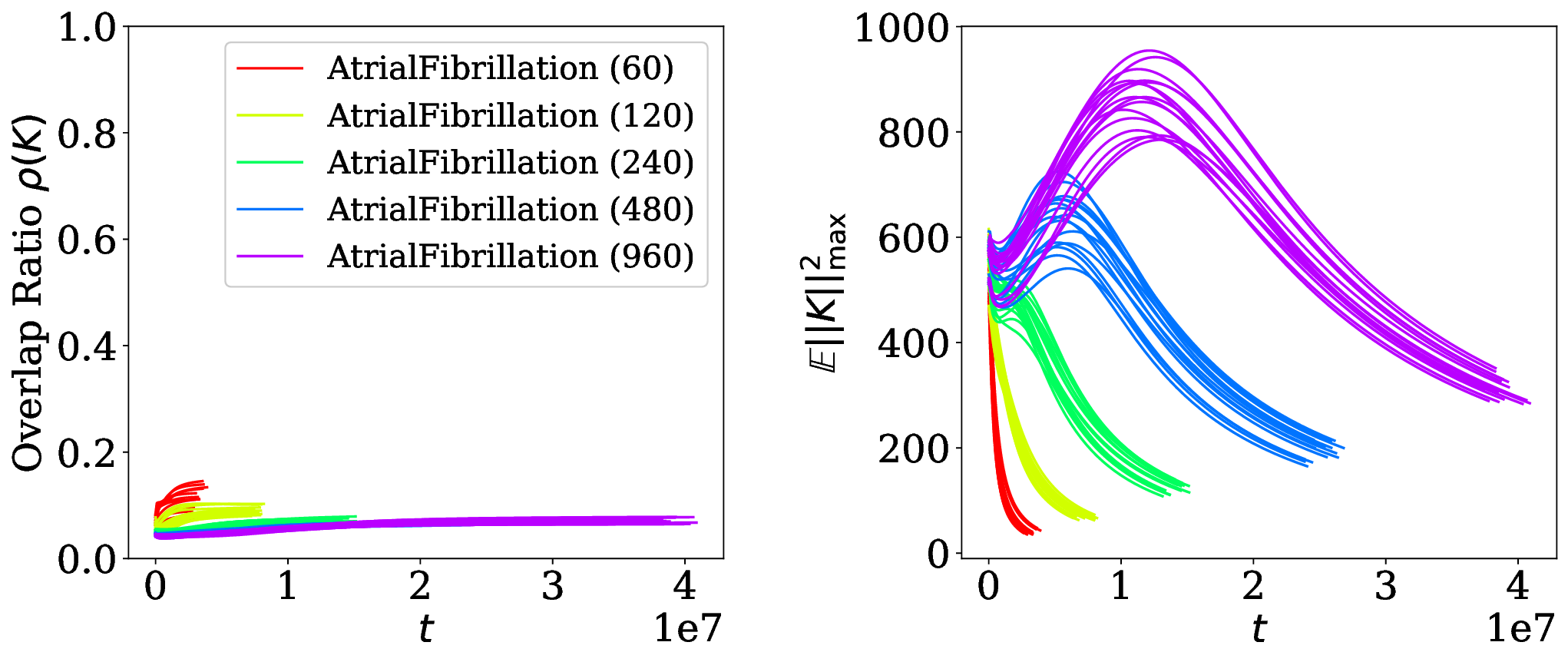}
		\caption{\label{fig:mlp-ucr-cross-entropy-overlap-ratio}$\rho(K)$ (left) and $\mathbb{E}||K||^2_{\max}$ (right) when trained on AtrialFibrillation with cross-entropy}
	\end{center}
\end{figure}

\begin{figure}[h!]
	\begin{center}
		\includegraphics[width=\textwidth]{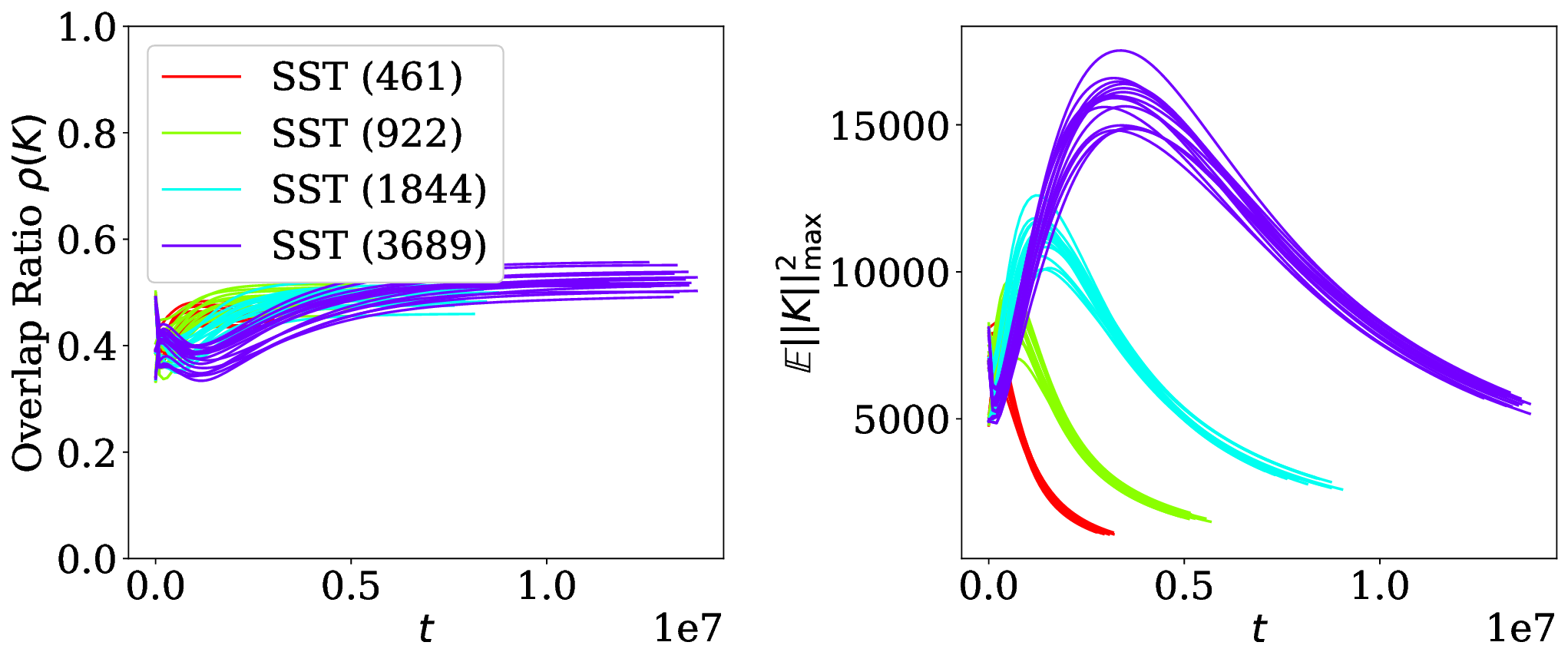}
		\caption{\label{fig:mlp-sst-cross-entropy-overlap-ratio}$\rho(K)$ (left) and $\mathbb{E}||K||^2_{\max}$ (right) when trained on SST with cross-entropy}
	\end{center}
\end{figure}

\begin{figure}
	\begin{center}
		\includegraphics[width=\textwidth]{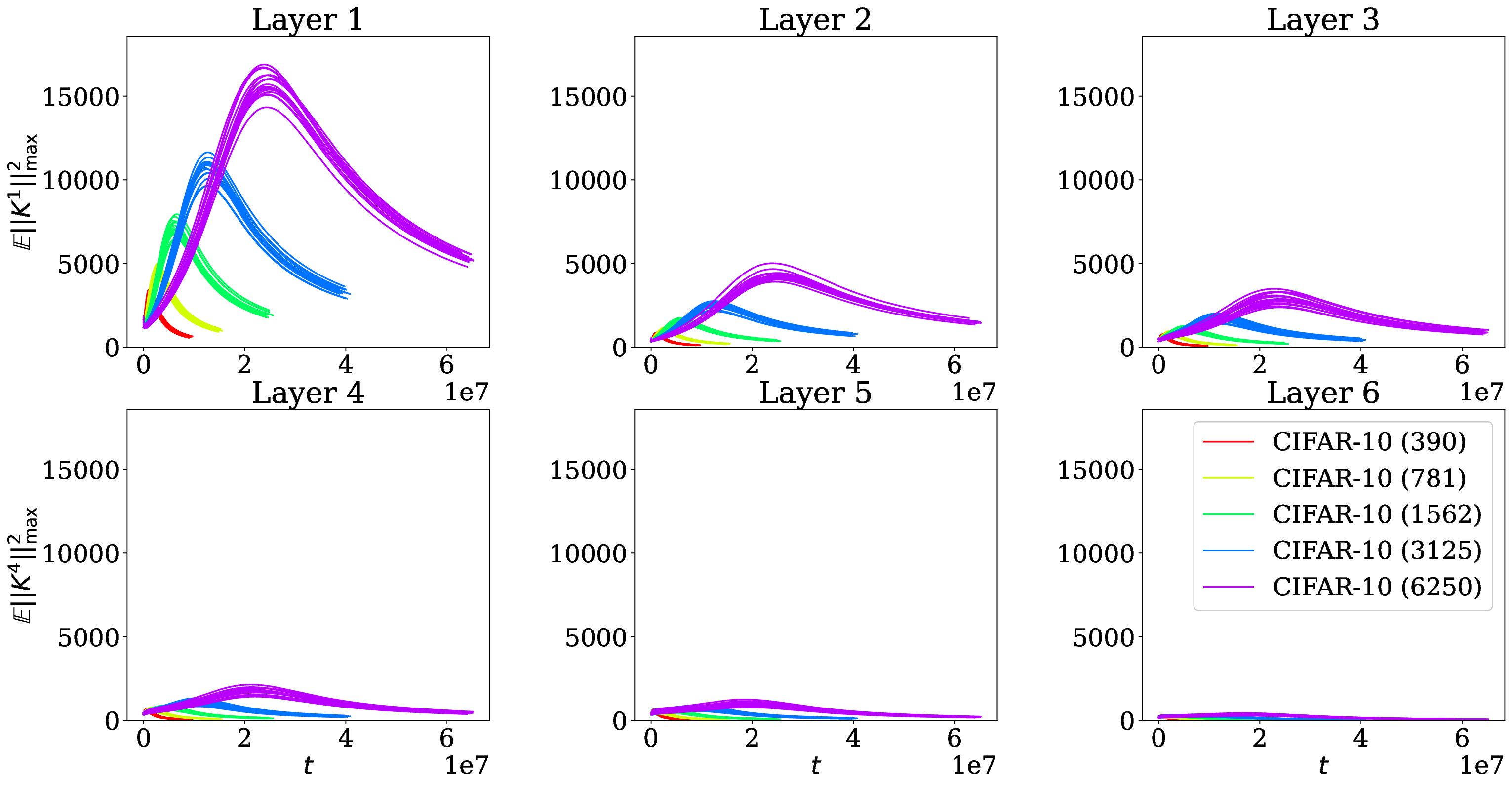}
		\caption{\label{fig:mlp-cifar10-cross-entropy-layerwise-sharpness}$\mathbb{E} ||K_i||^2_{\max}$ for layers $i = 1$ (top left), 2 (top middle), 3 (top right), 4 (bottom left), 5 (bottom middle), and 6 (bottom right) when trained on CIFAR-10 with cross-entropy}
	\end{center}
\end{figure}

\begin{figure}
	\begin{center}
		\includegraphics[width=\textwidth]{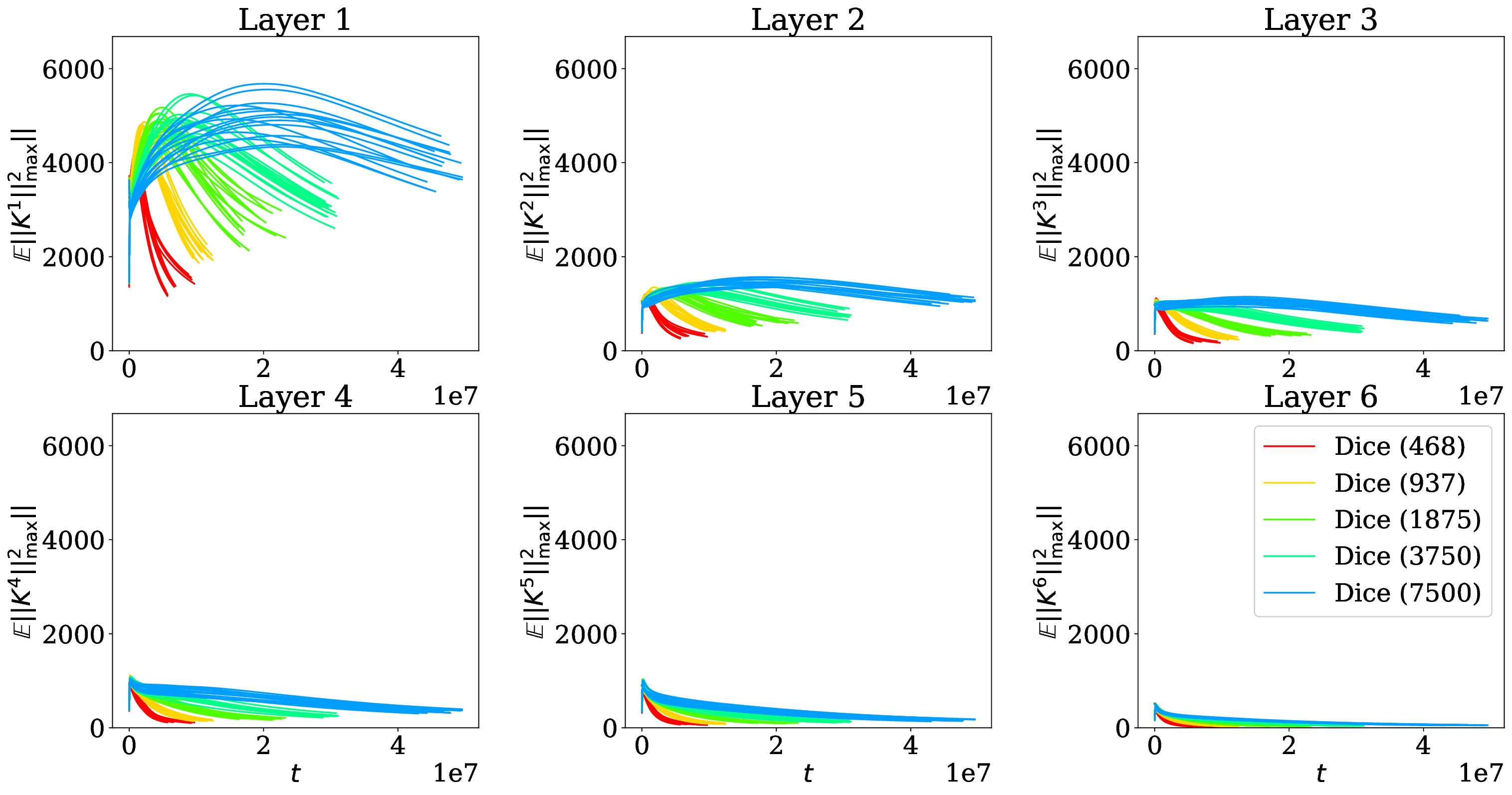}
		\caption{\label{fig:mlp-dice-cross-entropy-layerwise-sharpness}$\mathbb{E} ||K_i||^2_{\max}$ for layers $i = 1$ (top left), 2 (top middle), 3 (top right), 4 (bottom left), 5 (bottom middle), and 6 (bottom right) when trained on dice dataset with cross-entropy}
	\end{center}
\end{figure}

\begin{figure}
	\begin{center}
		\includegraphics[width=\textwidth]{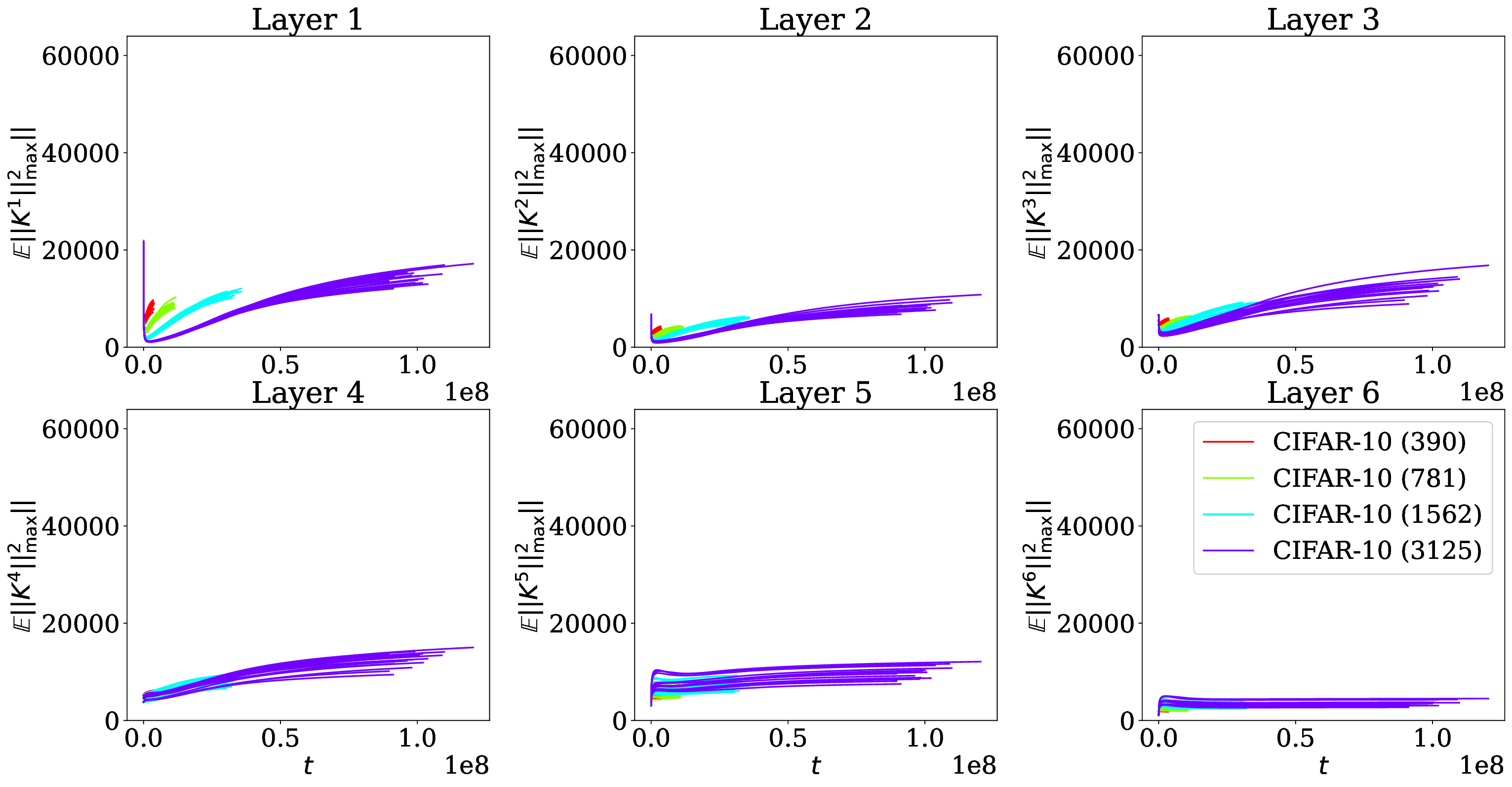}
		\caption{\label{fig:mlp-cifar10-mse-layerwise-sharpness}$\mathbb{E} ||K_i||^2_{\max}$ for layers $i = 1$ (top left), 2 (top middle), 3 (top right), 4 (bottom left), 5 (bottom middle), and 6 (bottom right) when trained on CIFAR-10 with MSE}
	\end{center}
\end{figure}

\begin{figure}
	\begin{center}
		\includegraphics[width=\textwidth]{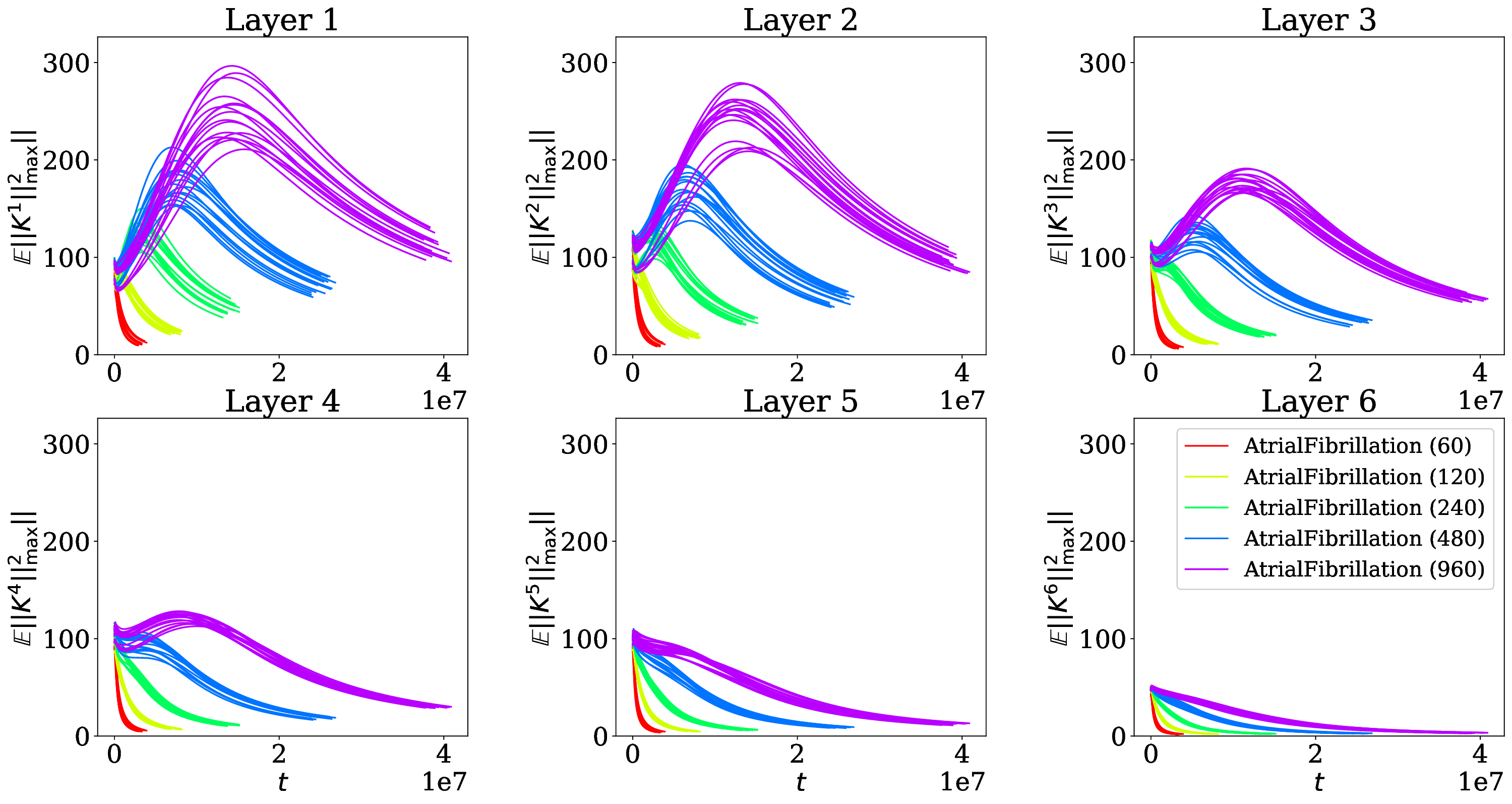}
		\caption{\label{fig:mlp-ucr-cross-entropy-layerwise-sharpness}$\mathbb{E} ||K_i||^2_{\max}$ for layers $i = 1$ (top left), 2 (top middle), 3 (top right), 4 (bottom left), 5 (bottom middle), and 6 (bottom right) when trained on AtrialFibrillation with cross-entropy}
	\end{center}
\end{figure}

\begin{figure}
	\begin{center}
		\includegraphics[width=\textwidth]{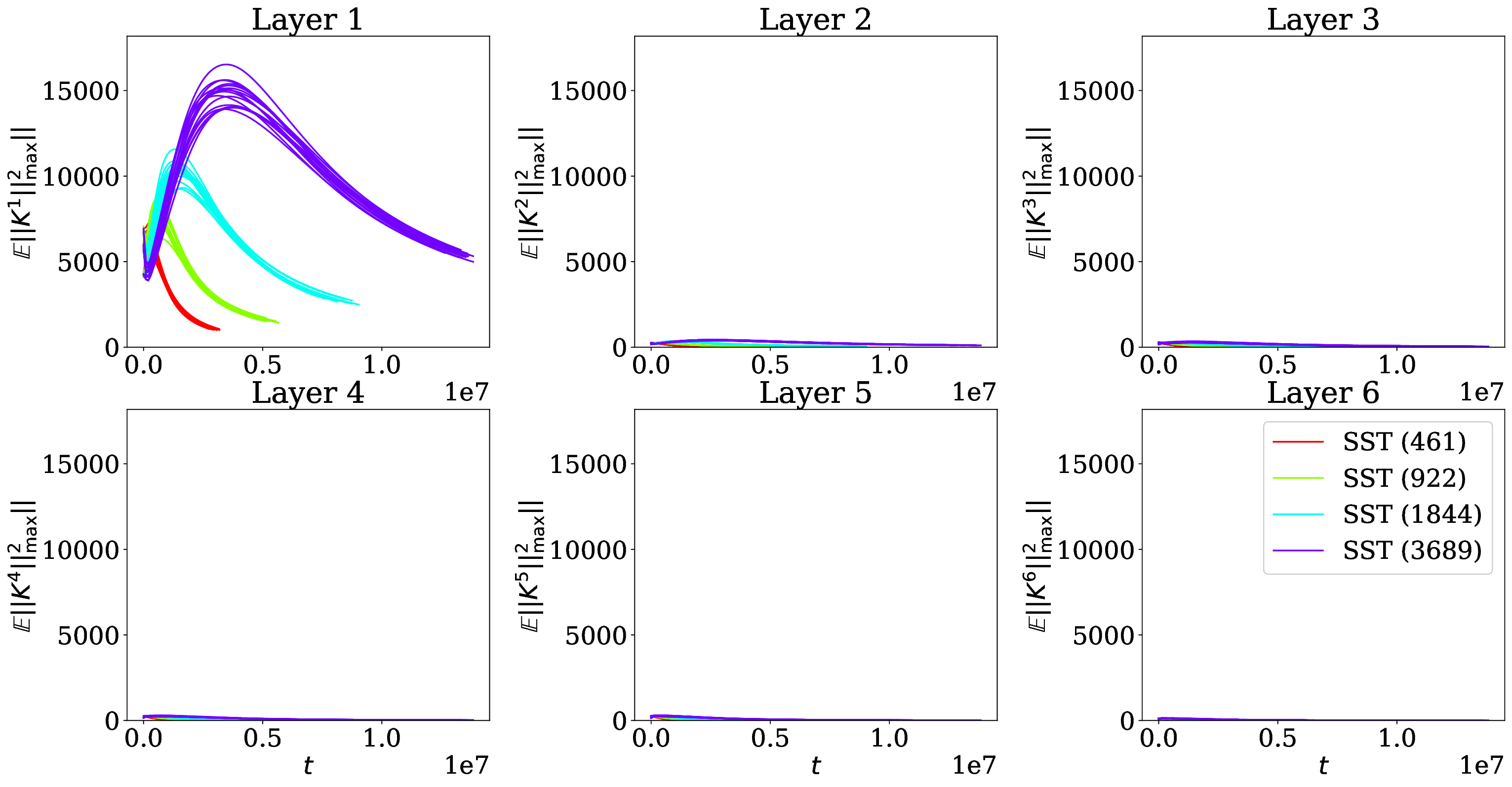}
		\caption{\label{fig:mlp-sst-cross-entropy-layerwise-sharpness}$\mathbb{E} ||K_i||^2_{\max}$ for layers $i = 1$ (top left), 2 (top middle), 3 (top right), 4 (bottom left), 5 (bottom middle), and 6 (bottom right) when trained on SST with cross-entropy}
	\end{center}
\end{figure}

\begin{figure}
	\begin{center}
		\includegraphics[width=\textwidth]{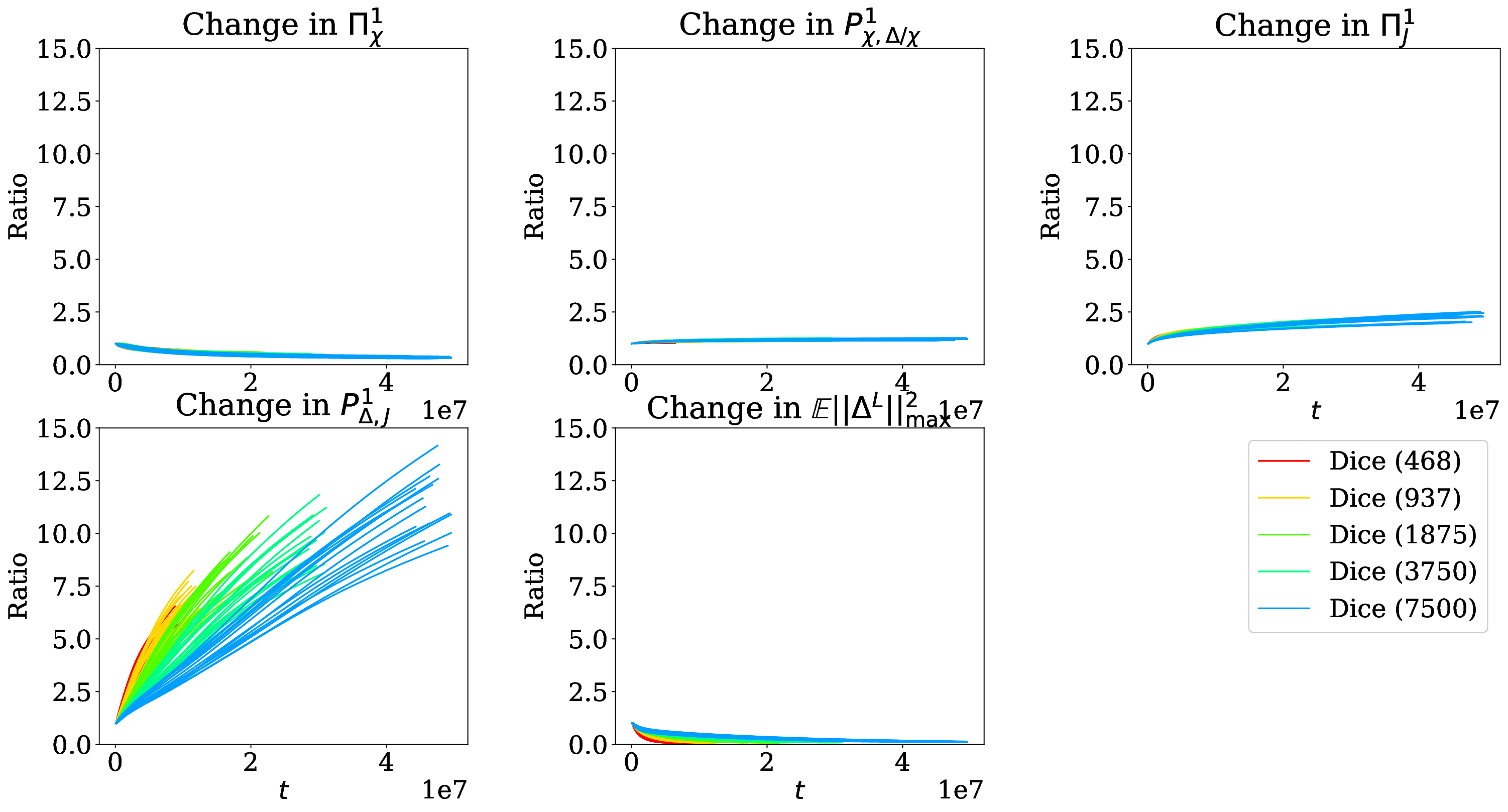}
		\caption{\label{fig:mlp-dice-cross-entropy-main}Change in components of $\mathbb{E}||\Delta^1||^2_{\max}$ when trained on dice dataset with cross-entropy: $\Pi_\chi^1$ (top left), $\Rho_{\chi,\Delta/\chi}^1$ (top middle), $\Pi_J^1$ (top right), $\Rho_{\Delta,J}^1$ (bottom left), and $\mathbb{E}||\Delta^L||^2_{\max}$ (bottom middle).}
	\end{center}
\end{figure}

\begin{figure}
	\begin{center}
		\includegraphics[width=\textwidth]{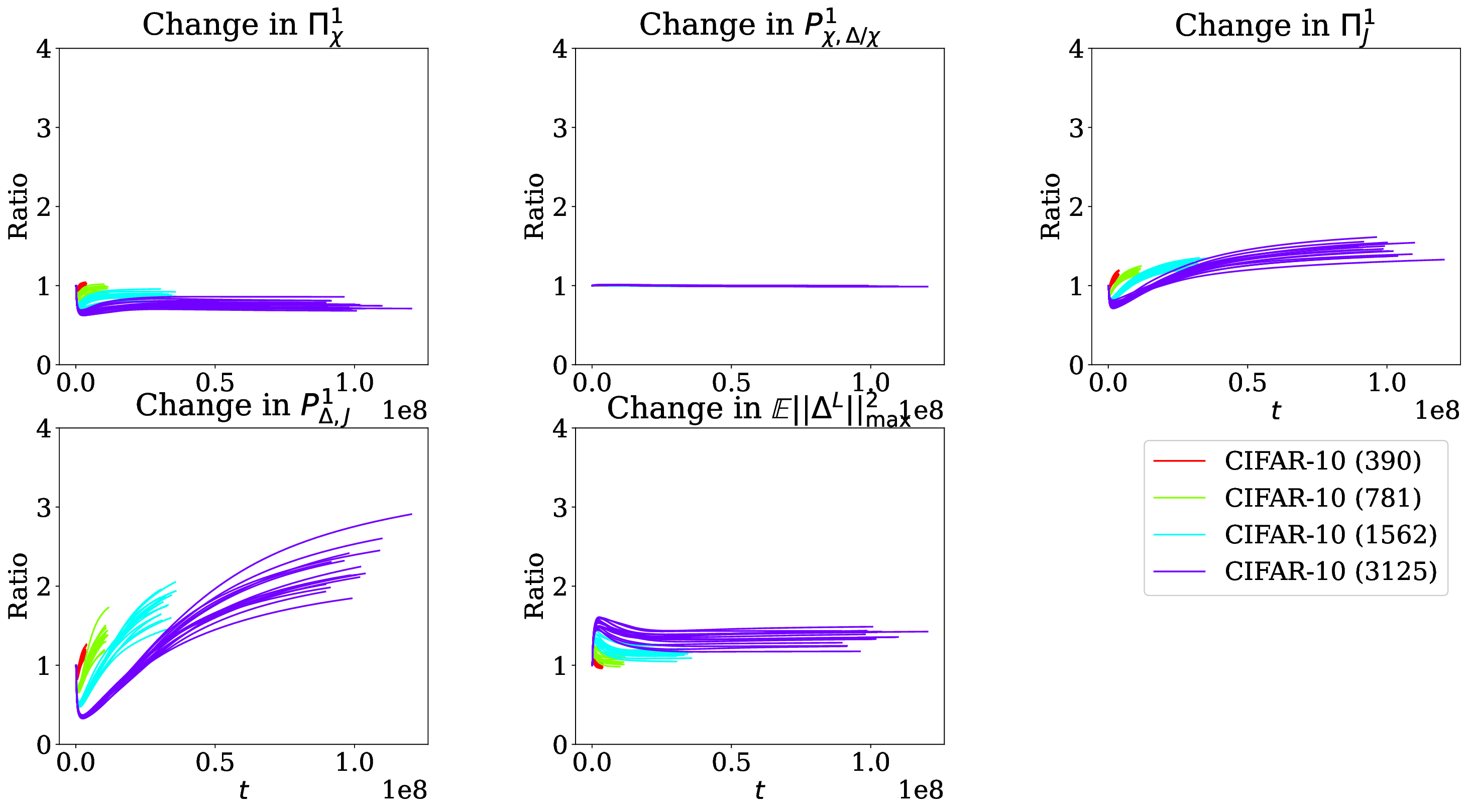}
		\caption{\label{fig:mlp-cifar10-mse-main}Change in components of $\mathbb{E}||\Delta^1||^2_{\max}$ when trained on CIFAR-10 with MSE: $\Pi_\chi^1$ (top left), $\Rho_{\chi,\Delta/\chi}^1$ (top middle), $\Pi_J^1$ (top right), $\Rho_{\Delta,J}^1$ (bottom left), and $\mathbb{E}||\Delta^L||^2_{\max}$ (bottom middle).}
	\end{center}
\end{figure}

We plot results from the experiments using CIFAR-10 and the cross-entropy loss in the main body, with other experiments plotted in the Supplemental Material section. In each figure, we plot each experiment as a separate curve, with the color of the curve denoting the dataset size. We plot the sharpness of $\mathcal{H}_\theta\widetilde{\mathcal{L}}$, $G$, and $H$ in Figures~\ref{fig:mlp-cifar10-cross-entropy-components-of-the-hessian}, \ref{fig:mlp-dice-cross-entropy-components-of-the-hessian}, \ref{fig:mlp-cifar10-mse-components-of-the-hessian}, \ref{fig:mlp-ucr-cross-entropy-components-of-the-hessian}, and \ref{fig:mlp-sst-cross-entropy-components-of-the-hessian}. We plot $\rho(K)$ and $\mathbb{E} ||K||^2_{\max}$ in Figures~\ref{fig:mlp-cifar10-cross-entropy-overlap-ratio}, \ref{fig:mlp-dice-cross-entropy-overlap-ratio}, \ref{fig:mlp-cifar10-mse-overlap-ratio}, \ref{fig:mlp-ucr-cross-entropy-overlap-ratio}, and \ref{fig:mlp-sst-cross-entropy-overlap-ratio}. We plot $\mathbb{E}||K^i||^2_{\max}$ for layers 1, 3, and 5 in Figure~\ref{fig:mlp-cifar10-cross-entropy-layerwise-sharpness-cutdown}. We plot $\mathbb{E}||K^i||^2_{\max}$ for all layers $i$ for all experiments in Figures~\ref{fig:mlp-cifar10-cross-entropy-layerwise-sharpness}, \ref{fig:mlp-dice-cross-entropy-layerwise-sharpness}, \ref{fig:mlp-cifar10-mse-layerwise-sharpness}, \ref{fig:mlp-ucr-cross-entropy-layerwise-sharpness}, and \ref{fig:mlp-sst-cross-entropy-layerwise-sharpness}. We plot each of the five components of $\mathbb{E} ||\Delta^1||^2_{\max}$ from Equation~\ref{eqn:main} in Figure~\ref{fig:mlp-cifar10-cross-entropy-main}. We plot analogous figures for the other experiments in Figures~\ref{fig:mlp-dice-cross-entropy-main}, \ref{fig:mlp-cifar10-mse-main}, \ref{fig:mlp-ucr-cross-entropy-main}, and \ref{fig:mlp-sst-cross-entropy-main} in the Supplemental Material section.

\section{Discussion}
\label{sec:discussion}

In all of our experiments, we observe an initial transient in which sharpness fluctuates wildly. Since this period typically lasts no more than 10 iterations, we ignored the first ten iterations in our analyses.

% Network:  Dataset:     Criterion: Max H Sharpness: Max H/Hess Ratio:  Max G-Hess Err:  Max Hess Increase:
%     MLP      Dice  Cross-Entropy   0.00            0.000              0.101            x1.41
%     MLP  CIFAR-10  Cross-Entropy  19.60            0.021              0.095            x5.45
%     MLP  CIFAR-10            MSE  56.74            0.003              0.041            x2.46
%     MLP       UCR  Cross-Entropy   2.79            0.099              0.068            x2.42
%     MLP       SST  Cross-Entropy  16.95            0.009              0.016            x3.07

The sharpness of $\mathcal{H}_\theta\widetilde{\mathcal{L}}$ can increase by as much as $\times5.45$ over training. Recall from Equation~\ref{eqn:g-plus-h} that $\mathcal{H}_\theta\widetilde{\mathcal{L}} = H + G$. Consistent with \cite{edge-of-stability}, we find that the increase in the sharpness of $\mathcal{H}_\theta\widetilde{\mathcal{L}}$ is caused by an increase in the sharpness of $G$. As shown in Figure~\ref{fig:mlp-cifar10-cross-entropy-components-of-the-hessian}, the sharpness of $H$ is so low that it is hard to see on the graph. After the initial fluctuations, the sharpness of $H$ never exceeds 56.74, or more than 9.9\% of the sharpness of $\mathcal{H}_\theta\widetilde{\mathcal{L}}$, in any experiment. The sharpness of $G$ is within 10.1\% of the sharpness of $\mathcal{H}_\theta\widetilde{\mathcal{L}}$ at all times in all experiments.

% Network:  Dataset:     Criterion: Max rho(K) change:
%     MLP      Dice  Cross-Entropy  x1.70
%     MLP  CIFAR-10  Cross-Entropy  x1.00
%     MLP  CIFAR-10            MSE  x1.15
%     MLP       UCR  Cross-Entropy  x1.90
%     MLP       SST  Cross-Entropy  x1.43

Recall from Equation~\ref{eqn:rho} that the sharpness of $G$ is equal to the product $\rho(K) \cdot \mathbb{E} ||K||^2_{\max}$. As shown in Figure~\ref{fig:mlp-cifar10-cross-entropy-overlap-ratio}, the increase in the sharpness of $G$ is caused by an increase in $\mathbb{E} ||K||^2_{\max}$. The overlap ratio $\rho(K)$ never increases by more than a factor of $\times1.90$, insufficient to explain the rise in the sharpness of $G$.

% Network:  Dataset:     Criterion: Max Ratio:
%     MLP      Dice  Cross-Entropy  x0.31
%     MLP  CIFAR-10  Cross-Entropy  x0.30
%     MLP  CIFAR-10            MSE  x1.07
%     MLP       UCR  Cross-Entropy  x1.37
%     MLP       SST  Cross-Entropy  x0.05

Recall from Equation~\ref{eqn:kkt} that the squared operator norm $KK^T$ can be decomposed into the sum of layerwise components $K_1K_1^T + ... + K_LK_L^T$. Consistent with \cite{layerwise-sharpness}, we observe that the highest operator norm is usually found in the first layer. $\mathbb{E}||K^i||^2_{\max}$ for $i > 1$ never exceeds $\mathbb{E}||K^1||^2_{\max}$ by more than 37\%. By the triangle inequality:
\begin{equation}\label{eqn:triangle-inequality}
\mathbb{E}||K^1||_{\max}^2 - \sum_{i=2}^{L} \mathbb{E}||K^i||_{\max}^2 \leq \mathbb{E} ||K||_{\max}^2 \leq\mathbb{E}||K^1||_{\max}^2 + \sum_{i=2}^{L} ||K^i||_{\max}^2
\end{equation}
Since $\mathbb{E}||K||_{\max}^2$ is usually dominated by $\mathbb{E}||K^1||_{\max}^2$, the increase in the sharpness of $\mathcal{H}_\theta \widetilde{\mathcal{L}}(\theta)$ can be analyzed by examining $\mathbb{E}||K^1||_{\max}^2 = \mathbb{E}||\Delta^1||^2_{\max}$. Recall from Equation~\ref{eqn:main} that $\mathbb{E} ||\Delta^1||^2_{\max}$ can be decomposed into five components:

\begin{itemize}

% Network:  Dataset:     Criterion: Max Delta^L increase:
%     MLP      Dice  Cross-Entropy  x1.019
%     MLP  CIFAR-10  Cross-Entropy  x2.115
%     MLP  CIFAR-10            MSE  x1.609
%     MLP       UCR  Cross-Entropy  x1.004
%     MLP       SST  Cross-Entropy  x1.086

	\item The behavior of $\mathbb{E} ||\Delta^L||^2_{\max}$ is variable, but it never increases by more than $\times2.12$.

% Network:  Dataset:     Criterion: Max chi increase:
%     MLP      Dice  Cross-Entropy  x1.016
%     MLP  CIFAR-10  Cross-Entropy  x1.057
%     MLP  CIFAR-10            MSE  x1.042
%     MLP       UCR  Cross-Entropy  x1.000
%     MLP       SST  Cross-Entropy  x1.329

	\item $\Pi_\chi^1$ is consistently stable or declining, never increasing by more than $\times1.33$.

% Network:  Dataset:     Criterion: Max r(chi, Delta/chi) increase:
%     MLP      Dice  Cross-Entropy  x1.272
%     MLP  CIFAR-10  Cross-Entropy  x1.016
%     MLP  CIFAR-10            MSE  x1.013
%     MLP       UCR  Cross-Entropy  x1.001
%     MLP       SST  Cross-Entropy  x1.011

	\item $\Rho_{\chi,\Delta/\chi}^1$ is consistently stable, never increasing by more than $\times 1.27$.

% Network:  Dataset:     Criterion: Max layerwise Jacobian norm increase:
%     MLP      Dice  Cross-Entropy  x2.515
%     MLP  CIFAR-10  Cross-Entropy  x3.926
%     MLP  CIFAR-10            MSE  x1.613
%     MLP       UCR  Cross-Entropy  x2.359
%     MLP       SST  Cross-Entropy  x2.026

	\item $\Pi_J^1$ is consistently increasing, but never by more than $\times3.93$.

% Network:  Dataset:     Criterion: Max Align Change: Max Always Stable LR:
%     MLP      Dice  Cross-Entropy  x14.160           0.00061
%     MLP  CIFAR-10  Cross-Entropy  x21.312           0.00115
%     MLP  CIFAR-10            MSE  x2.910            0.00004
%     MLP       UCR  Cross-Entropy  x26.902           0.03821
%     MLP       SST  Cross-Entropy  x28.970           0.00062

	\item $\Rho_{\Delta,J}^1$ is consistently increasing, by as much as $\times28.97$.

\end{itemize}

If we assume $\lambda_1(\mathcal{H}_\theta\widetilde{\mathcal{L}}) \approx \lambda_1(G)$, we can use Equations~\ref{eqn:rho}, \ref{eqn:kkt}, \ref{eqn:main}, and \ref{eqn:triangle-inequality} to bound the sharpness of $\mathcal{H}_\theta\widetilde{\mathcal{L}}$ under the counterfactual that $\Rho_{\delta,J}^i$ remains constant:
$$
\lambda_1(\mathcal{H}_\theta\widetilde{\mathcal{L}}) \leq
\rho(K) \cdot \sum_i \frac{\Rho_{\Delta,J}^i(t)}{\Rho_{\Delta,J}^i(0)} \mathbb{E} ||\Delta^i||^2_{\max}
$$
Using this bound, we find that if $\Rho^i_{\Delta,J}$ did not increase after the initial transient period, our network training with cross-entropy loss would be stable for $\eta < 0.0006$ in all of our experiments, and our network training with MSE would be stable for $\eta < 0.00003$ in all experiments. We therefore attribute the training on the edge of stability phenomenon primarily to the increase in $\Rho_{\Delta,J}^1$.

Recall that $\Rho_{\Delta,J}^1$ is the product of the alignment ratios of $\Delta^{i+1}$ and the layerwise Jacobian $\frac{\partial \hat{x}^{i+1}}{\partial \hat{x}^i}$. This term measures the overlap between the image of $\Delta^{i+1}$ and the top singular vectors of the layerwise Jacobian. Since the layerwise Jacobian is ill-conditioned, when the network is first initialized, much of the image of $\Delta^{i+1}$ will tend to fall onto singular vectors with low singular values. As training proceeds, the singular vectors rotate so that more of the image falls onto singular vectors with high singular values. This implies that a gradient signal at the output of the network will increase in magnitude as it is backpropagated to the inputs of the network. Equivalently, for any input $x\in\mathcal{T}$, there will exist a subspace so that a small change in $x$ along that subspace will result in a large change in the network outputs. This is reminescent of past theoretical work that showed such a phenomenon would occur on deep linear networks \citep{layerwise-alignment,layerwise-alignment-2}. However, to the best of our knowledge this has never previously been observed in nonlinear networks, because training on the edge of stability puts a limit on the maximum sharpness of $\mathcal{H}_\theta\widetilde{\mathcal{L}}$, and hence on the maximum value that $\Rho_{\Delta,J}^i$ can attain.

Further, the degree of alignment consistently scales with the dataset size, following a power law of the form \citep{scaling-laws-1,scaling-laws-4,scaling-laws-5}:
\begin{equation}\label{eqn:power-law}
	\max \Rho^1_{\Delta,J} \approx c_1 D^{c_2}
\end{equation}
As shown in Table~\ref{tab:results}, this power law holds with a coefficient of determination between 0.74 and 0.98 in all experiments. We hypothesize that this occurs because the layerwise Jacobian alignment is correlated to the complexity of the decision boundary, but leave further investigation to future work.

\begin{table}
	\begin{center}
	\begin{tabular}{ll|lll}
		Dataset&Criterion&$c_1$&$c_2$&$R^2$\\
		\hline
			      Dice&Cross-Entropy&0.0017&0.23&0.84\\ % EoS-67
		          CIFAR-10&Cross-Entropy&0.0004&0.49&0.97\\ % EoS-68
		          CIFAR-10&          MSE&0.0004&0.20&0.74\\ % EoS-69
		AtrialFibrillation&Cross-Entropy&0.0026&0.45&0.93\\ % EoS-74
		               SST&Cross-Entropy&0.0001&0.67&0.98\\ % EoS-75
		%AlexNet&              Dice&Cross-Entropy& 0.39&5.50&0.97   % EoS-56
	\end{tabular}
	\end{center}
	\caption{\label{tab:results}\textbf{Power Laws for Peak Layerwise Jacobian Alignment}. $c_1, c_2$ are coefficients from Equation~\ref{eqn:power-law}. $R^2$ is coefficient of determination.}
\end{table}

\section{Conclusion}
\label{sec:conclusion}

In this paper, we have explored the behavior of neural networks trained by solving the actual gradient flow, without falling into the edge of stability. We have used those experiments to show that training on the edge of stability is caused by alignment of the layerwise Jacobian matrices of the network, and that that alignment increases with dataset size.

\textbf{Acknowledgements:} This work was supported in part by high-performance computer time and resources from the DoD High Performance Computing Modernization Program.

\small{Approved for public release, NGA-U-2024-01088. The views expressed in this paper do not necessarily reflect those of the National Geospatial-Intelligence Agency, the Department of Defense, or any other department or agency of the US Government.}

\bibliography{citations}{}
\bibliographystyle{plain}

\section{Supplemental Material}

\subsection{Calculating the Eigenvalues}
\label{sec:power-iteration}

We use the power iteration method to calculate the top $k$ eigenvalues and eigenvectors of $\mathcal{H}_\theta\widetilde{\mathcal{L}}(\theta)$ by absolute value $\tilde{\lambda}_1, ..., \tilde{\lambda}_k$ \citep{power-iteration}:

\begin{enumerate}

	\item Randomly initialize an $N\times k$ matrix $\tilde{V}_0$ and initialize $\tilde{\Lambda}_0$ as a $k$-dimensional vector whose entries are $\inf$. Let $i = 0$.

	\item Calculate the matrix-vector product $W_i = A\tilde{V}_i$.

	\item Calculate the QR decomposition of $W_i$ as $Q_{i + 1}, \tilde{V}_{i+1}$.

	\item Set $\tilde{\Lambda}_{i + 1}$ to be equal to the diagonal entries of $Q_{i + 1}$.

	\item Calculate the difference between $\tilde{\Lambda}_{i + 1}, \tilde{\Lambda}_i$. If this falls below a threshold, terminate the algorithm: the eigenvalues are the entries of $\tilde{\Lambda}_{i + 1}$, and the eigenvectors are the columns of $\tilde{V}_{i + 1}$. If not, increment $i$ and return to step 2.

\end{enumerate}

Since the size of $\mathcal{H}_\theta\widetilde{\mathcal{L}}(\theta)$ is quadratic with the number of network parameters $P$, it is too big to be materialized in memory for neural networks of practical interest. However, the power iteration method requires only the calculation of matrix-vector products, which are more tractable. For a vector $v$, the $\mathcal{H}_\theta\widetilde{\mathcal{L}}(\theta)$-vector product is given by:
$$
(\mathcal{H}_\theta \widetilde{\mathcal{L}})v = \underset{x,y\sim\mathcal{T}}{\mathbb{E}} \left((\mathcal{H}_\theta(l(f(x, \theta), y)))v\right)
$$
For an individual data point $(x, y) \in \mathcal{T}$, we can calculate the Hessian-vector product $(\mathcal{H}_\theta(l(f(x, \theta), y)))v$ efficiently using the PyTorch autograd engine \citep{pytorch}. The product $(\mathcal{H}_\theta \widetilde{\mathcal{L}})v$ is then obtained by averaging over the individual Hessian-vector products for all $(x, y) \in \mathcal{T}$.

This procedure is expensive, and as discussed in Section~\ref{sec:exponential-integration}, we will perform it many times over the course of training. However, we can substantially reduce the number of iterations required by reusing the eigenvectors from the previous calculation in the next iteration to initialize $\tilde{V}_0$. Since the eigenvectors change slowly, the power iteration method will generally require only 1 -- 3 iterations to converge.

% Experiment:                  % NaN Eigenvalues:
% AlexNet, Dice, Cross-Entropy 0.000090           (EoS-56)
% MLP, Dice, Cross-Entropy     0.000026           (EoS-67)
% MLP, CIFAR-10, MSE           0.000003           (EoS-69)
% MLP, CIFAR-10, Cross-Entropy 0.000013           (EoS-68)
% MLP, UCR, Cross-Entropy      0.000001           (EoS-74)
% MLP, SST, Cross-Entropy      0.000006           (EoS-75)

Since the power iteration method finds the highest eigenvalues by \emph{absolute value}, on rare occasions it will return a negative eigenvalue. This never occurs for more than 0.009\% of the iterations in any experiment. We treat these as NaN values and ignore them in our analyses.

We can calculate the eigenvalues of $G(\theta), H(\theta)$ similarly. The $G(\theta)$-vector product is given by:
$$
G(\theta)v = \underset{x,y\sim\mathcal{T}}{\mathbb{E}}\left(
\left(\frac{\partial f}{\partial \theta}\right)^T
\left(\mathcal{H}_z l(f(x, \theta), y)\right)
\left(\frac{\partial f}{\partial \theta}\right) v
\right)
$$
The Jacobian-vector product can be efficiently calculated using the PyTorch autograd engine, and the Hessian of the criterion can be calculated explicitly. We can then calculate the product for each individual data point $(x, y) \in \mathcal{T}$ and take the mean, just as with the Hessian-vector product. The $H(\theta)$-vector product can be obtained by calculating the $\mathcal{H}_\theta\widetilde{\mathcal{L}}(\theta)$-vector product and the $G(\theta)$-vector product and subtracting them:
$$
	H(\theta) v = (\mathcal{H}_\theta\widetilde{\mathcal{L}}(\theta)) v - G(\theta) v
$$

The matrix $K(x, \theta, y)$ is small enough to be materialized in memory. The Jacobian $\frac{\partial f}{\partial \theta}(x, \theta)$ can be efficiently calculated in a batchwise fashion using functorch \citep{functorch}, and the Hessian $\mathcal{H}_zl(f(x, \theta), y)$ can be calculated explicitly for a desired loss function. $R$ can then be calculated by performing a matrix square root using the eigendecomposition. Matrix multiplication then gives $K(x, \theta, y)$. Since the nonzero eigenvalues of $K^TK$ are equal to the eigenvalues of $KK^T$, we can multiply $K$ by $K^T$ to obtain a small matrix whose eigenvalues we can calculate explicitly. (We found this procedure to be marginally more efficient than calculating the singular values directly.) By calculating these eigenvalues for every $x, y$, we obtain $\mathbb{E} ||K(x, \theta, y)||_{\max}^2$. We can then obtain the overlap ratio $\rho(K)$ by:
$$
\rho(K) = \frac{\lambda^{1}(G(\theta))}{\mathbb{E} ||K(x, \theta, y)||_{\max}^2}
$$

Given the Jacobian matrices $\frac{\partial f}{\partial \theta}$ calculated for the previous step, we can calculate the Jacobian matrices $\frac{\partial f}{\partial A^i}, \frac{\partial f}{\partial b^i}$ by breaking $\frac{\partial f}{\partial \theta}$ into blocks corresponding to those parameters. The $K^i$ matrices can then be calculated by left-multiplying by $R$, giving us $\mathbb{E} ||K^i||_{\max}^2 = \mathbb{E} ||\Delta^i||_{\max}^2$.

We calculate the expected squared activation ratios $\mathbb{E} (\chi^i(x, \theta))^2$ explicitly, by performing a forward pass for each $x, y$, recording the norms of its activations, taking the ratio, and averaging over $x, y \sim \mathcal{T}$.

We calculate the expected squared operator norm of the layerwise Jacobians using functorch and the PyTorch autograd engine to calculate the layerwise Jacobians $\frac{\partial \hat{x}^{i+1}}{\partial \hat{x}^i}$ just as we calculated $\frac{\partial f}{\partial \theta}.$ We reuse the activations from the calculation of the expected activation ratio to avoid needing a second forward pass. We then use the PyTorch implementation of the LOBPCG algorithm \citep{lobpcg} to calculate the top singular values of $\frac{\partial \hat{x}^{i+1}}{\partial\hat{x}^i}$. This gives us the squared operator norm for a specific $x$, and we take the average over all $x, y \sim \mathcal{T}$. We similarly calculate the alignment ratio of $\Delta^{i+1}$ and the layerwise Jacobian $\frac{\partial \hat{x}^{i+1}}{\partial \hat{x}^i}$ by multiplying them together, calculating their self-adjoint product, calculating the top eigenvalue, and averaging over all inputs $x$.

\subsection{Exponential Integration}
\label{sec:exponential-integration}

Due to the high sharpness of the Hessian matrix $\mathcal{H}_\theta\widetilde{\mathcal{L}}(\theta)$, training neural networks using gradient descent does not accurately approximate the underlying system of ordinary differential equations. To achieve a more accurate approximation, we calculate the top eigenvalues $\lambda_m(\theta)$ and eigenvectors $v_m(\theta)$ of the Hessian matrix at every iteration as described in Section~\ref{sec:power-iteration}. We can then decompose the gradient into the sum of its projections onto those eigenvectors and an orthogonal residual:
$$
\frac{d\theta_t}{dt} = -\sum_{m=1}^{k} c_m(\theta_t) v_m(\theta_t) - w(\theta_t)
$$
where $c_m(\theta_t) = \nabla_\theta\widetilde{\mathcal{L}}(\theta_t)^Tv_m(\theta_t)$, and $w(\theta_t)$ is the residual. Exploiting the Taylor decomposition, we can expand this to:
$$
\frac{d\theta_t}{dt} = -\sum_{m=1}^{k} \bigg(
c_m(\theta_0) v_m(\theta_0) + \lambda_m(\theta_0)(\theta_t - \theta_0)^T v_m(\theta_0)
\bigg) - w(\theta_0) + \mathcal{O}(||\theta_t - \theta_0||)
$$
Define $r_m(t) = (\theta_t - \theta_0)^Tv_m(\theta_0)$ to be the coordinate of $\theta_t$ in the portion of the eigenbasis we have retrieved. Then:
$$
\frac{dr_m}{dt} \approx -c_m(\theta_0) - \lambda_m(\theta_0)r_m(t)
$$
which has the solution:
$$
r_m(t) \approx \frac{c_m(\theta_0)}{\lambda_m(\theta_0)}\left(e^{-\lambda_m(\theta_0) t} - 1\right)
$$
Early in training, before the top eigenvalue has risen significantly, $r_m(t)$ may be unreasonably large. We therefore instead use in practice:
$$
	r_m(t) = \min\left(
		\frac{1}{\lambda_m(\theta_t)}\left(e^{-\lambda_m(\theta_0) t} - 1\right),
		\eta
	\right) \cdot c_m(\theta_0)
$$
This gives us:
$$
\theta_t \approx
\sum_{m=1}^k r_m(0) v_m(\theta_0) - w(\theta_0)t + \theta_0
$$
Since we are calculating multiple eigenvalues and eigenvectors anyway, we calculate one additional $(k + 1)$th eigenvalue and use it to determine the step size:
$$
\Delta t = \min\left(\frac{1}{\lambda_{k+1}(\theta)}, \eta\right)
$$
where $\eta$ is a learning rate. This is effectively the exponential Euler method \citep{exponential-integration-origin,exponential-integration-survey}, but using only a small number of top eigenvectors instead of the full eigendecomposition. In practice, we find that for most criterion functions, high sharpness occurs along only $O$ top eigenvectors, where $O$ is equal to the number of network outputs, and so use $k = O$.

\subsection{Synthetic Data}
\label{sec:dice}

\begin{figure}
	\begin{center}
		\includegraphics[width=14cm]{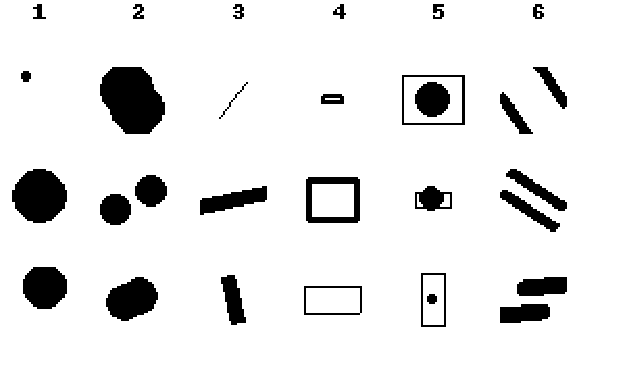}
		\caption{\label{fig:example-dice}Examples of synthetic imagery. Each column contains examples from a different class.}
	\end{center}

\end{figure}

In order to study the connection between generalization and sharpness, we needed a dataset on which we can achieve arbitrarily low test loss with a computationally tractable training dataset. We therefore created a simple synthetic dataset with six classes consisting of abstract symbols inspired by dice. Examples are shown in Figure~\ref{fig:example-dice}. We randomly generated 20,000 examples of each symbol for each class, or 120,000 in total, and split them evenly into train and test sets of 60,000 examples each. We used a consistent train-test split across all experiments.

The basic idea was to create a collection of images using the pips on a standard die as the defining relationship for classication, yielding six distinct classes. The placement of the graphical objects and defining shapes were randomized during creation, to create a varied set of images within each class. Each output image was set at a fixed size, with the horizontal and vertical axis scaled from -1 to 1.

Images for the ``one'' class were a single point randomly placed in the image field using the ``points'' function of the R base plot package \citep{RSoftware}. The point size, as determined by the cex parameter in the code, and $x$ and $y$ coordinates for placement were randomly selected as parameters $P_{size}$, $x_{coord}$, and $y_{coord}$, respectively.

Images for the ``twos'' class were created using the same parameters for the ``ones'' class and then reflecting the resulting coordinates across the x-y axis. The first pip is generated from randomly selected parameters $P_{size}$, $x_{coord}$, and $y_{coord}$ and the second pip generated using the same values to generate a corresponding $P_{size}$, $-x_{coord}$, and $-y_{coord}$.

Images for ``threes'' were created by generating a single line (in the spirit of three connected dots) using the ``lines'' function in the base R \citep{RSoftware} package . The starting point of the line was determined in the same method as a ``one'' point by randomly generateing two coordainates, $x_{coord}$ and $y_{coord}$. The end point of the line was the reflection of those two coodinates across the $x$ and $y$ axis as $-x_{coord}$ and $-y_{coord}$, akin to how the second point of the ``twos'' class was generated. The line weight is generated by a random parameter $L_{weight}$ and the end type, $E_{type}$, is selected randomly from the three types available (round, butt, and square).

Images for the ``fours'' class were created by using the ``polygon'' function in the base R \citep{RSoftware} package. An $x$ and $y$ coordinate were generated using the process used for ``ones'' with the other three points generated by reflecting values across the $x$/$y$ axis. The four points of the box are therefore defined as $x_{coord}$ and $y_{coord}$, $-x_{coord}$ and $y_{coord}$, $x_{coord}$ and $-y_{coord}$, and $-x_{coord}$ and $-y_{coord}$. As in generating ``threes,'' the box also has parameters for line weight and end types, $L_{weight}$ and $E_{type}$.

Images for the ``fives'' class were created by generating a box in the same manner as for the ``fours'' class and then generating a single point at (0,0) using the variable size parameter, $P_{size}$, used for the ``ones'' class.

Images for the ``sixes'' class used the same method as the ``threes'' class to define an initial line with an $x_{coord}$, $y_{coord}$, $L_{weight}$, and $E_{type}$. Two additional parameters were included prior to drawing to define $x$ and $y$ translations, namely $x_{shift}$ and $y_{shift}$. The first line was then defined by the two points $x_{coord} - x_{shift}$, $y_{coord} - y_{shift}$ and $-x_{coord} - x_{shift}$, $-y_{coord} - y_{shift}$. The second line is defined by the two points $x_{coord} + x_{shift}$, $y_{coord} + y_{shift}$ and $-x_{coord} + x_{shift}$, $-y_{coord} + y_{shift}$, thus creating two parallel lines, equally offset from the origin.

In generating the images used in the experiments, the output image size was fixed at $32\times32$ pixels. 20,000 images of each class were created with an index included in the file name to denote which instance it was. The parameters were adjusted to accommodate the image size when appropriate, such as when the point sizes of the pips had to be smaller than $32\times32$. The randomly selected numbers were sampled from a uniform distribution from ranges detailed in Table \ref{table:dicegenparameters} using the ``runif'' function in the R \citep{RSoftware} package. The $E_{type}$, being the sole integer parameter, was selected using the ``sample'' package.

\begin{table}
\centering
\begin{tabular}{l c c} 

\hline

\\Parameter & Min & Max\\ [0.5ex] 

\hline

\\ \textbf{Ones}
\\$x_{coord}$
&-1
&1

\\$y_{coord}$
&-1
&1

\\$P_{size}$
&0.1
&1

\\
\\ \textbf{Twos}
\\$x_{coord}$
&-1
&1

\\$y_{coord}$
&-1
&1

\\$P_{size}$
&0.1
&1

\\
\\ \textbf{Threes}
\\$x_{coord}$
&-1
&1

\\$y_{coord}$
&-1
&1

\\$L_{weight}$
&0.1
&2.5

\\$E_{type}$
&0
&2

\\
\\ \textbf{Fours}
\\$x_{coord}$
&-1
&1

\\$y_{coord}$
&-1
&1

\\$L_{weight}$
&0.1
&1

\\$E_{type}$
&0
&2

\\
\\ \textbf{Fives}
\\$x_{coord}$
&0.2
&1

\\$y_{coord}$
&0.2
&1

\\$P_{size}$
&0.1
&0.7

\\$L_{weight}$
&0.1
&1

\\$E_{type}$
&0
&2

\\
\\ \textbf{Sixes}
\\$x_{coord}$
&-0.9
&0.9

\\$y_{coord}$
&-0.9
&0.9

\\$x_{shift}$
&0.1
&0.8

\\$y_{shift}$
&0.1
&0.8

\\$L_{weight}$
&0.1
&2.5

\\$E_{type}$
&0
&2

\end{tabular}
\caption{\textbf{Image generation parameters by class}}
\label{table:dicegenparameters}
\end{table}

\subsection{Additional Proofs}
\label{sec:random-matrix}

\begin{lemma}\label{lemma:mmt}Let $M$ be a matrix, and let $\lambda, v$ be a nonzero eigenvalue, eigenvector of $M^TM$. Then $\lambda, Mv / ||Mv||$ are a nonzero eigenvalue, eigenvector of $MM^T$.\end{lemma}

\proof{
	The result is immediate:
$$
MM^T \left(\frac{Mv}{||Mv||}\right) = \frac{M}{||Mv||} (M^TM v) = \lambda\left(\frac{Mv}{||Mv||}\right)
$$
}

Let $M$ be a random symmetric matrix. We define the \textbf{overlap ratio} of $M$ to be:
$$
\rho(M) = \frac{\lambda^1(\mathbb{E} (M^TM))}{\mathbb{E} \lambda^1(M^TM)}
$$
where $\lambda^1$ denotes the top eigenvalue of a symmetric matrix.

\begin{lemma}\label{lemma:overlap-ratio} For any random matrix $M$, $0 \leq \rho(M) \leq 1$.\end{lemma}

	\proof{
		$\lambda^{1}$ is equivalent to the operator norm, and so is a convex function due to the triangle inequality. Therefore, by Jensen's inequality, $\lambda^{1}(\mathbb{E} (M^TM)) \leq \mathbb{E} \lambda^{\max}(M^TM)$. Further, let $v$ be the eigenvector of $\mathbb{E} (M^TM)$ corresponding to $\lambda^{1}(\mathbb{E} (M^TM))$. We may assume without loss of generality that $||v|| = 1$. Then:
	$$
	\lambda^{1}(\mathbb{E}(M^TM)) = \lambda^{1}(\mathbb{E}(M^TM)) v^Tv = v^T(\mathbb{E} (M^TM))v = \mathbb{E} (v^TM^TMv) = \mathbb{E} ||Mv||^2 \geq 0
	$$
Therefore:
$$
	0 \leq \lambda^{1}(\mathbb{E} (M^TM)) \leq \mathbb{E} \lambda^{1}(M^TM)
$$
Dividing by $\mathbb{E}\lambda^{1}(M^TM)$ gives us:
$$
	0 \leq \rho(M) \leq 1
$$
}

\newpage

\subsection{Additional Figures}

\subsubsection{Components of the Hessian Matrix}

\subsubsection{Overlap Ratio $\rho(K)$}

\newpage

\subsubsection{Layerwise Sharpness}

\newpage

\subsubsection{Components of $\mathbb{E} ||\Delta^1||^2_{\max}$}

\begin{figure}
	\begin{center}
		\includegraphics[width=\textwidth]{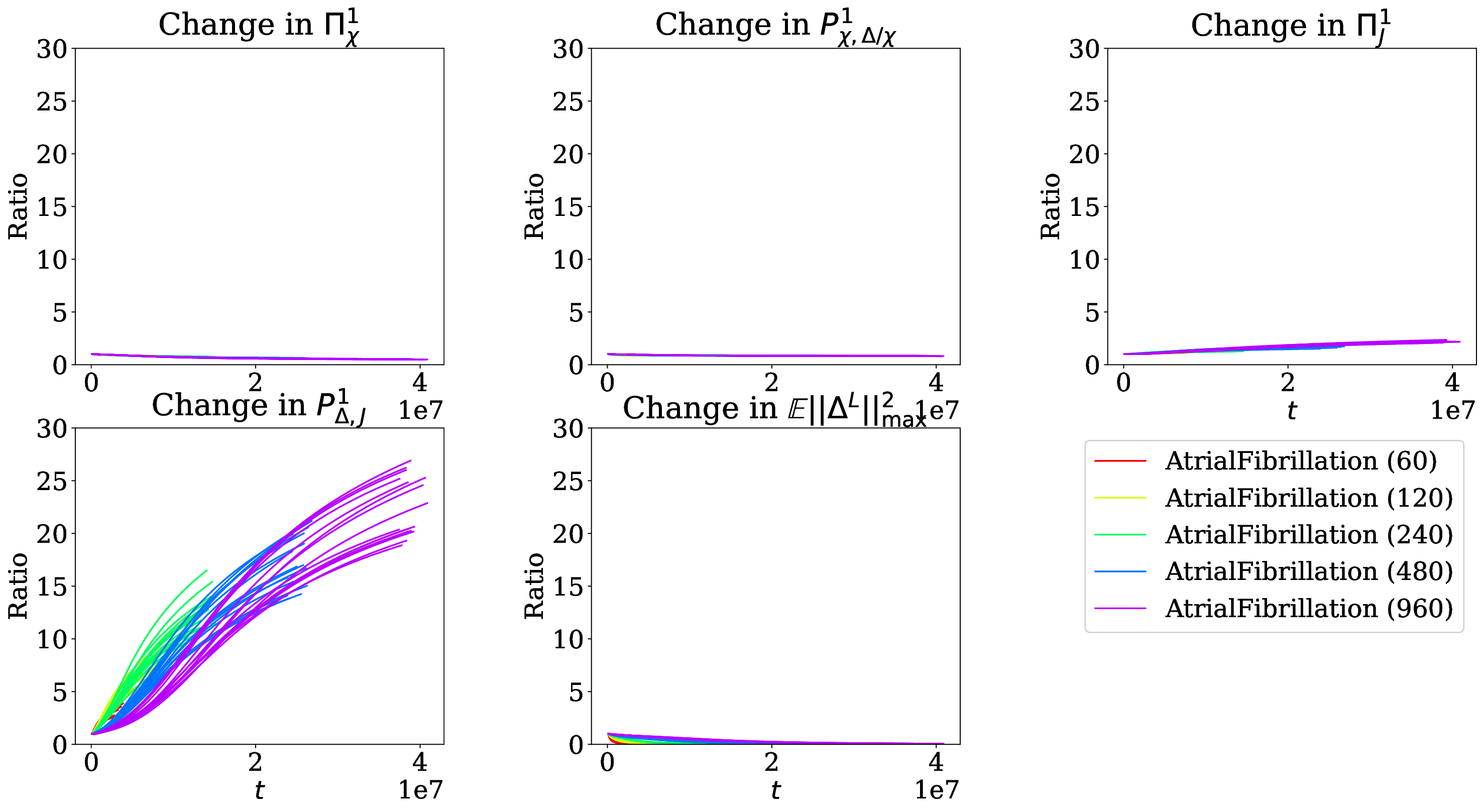}
		\caption{\label{fig:mlp-ucr-cross-entropy-main}Change in components of $\mathbb{E}||\Delta^1||^2_{\max}$ when trained on AtrialFibrillation with cross-entropy: $\Pi_\chi^1$ (top left), $\Rho_{\chi,\Delta/\chi}^1$ (top middle), $\Pi_J^1$ (top right), $\Rho_{\Delta,J}^1$ (bottom left), and $\mathbb{E}||\Delta^L||^2_{\max}$ (bottom middle).}
	\end{center}
\end{figure}

\begin{figure}
	\begin{center}
		\includegraphics[width=\textwidth]{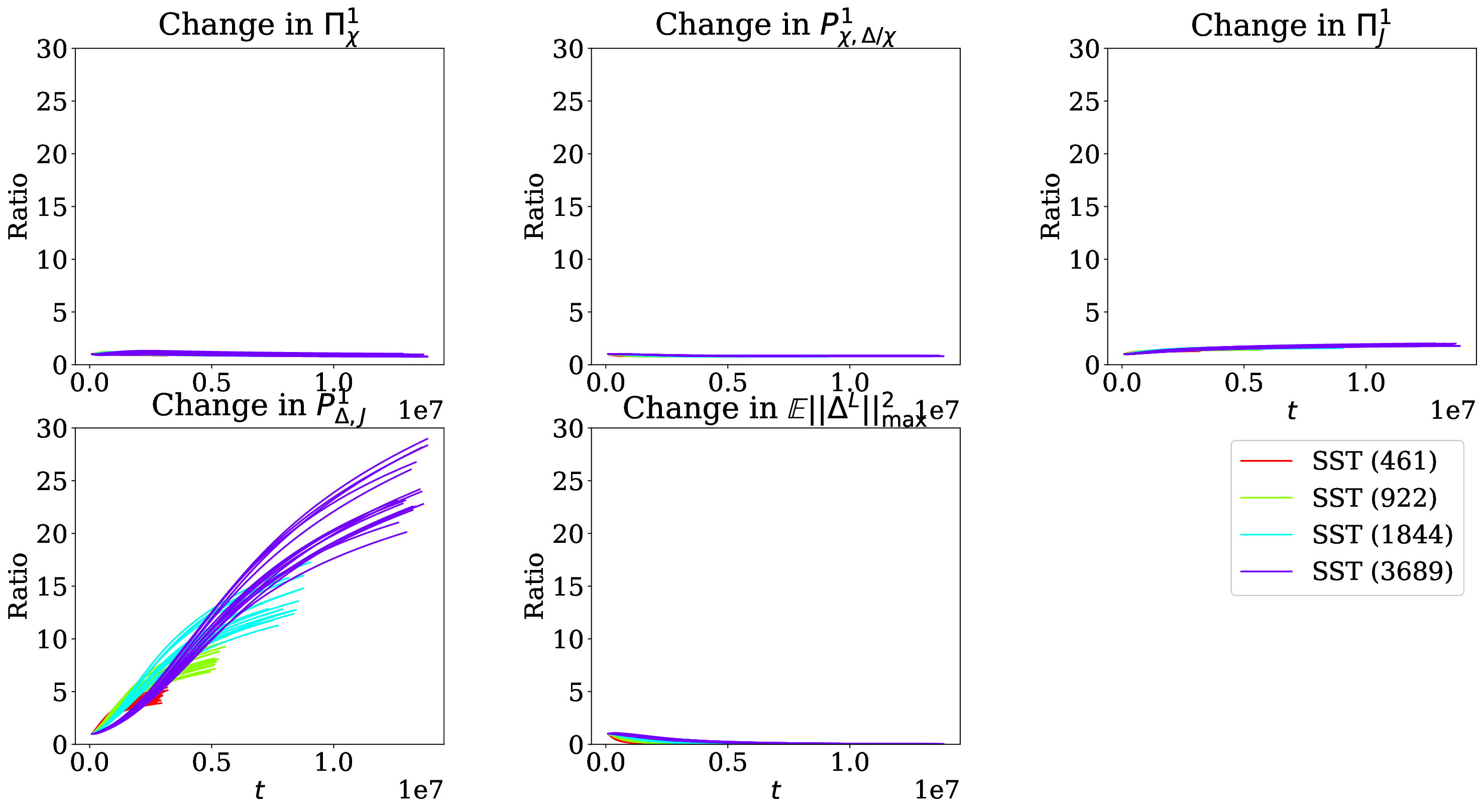}
		\caption{\label{fig:mlp-sst-cross-entropy-main}Change in components of $\mathbb{E}||\Delta^1||^2_{\max}$ when trained on SST with cross-entropy: $\Pi_\chi^1$ (top left), $\Rho_{\chi,\Delta/\chi}^1$ (top middle), $\Pi_J^1$ (top right), $\Rho_{\Delta,J}^1$ (bottom left), and $\mathbb{E}||\Delta^L||^2_{\max}$ (bottom middle).}
	\end{center}
\end{figure}

\newpage

\subsubsection{Activation Ratios}

\begin{figure}
	\begin{center}
		\includegraphics[width=\textwidth]{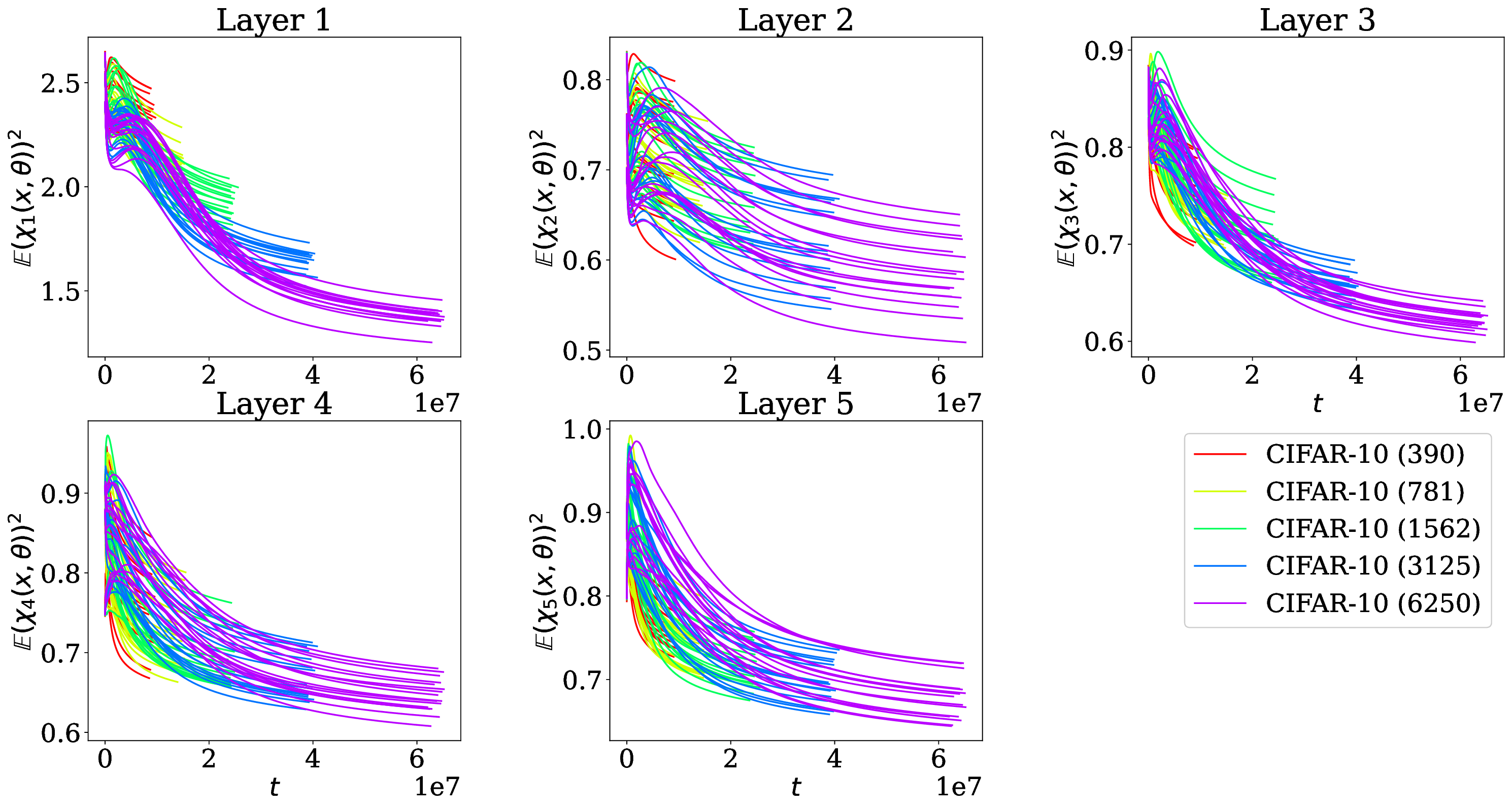}
		\caption{\label{fig:mlp-cifar10-cross-entropy-activation-ratios}Mean squared activation ratios $\mathbb{E} (\chi^i(x, \theta))^2$ when trained on CIFAR-10 using cross-entropy for layers $i = 1$ (top left), 2 (top middle), 3 (top right), 4 (bottom left), and 5 (bottom middle)}
	\end{center}
\end{figure}

\begin{figure}
	\begin{center}
		\includegraphics[width=\textwidth]{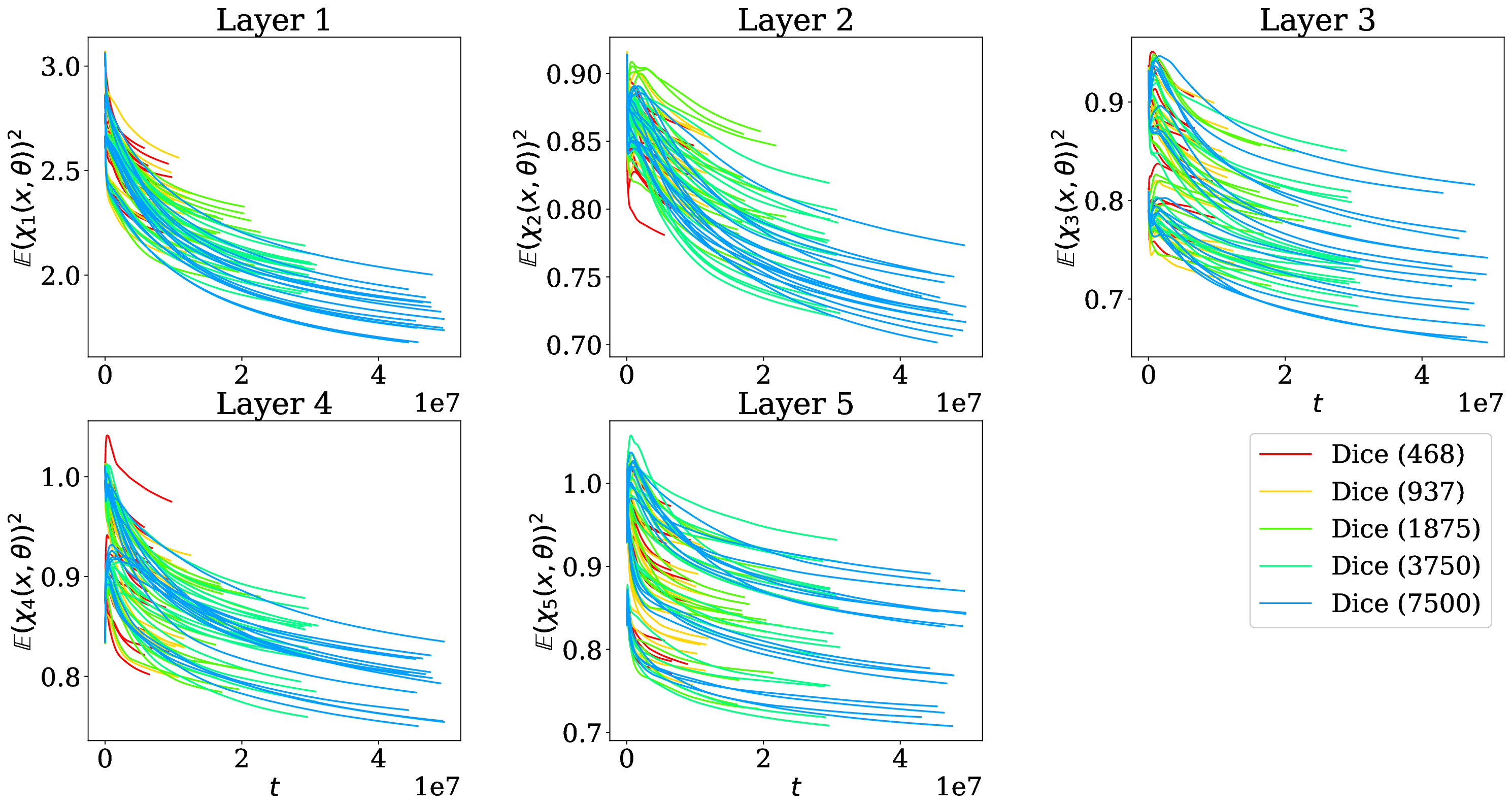}
		\caption{\label{fig:mlp-dice-cross-entropy-activation-ratios}Mean squared activation ratios $\mathbb{e} (\chi^i(x, \theta))^2$ when trained on dice dataset using cross-entropy for layers $i = 1$ (top left), 2 (top middle), 3 (top right), 4 (bottom left), and 5 (bottom middle)}
	\end{center}
\end{figure}

\begin{figure}
	\begin{center}
		\includegraphics[width=\textwidth]{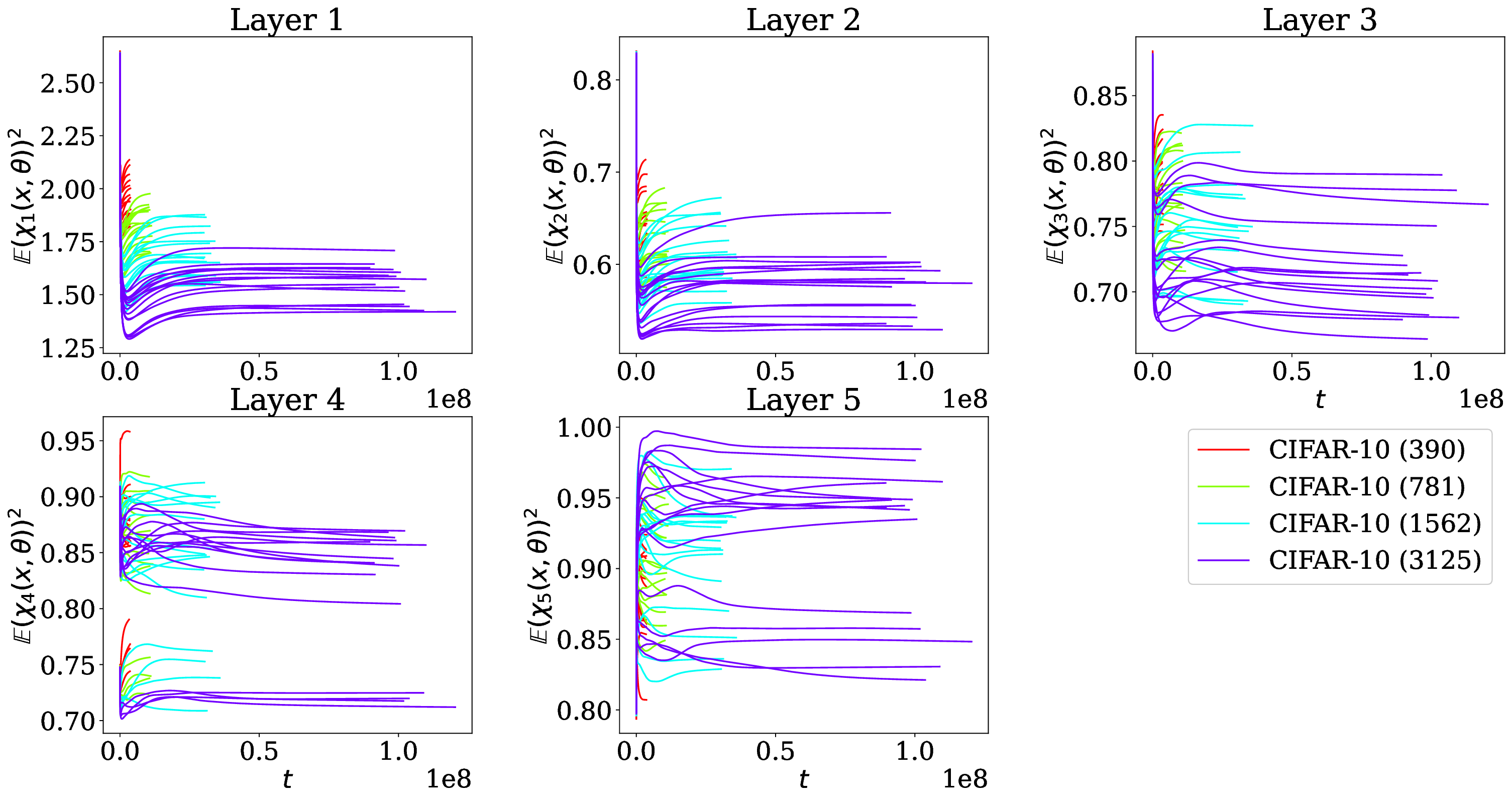}
		\caption{\label{fig:mlp-cifar10-mse-activation-ratios}Mean squared activation ratios $\mathbb{E} (\chi^i(x, \theta))^2$ when trained on CIFAR-10 using MSE for layers $i = 1$ (top left), 2 (top middle), 3 (top right), 4 (bottom left), and 5 (bottom middle)}
	\end{center}
\end{figure}
 
\begin{figure}
	\begin{center}
		\includegraphics[width=\textwidth]{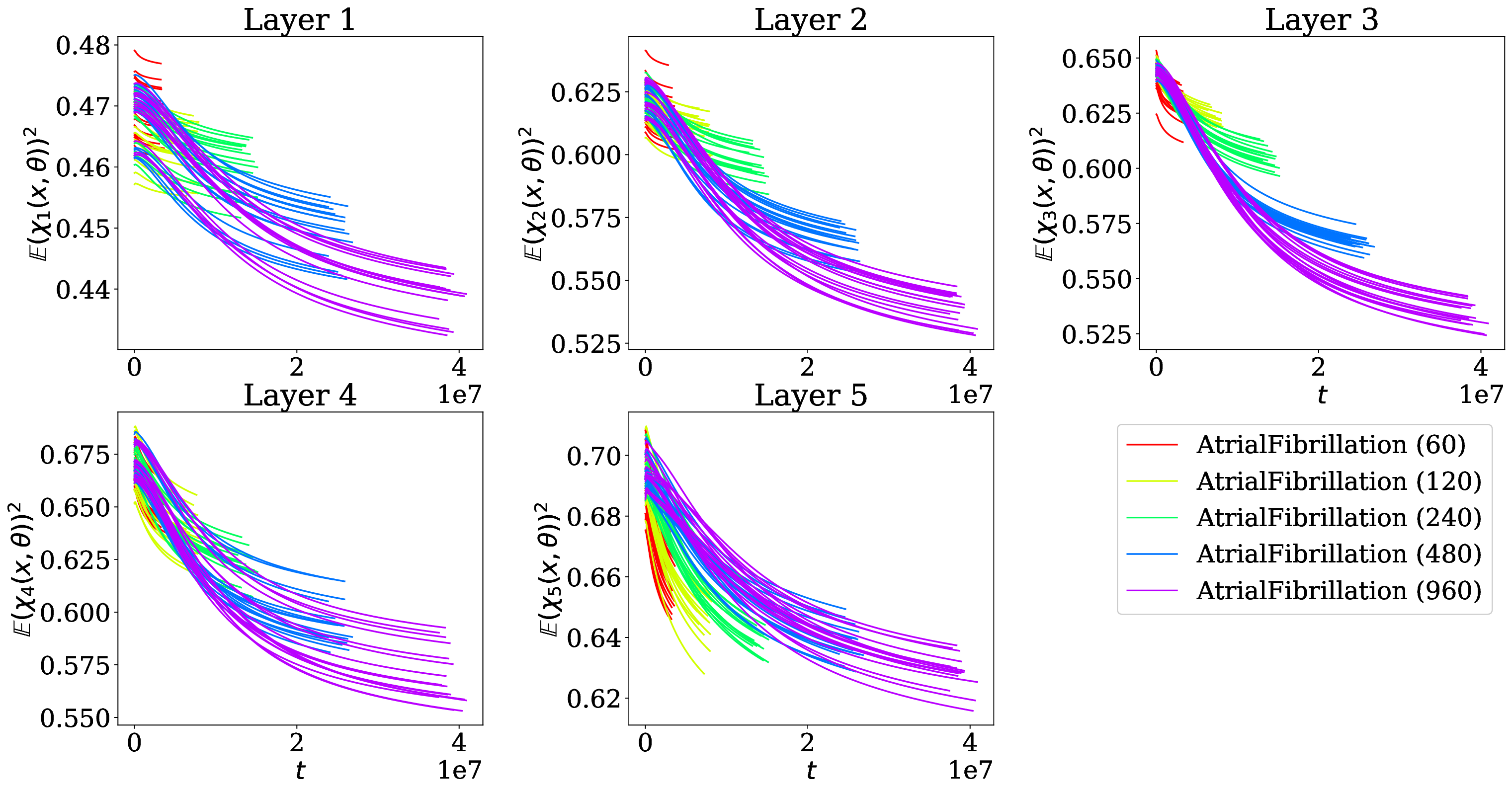}
		\caption{\label{fig:mlp-ucr-cross-entropy-activation-ratios}Mean squared activation ratios $\mathbb{e} (\chi^i(x, \theta))^2$ when trained on AtrialFibrillation using cross-entropy for layers $i = 1$ (top left), 2 (top middle), 3 (top right), 4 (bottom left), and 5 (bottom middle)}
	\end{center}
\end{figure}

\begin{figure}
	\begin{center}
		\includegraphics[width=\textwidth]{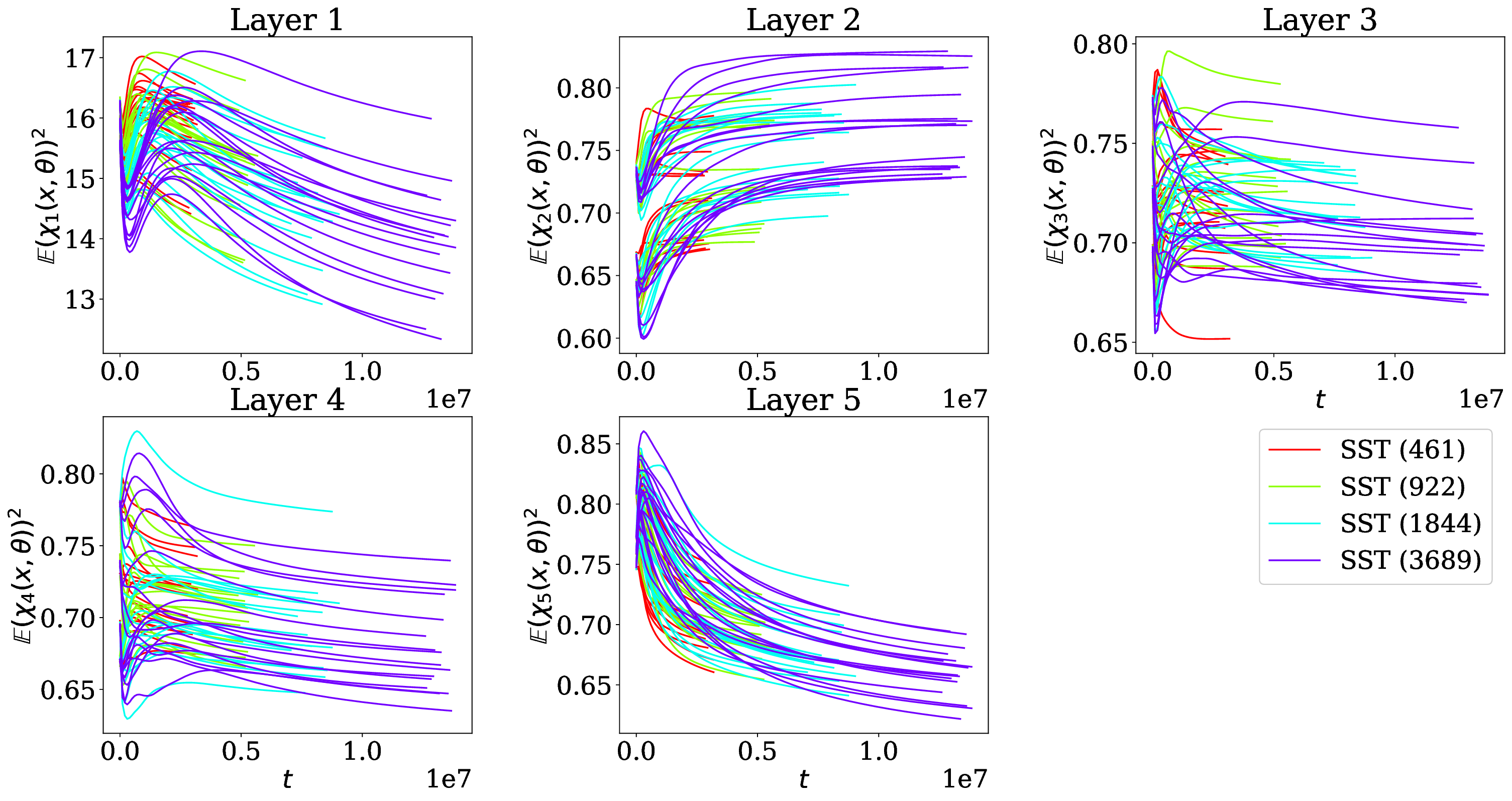}
		\caption{\label{fig:mlp-sst-cross-entropy-activation-ratios}Mean squared activation ratios $\mathbb{e} (\chi^i(x, \theta))^2$ when trained on SST using cross-entropy for layers $i = 1$ (top left), 2 (top middle), 3 (top right), 4 (bottom left), and 5 (bottom middle)}
	\end{center}
\end{figure}

\newpage

\subsubsection{Alignment Ratios $r(\chi^i, \Delta^i / \chi^i)$}

\begin{figure}
	\begin{center}
		\includegraphics[width=\textwidth]{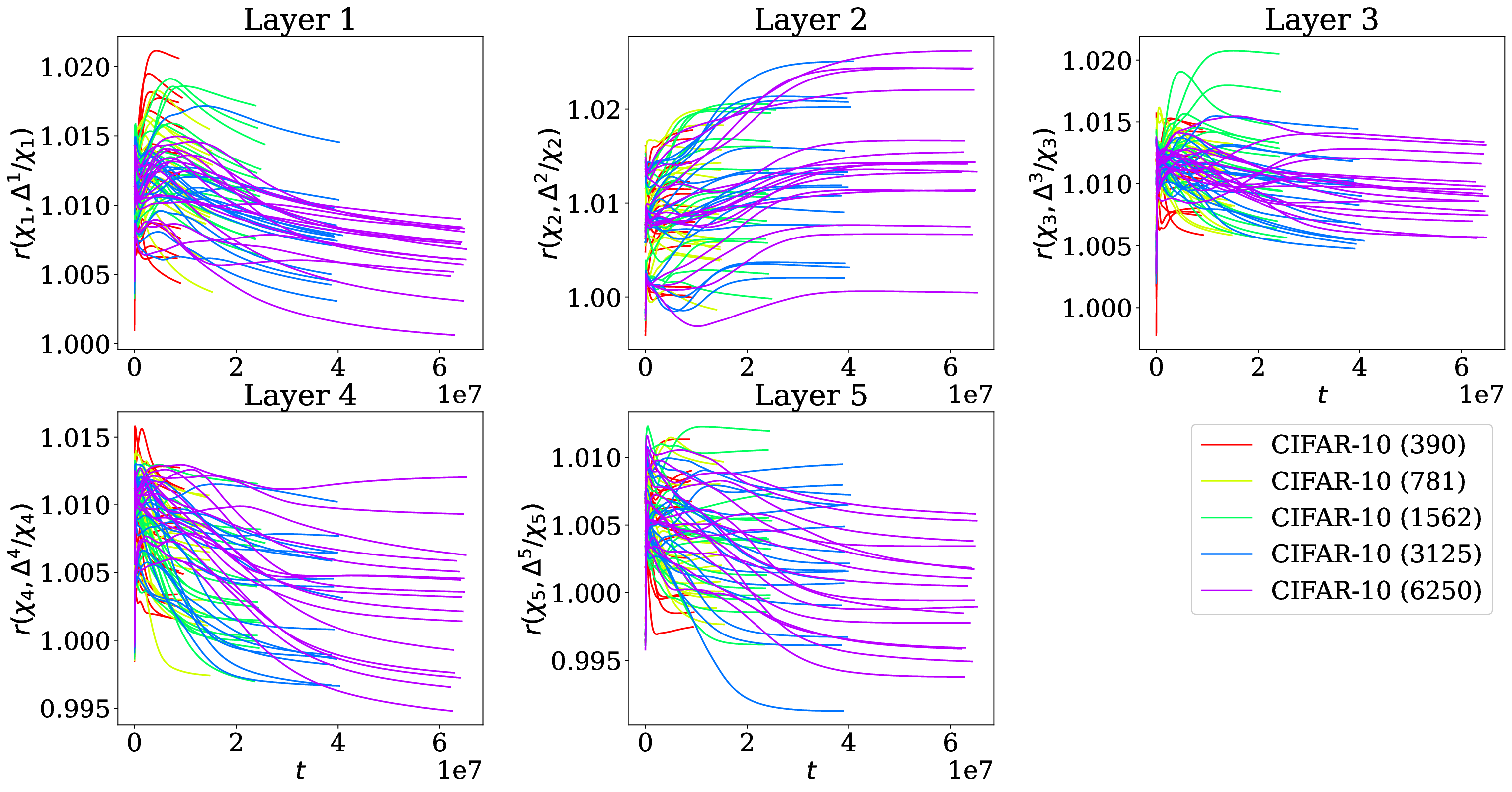}
		\caption{\label{fig:mlp-cifar10-cross-entropy-chi-delta-alignment-ratios}Alignment ratios $r(\chi^i, \Delta^i / \chi^i)$ when trained on CIFAR-10 using cross-entropy for layers $i = 1$ (top left), 2 (top middle), 3 (top right), 4 (bottom left), and 5 (bottom middle)}
	\end{center}
\end{figure}

\begin{figure}
	\begin{center}
		\includegraphics[width=\textwidth]{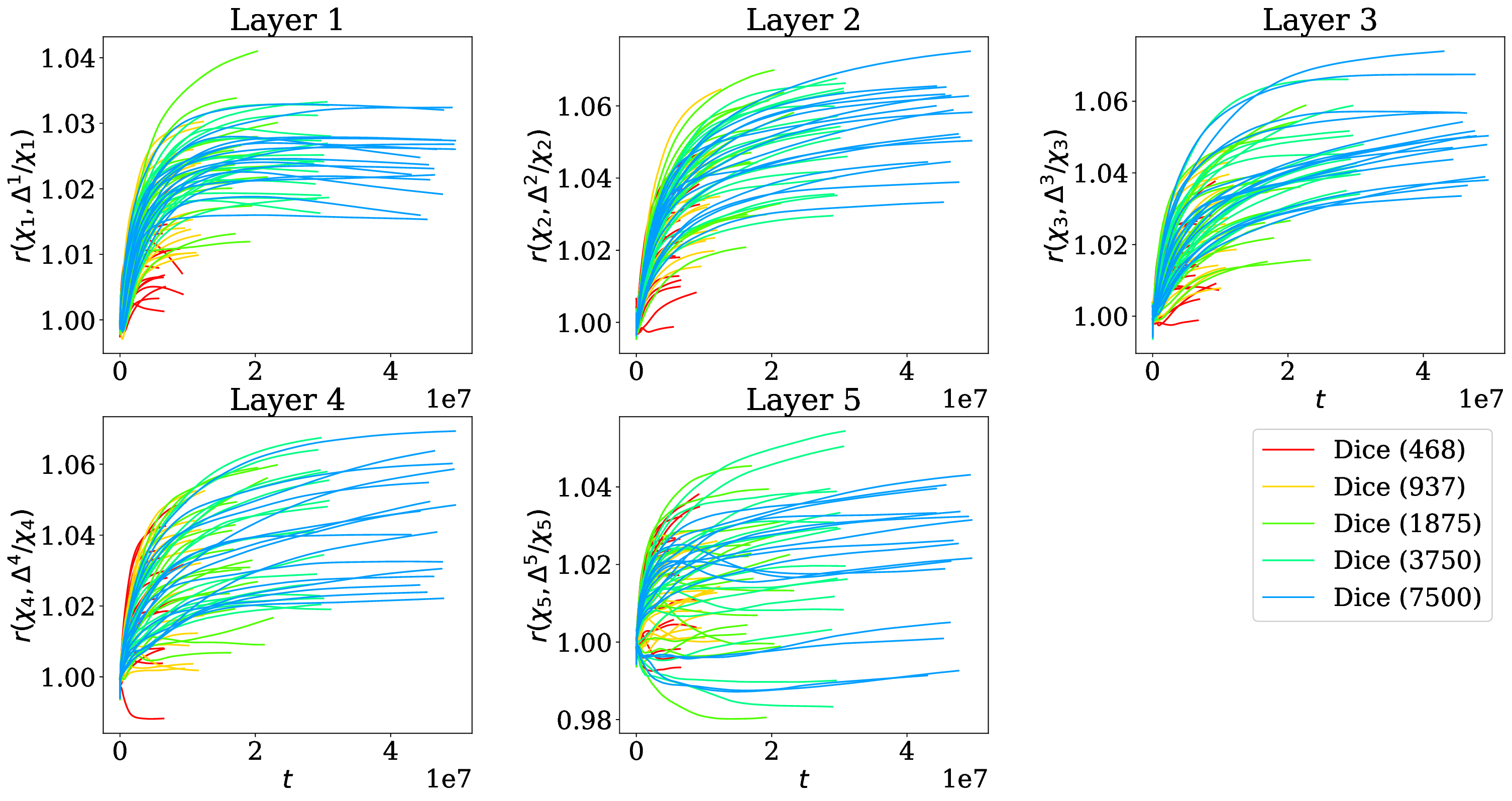}
		\caption{\label{fig:mlp-dice-cross-entropy-chi-delta-alignment-ratios}Alignment ratios $r(\chi^i, \Delta^i / \chi^i)$ when trained on dice dataset using cross-entropy for layers $i = 1$ (top left), 2 (top middle), 3 (top right), 4 (bottom left), and 5 (bottom middle)}
	\end{center}
\end{figure}

\begin{figure}
	\begin{center}
		\includegraphics[width=\textwidth]{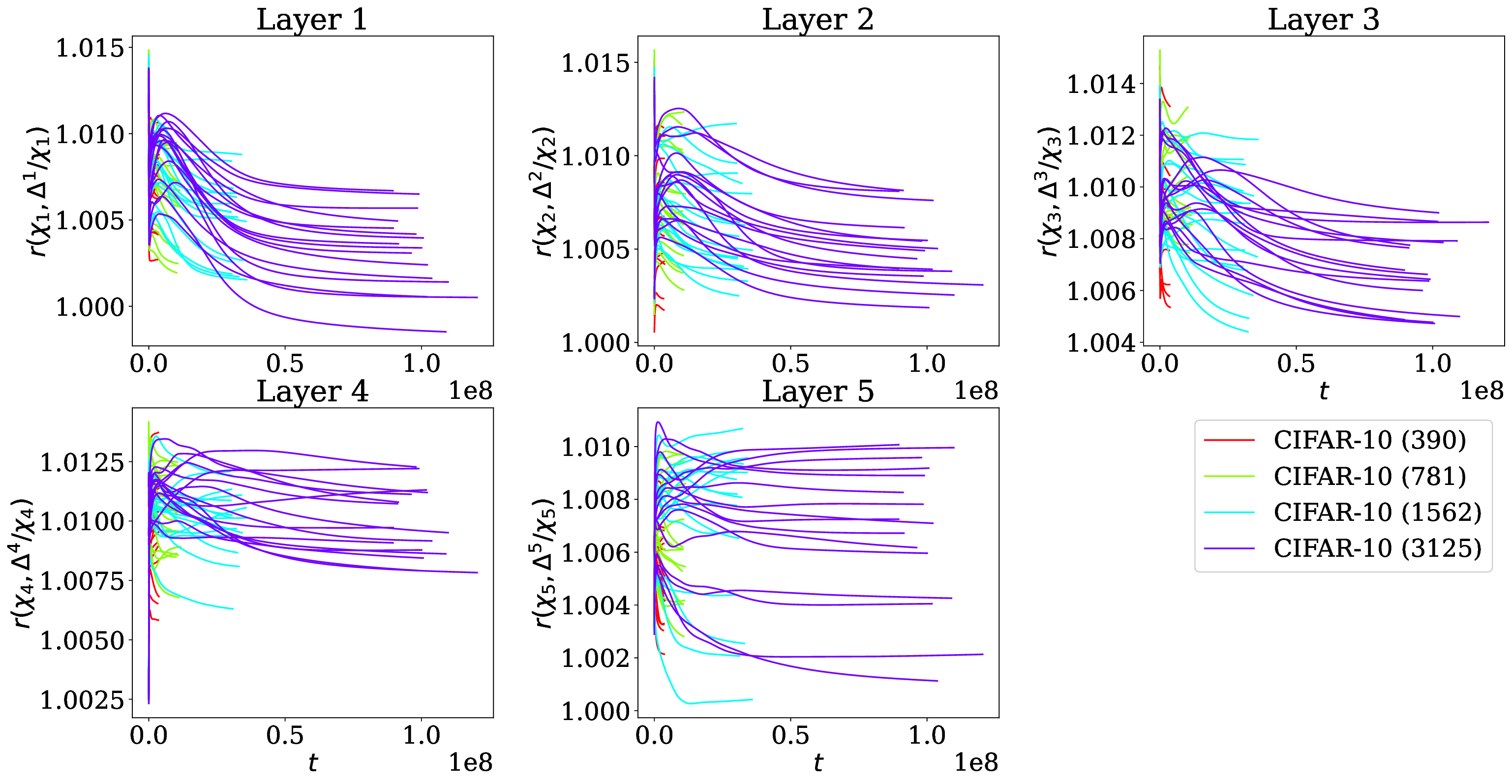}
		\caption{\label{fig:mlp-cifar10-mse-chi-delta-alignment-ratios}Alignment ratios $r(\chi^i, \Delta^i / \chi^i)$ when trained on CIFAR-10 using MSE for layers $i = 1$ (top left), 2 (top middle), 3 (top right), 4 (bottom left), and 5 (bottom middle)}
	\end{center}
\end{figure}

\begin{figure}
	\begin{center}
		\includegraphics[width=\textwidth]{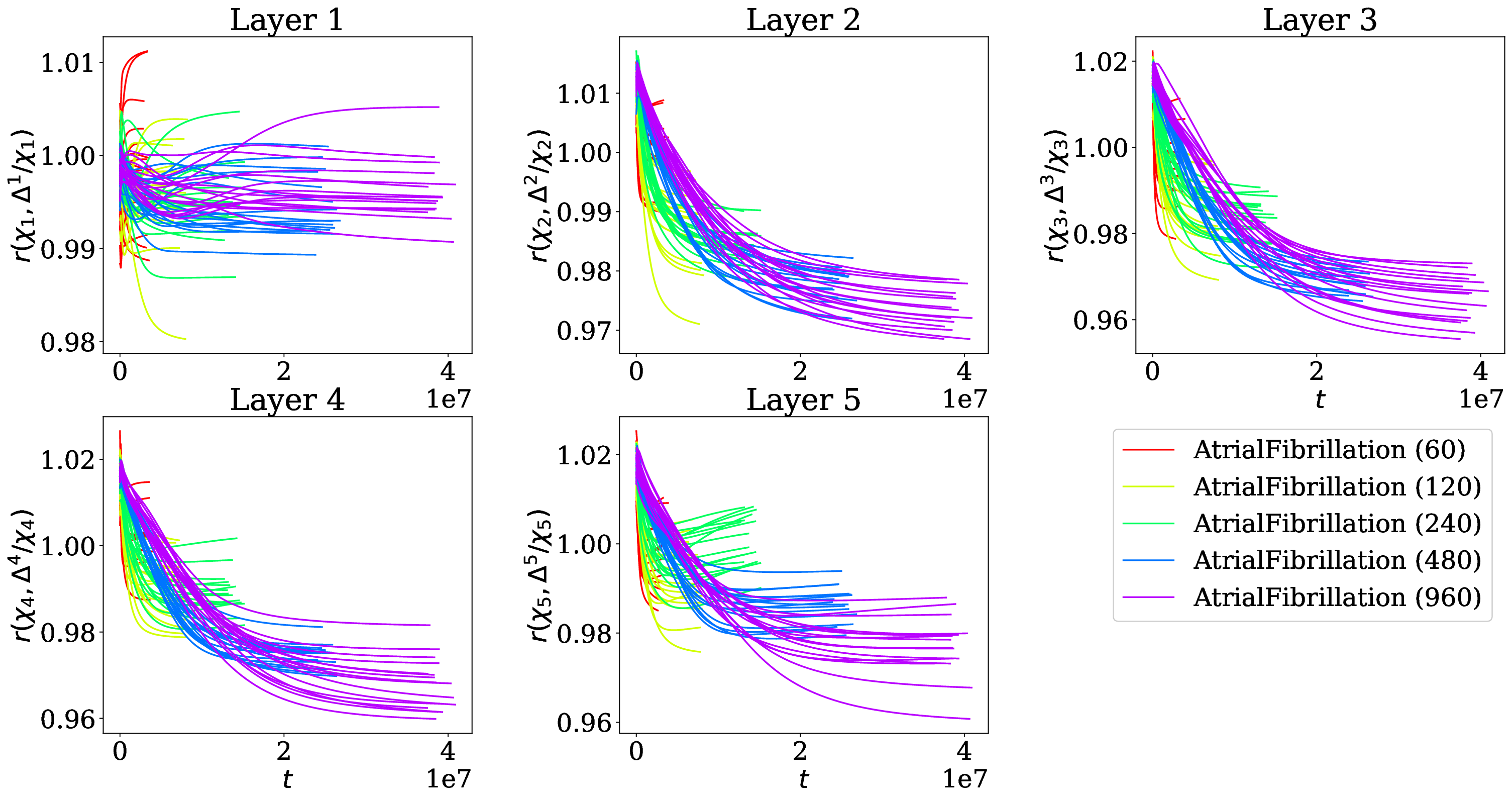}
		\caption{\label{fig:mlp-ucr-cross-entropy-chi-delta-alignment-ratios}Alignment ratios $r(\chi^i, \Delta^i / \chi^i)$ when trained on AtrialFibrillation using cross-entropy for layers $i = 1$ (top left), 2 (top middle), 3 (top right), 4 (bottom left), and 5 (bottom middle)}
	\end{center}
\end{figure}

\begin{figure}
	\begin{center}
		\includegraphics[width=\textwidth]{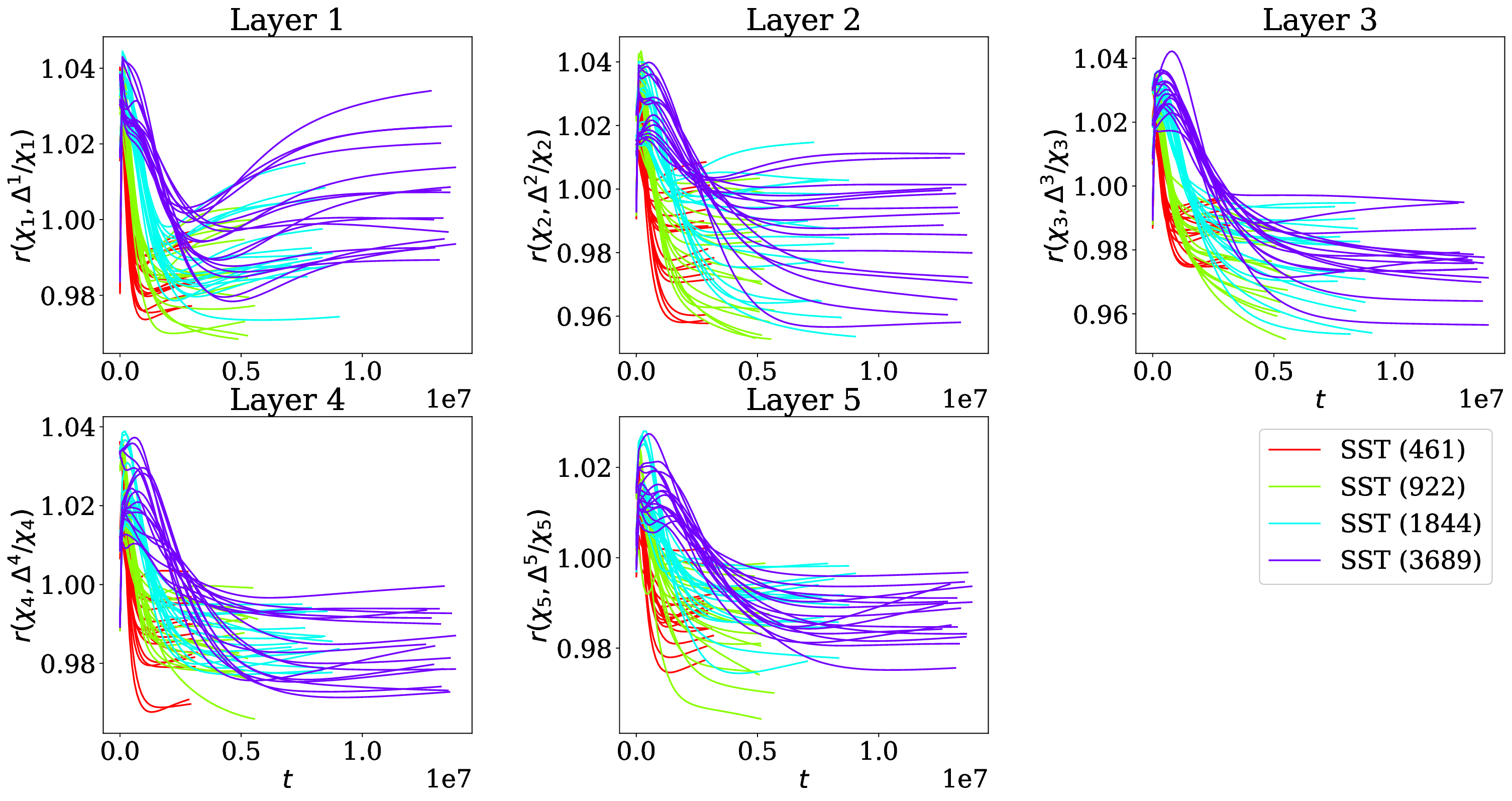}
		\caption{\label{fig:mlp-sst-cross-entropy-chi-delta-alignment-ratios}Alignment ratios $r(\chi^i, \Delta^i / \chi^i)$ when trained on SST using cross-entropy} for layers $i = 1$ (top left), 2 (top middle), 3 (top right), 4 (bottom left), and 5 (bottom middle)
	\end{center}
\end{figure}

\newpage

\subsubsection{Expected Squared Operator Norms of the Layerwise Jacobians $\mathbb{E} ||\frac{\partial \hat{x}^{i+1}}{\partial\hat{x}^i}||^2_{\max}$}

\begin{figure}
	\begin{center}
		\includegraphics[width=\textwidth]{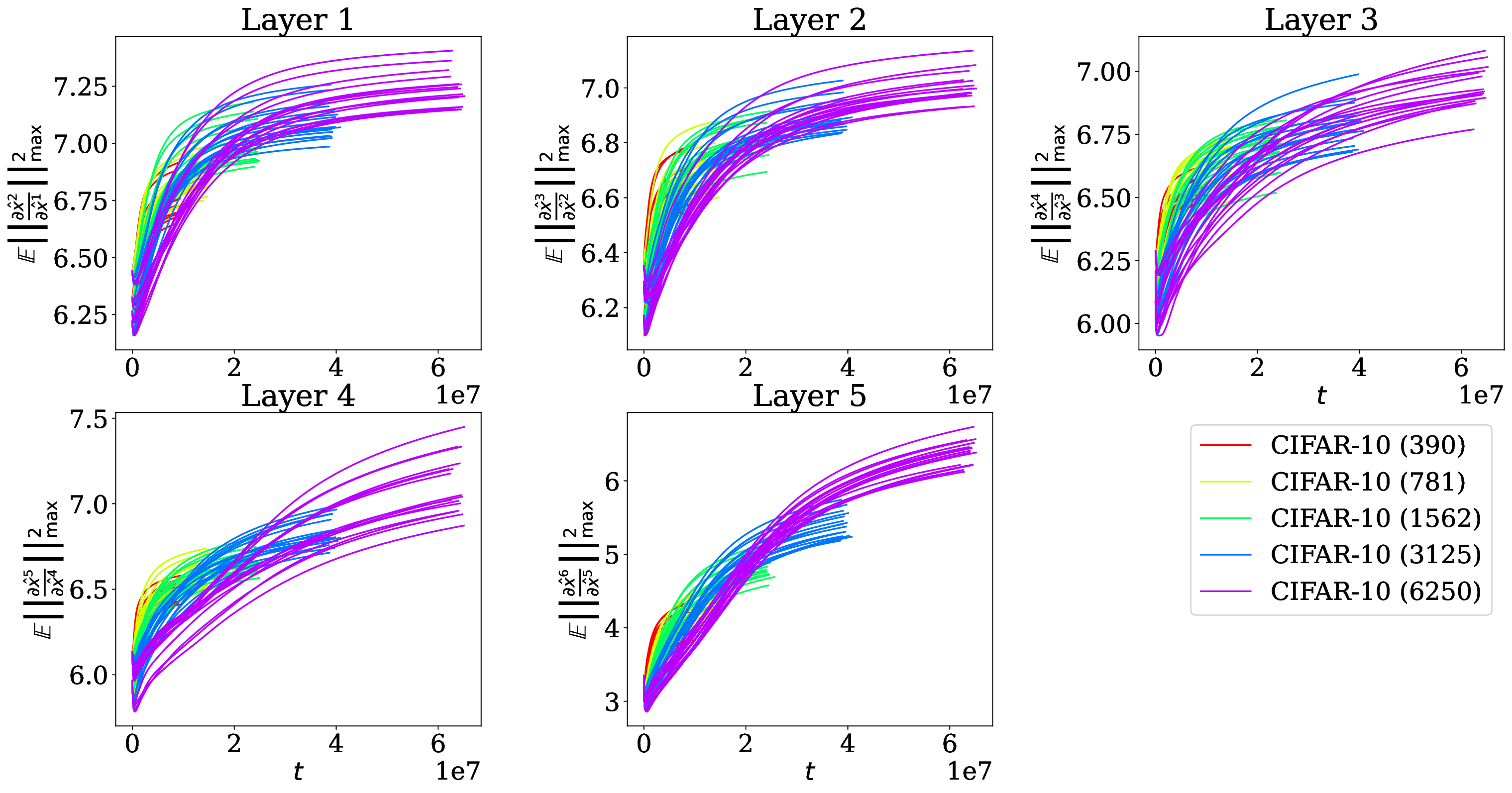}
		\caption{\label{fig:mlp-cifar10-cross-entropy-layerwise-operator-norm}Mean squared operator norms of the layerwise Jacobians $\mathbb{E} ||\frac{\partial \hat{x}^{i+1}}{\partial\hat{x}^i}||^2_{\max}$ when trained on CIFAR-10 using cross-entropy for layers $i = 1$ (top left), 2 (top middle), 3 (top right), 4 (bottom left), and 5 (bottom middle)}
	\end{center}
\end{figure}

\begin{figure}
	\begin{center}
		\includegraphics[width=\textwidth]{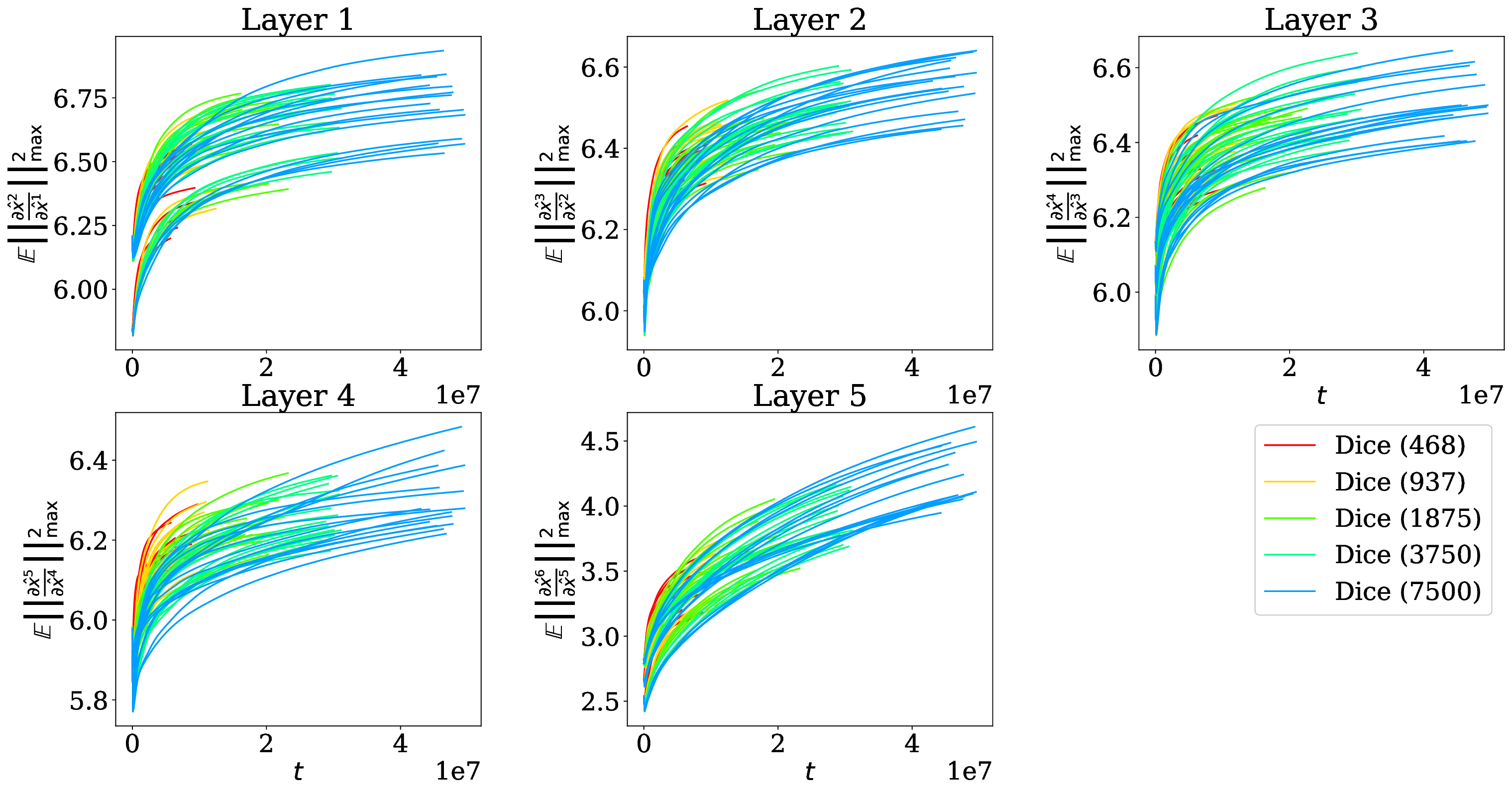}
		\caption{\label{fig:mlp-dice-cross-entropy-layerwise-operator-norm}Expected squared operator norms of the layerwise Jacobians $\mathbb{E} ||\frac{\partial \hat{x}^{i+1}}{\partial\hat{x}^i}||^2_{\max}$ when trained on dice dataset using cross-entropy for layers $i = 1$ (top left), 2 (top middle), 3 (top right), 4 (bottom left), and 5 (bottom middle)}
	\end{center}
\end{figure}

\begin{figure}
	\begin{center}
		\includegraphics[width=\textwidth]{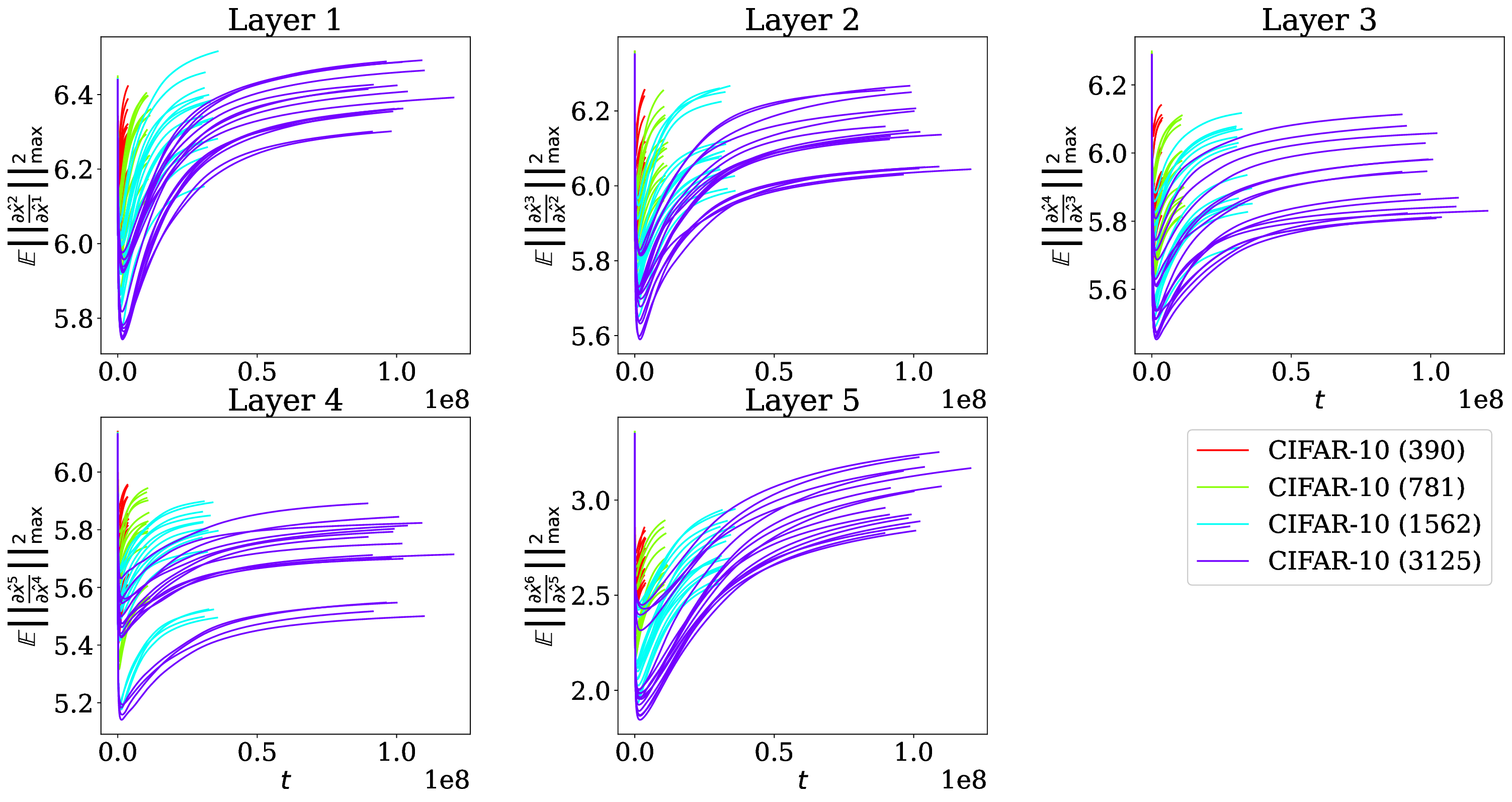}
		\caption{\label{fig:mlp-cifar10-mse-layerwise-operator-norm}Expected squared operator norms of the layerwise Jacobians $\mathbb{E} ||\frac{\partial \hat{x}^{i+1}}{\partial\hat{x}^i}||^2_{\max}$ when trained on CIFAR-10 using MSE for layers $i = 1$ (top left), 2 (top middle), 3 (top right), 4 (bottom left), and 5 (bottom middle)}
	\end{center}
\end{figure}

\begin{figure}
	\begin{center}
		\includegraphics[width=\textwidth]{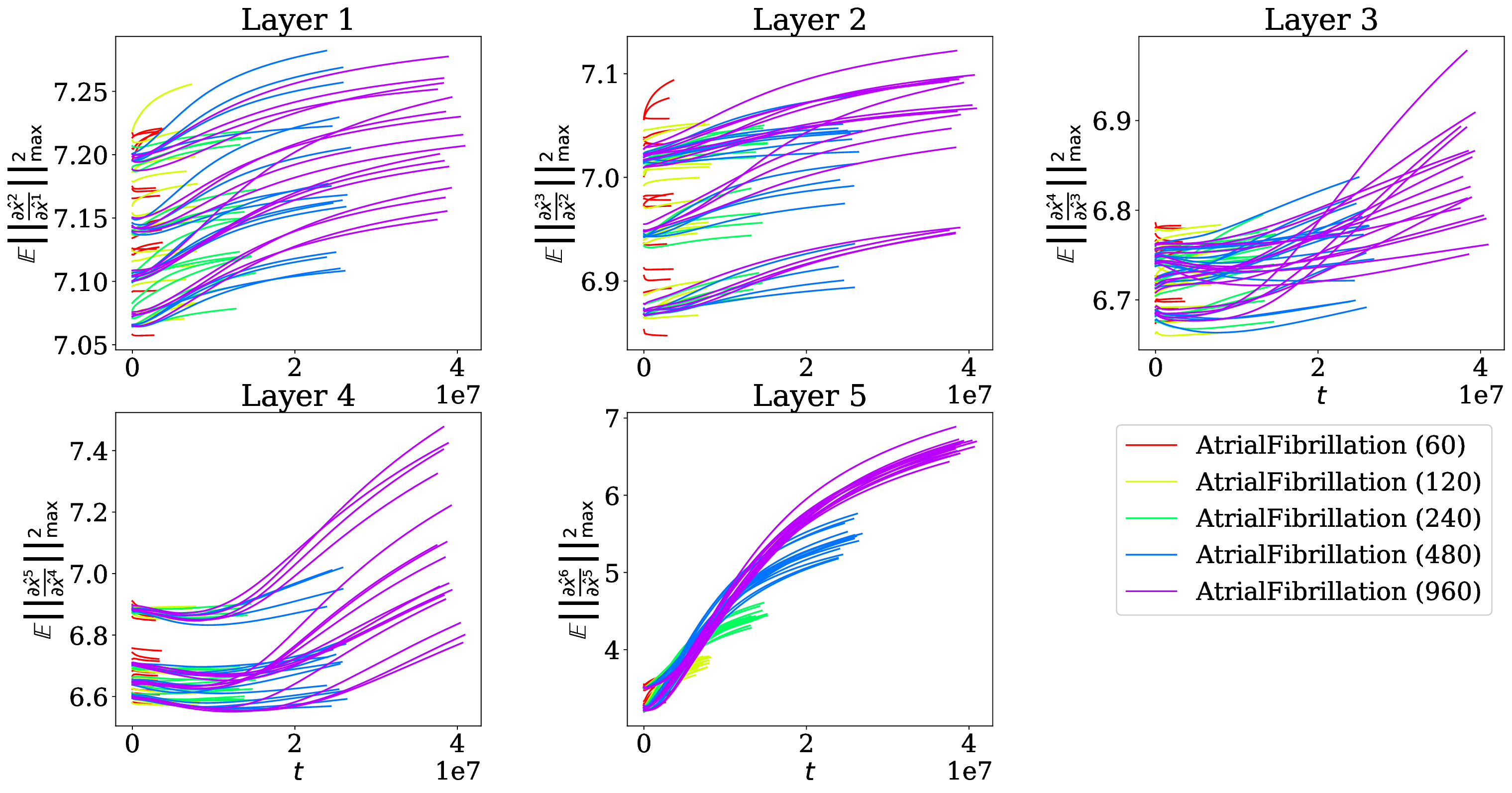}
		\caption{\label{fig:mlp-ucr-cross-entropy-layerwise-operator-norm}Expected squared operator norms of the layerwise Jacobians $\mathbb{E} ||\frac{\partial \hat{x}^{i+1}}{\partial\hat{x}^i}||^2_{\max}$ when trained on AtrialFibrillation using cross-entropy for layers $i = 1$ (top left), 2 (top middle), 3 (top right), 4 (bottom left), and 5 (bottom middle)}
	\end{center}
\end{figure}

\begin{figure}
	\begin{center}
		\includegraphics[width=\textwidth]{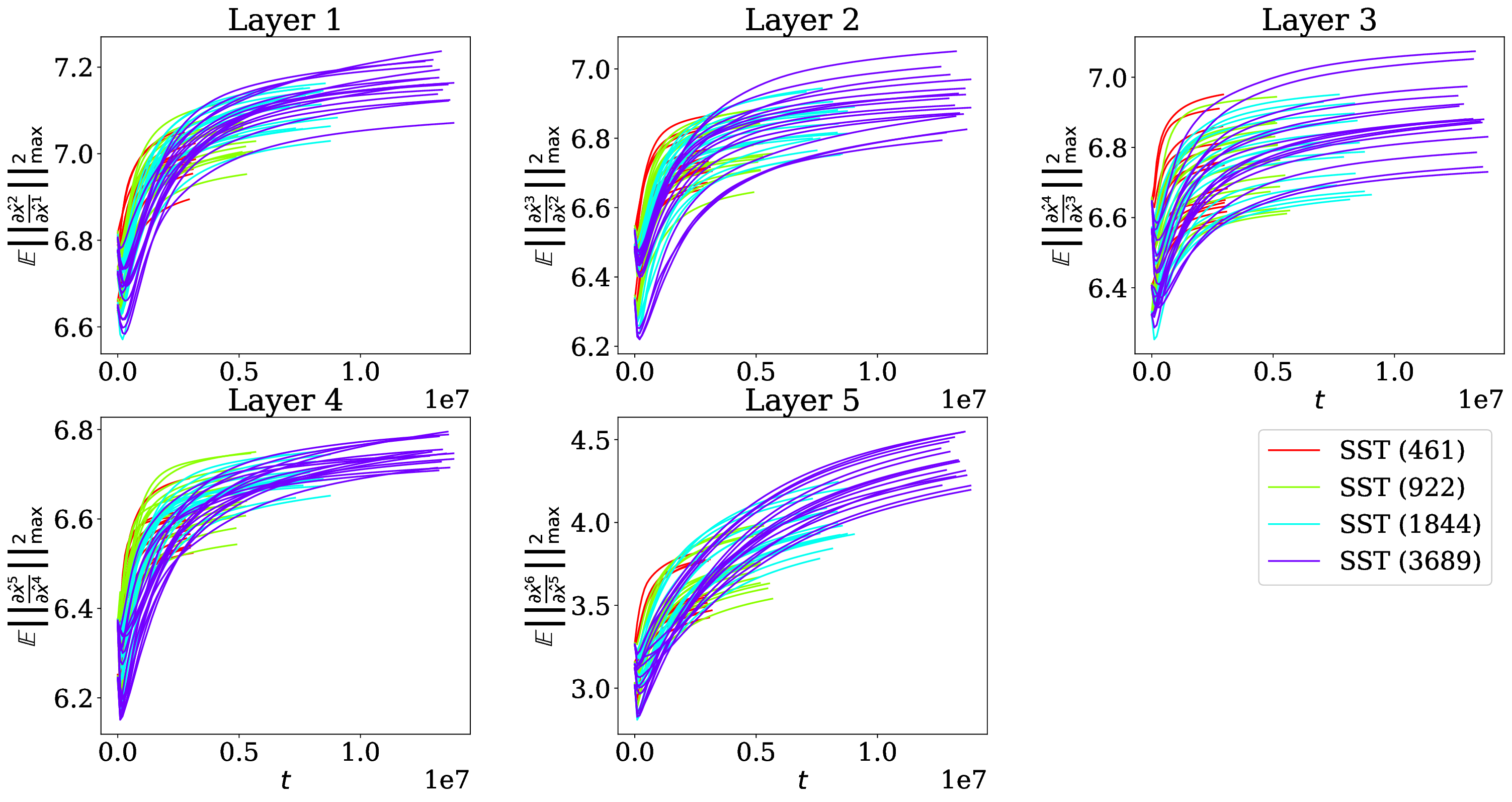}
		\caption{\label{fig:mlp-sst-cross-entropy-layerwise-operator-norm}Expected squared operator norms of the layerwise Jacobians $\mathbb{E} ||\frac{\partial \hat{x}^{i+1}}{\partial\hat{x}^i}||^2_{\max}$ when trained on SST using cross-entropy for layers $i = 1$ (top left), 2 (top middle), 3 (top right), 4 (bottom left), and 5 (bottom middle)}
	\end{center}
\end{figure}

\newpage

\subsubsection{Alignment Ratios $r(\Delta^i, \frac{\partial \hat{x}^{i+1}}{\partial \hat{x}^i})$}

\begin{figure}
	\begin{center}
		\includegraphics[width=\textwidth]{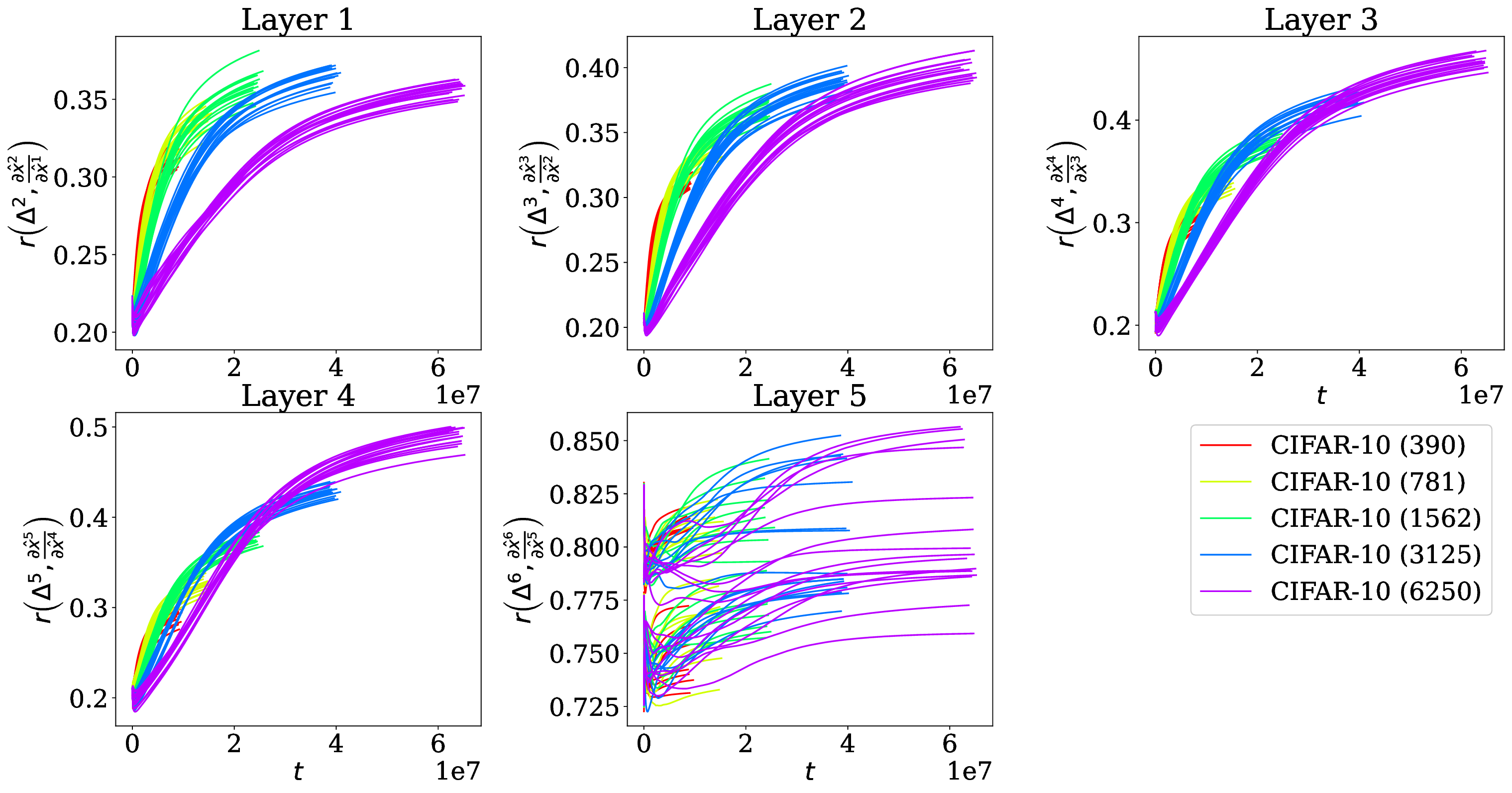}
		\caption{\label{fig:mlp-cifar10-cross-entropy-delta-jacobian-alignment-ratios}Alignment ratios $r(\Delta^i, \frac{\partial \hat{x}^{i+1}}{\partial \hat{x}^i})$ when trained on CIFAR-10 using cross-entropy for layers $i = 1$ (top left), 2 (top middle), 3 (top right), 4 (bottom left), and 5 (bottom middle)}
	\end{center}
\end{figure}

\begin{figure}
	\begin{center}
		\includegraphics[width=\textwidth]{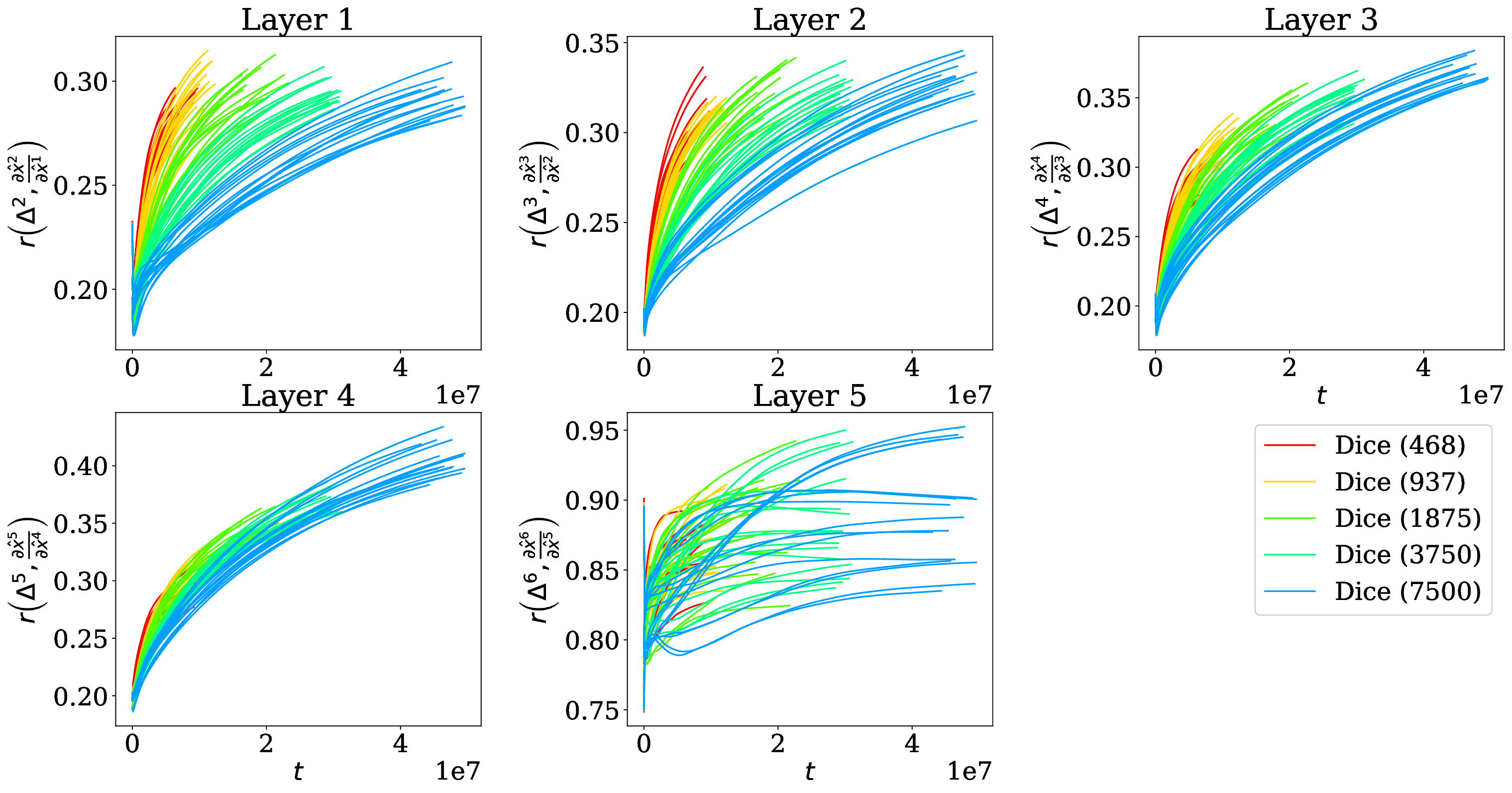}
		\caption{\label{fig:mlp-dice-cross-entropy-delta-jacobian-alignment-ratios}Alignment ratios $r(\Delta^i, \frac{\partial \hat{x}^{i+1}}{\partial \hat{x}^i})$ when trained on dice dataset using cross-entropy for layers $i = 1$ (top left), 2 (top middle), 3 (top right), 4 (bottom left), and 5 (bottom middle)}
	\end{center}
\end{figure}

\begin{figure}
	\begin{center}
		\includegraphics[width=\textwidth]{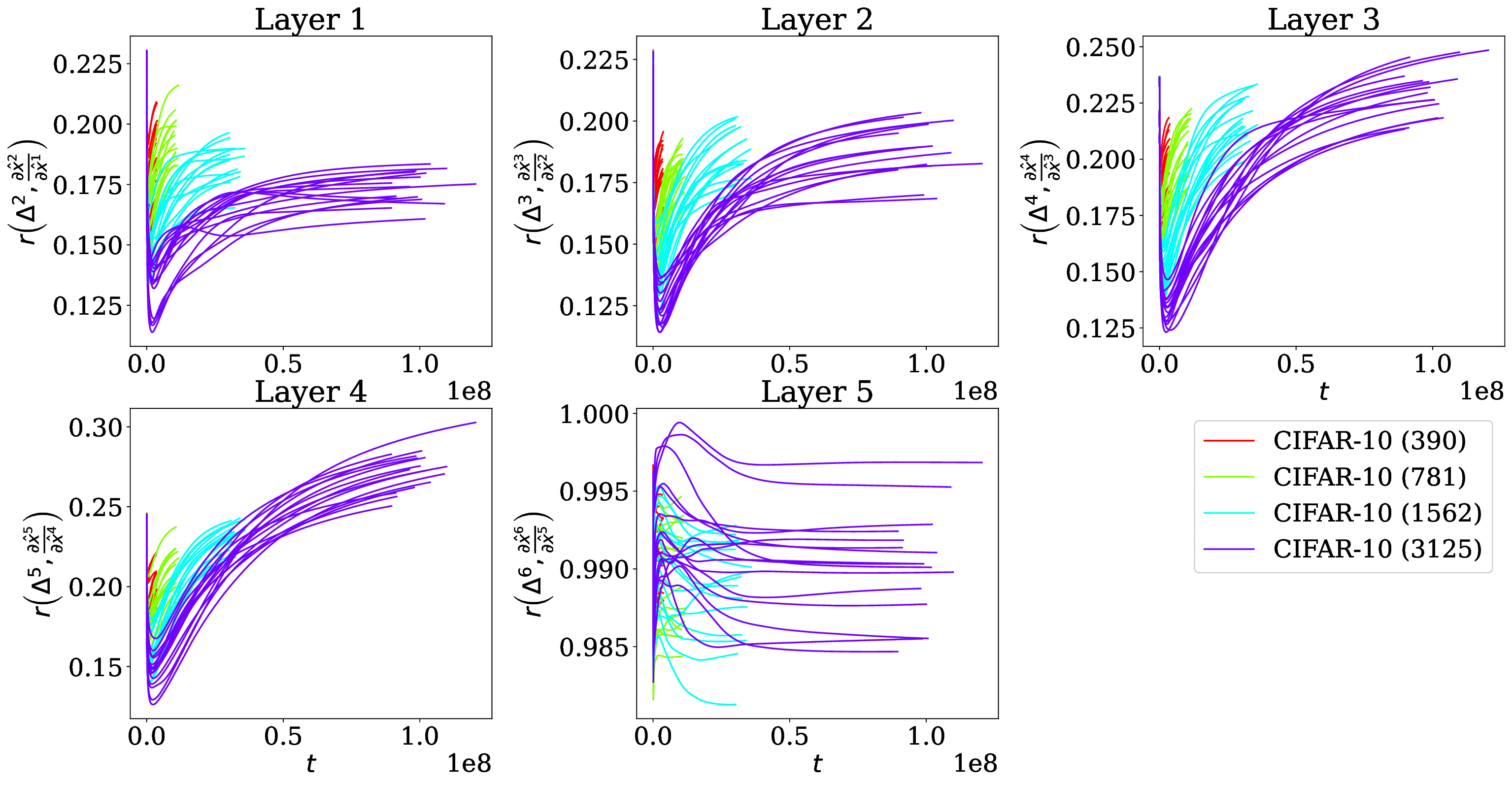}
		\caption{\label{fig:mlp-cifar10-mse-delta-jacobian-alignment-ratios}Alignment ratios $r(\Delta^i, \frac{\partial \hat{x}^{i+1}}{\partial \hat{x}^i})$ when trained on CIFAR-10 using MSE for layers $i = 1$ (top left), 2 (top middle), 3 (top right), 4 (bottom left), and 5 (bottom middle)}
	\end{center}
\end{figure}

\begin{figure}
	\begin{center}
		\includegraphics[width=\textwidth]{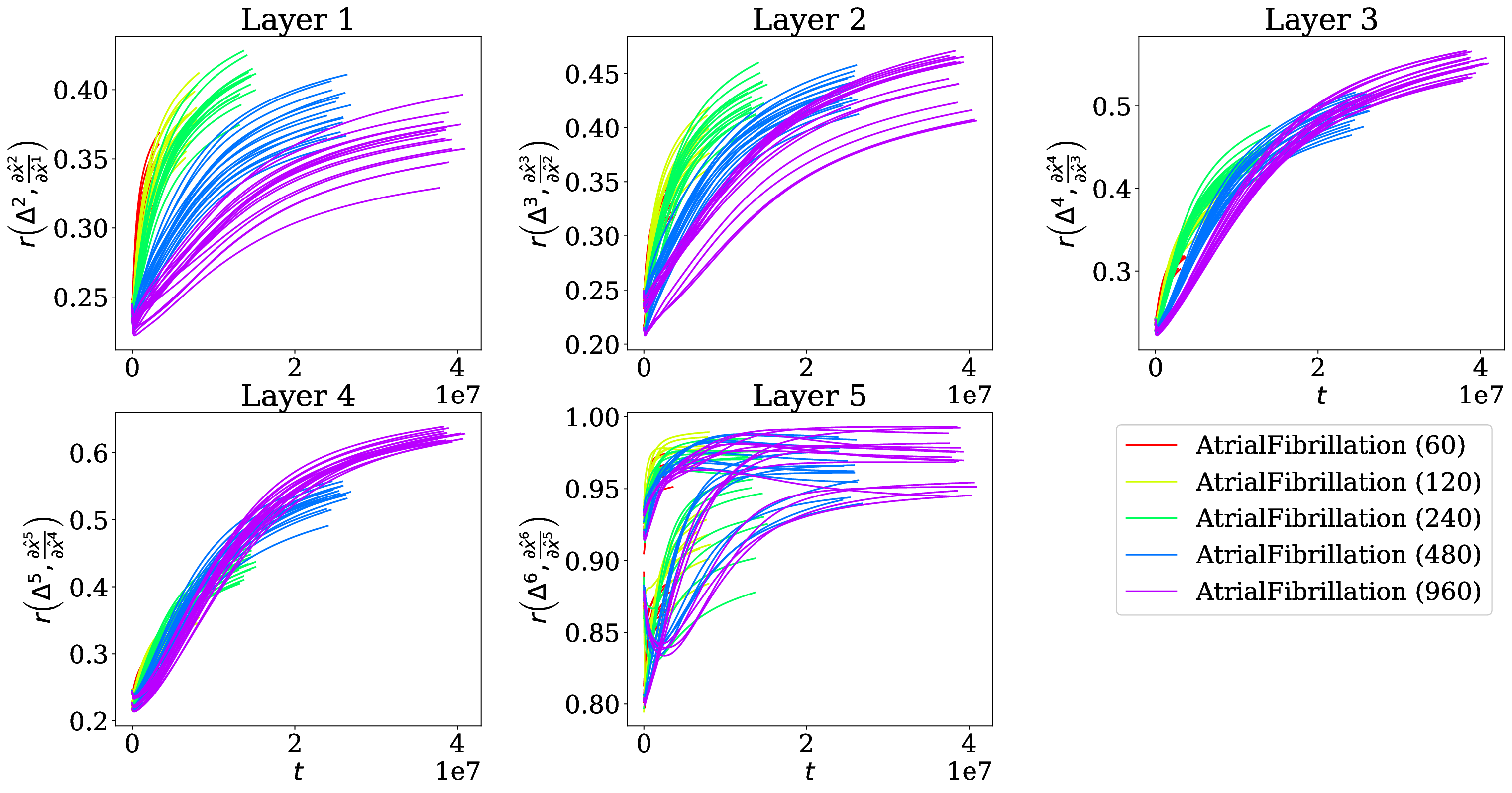}
		\caption{\label{fig:mlp-ucr-cross-entropy-delta-jacobian-alignment-ratios}Alignment ratios $r(\Delta^i, \frac{\partial \hat{x}^{i+1}}{\partial \hat{x}^i})$ when trained on AtrialFibrillation using cross-entropy for layers $i = 1$ (top left), 2 (top middle), 3 (top right), 4 (bottom left), and 5 (bottom middle)}
	\end{center}
\end{figure}

\begin{figure}
	\begin{center}
		\includegraphics[width=\textwidth]{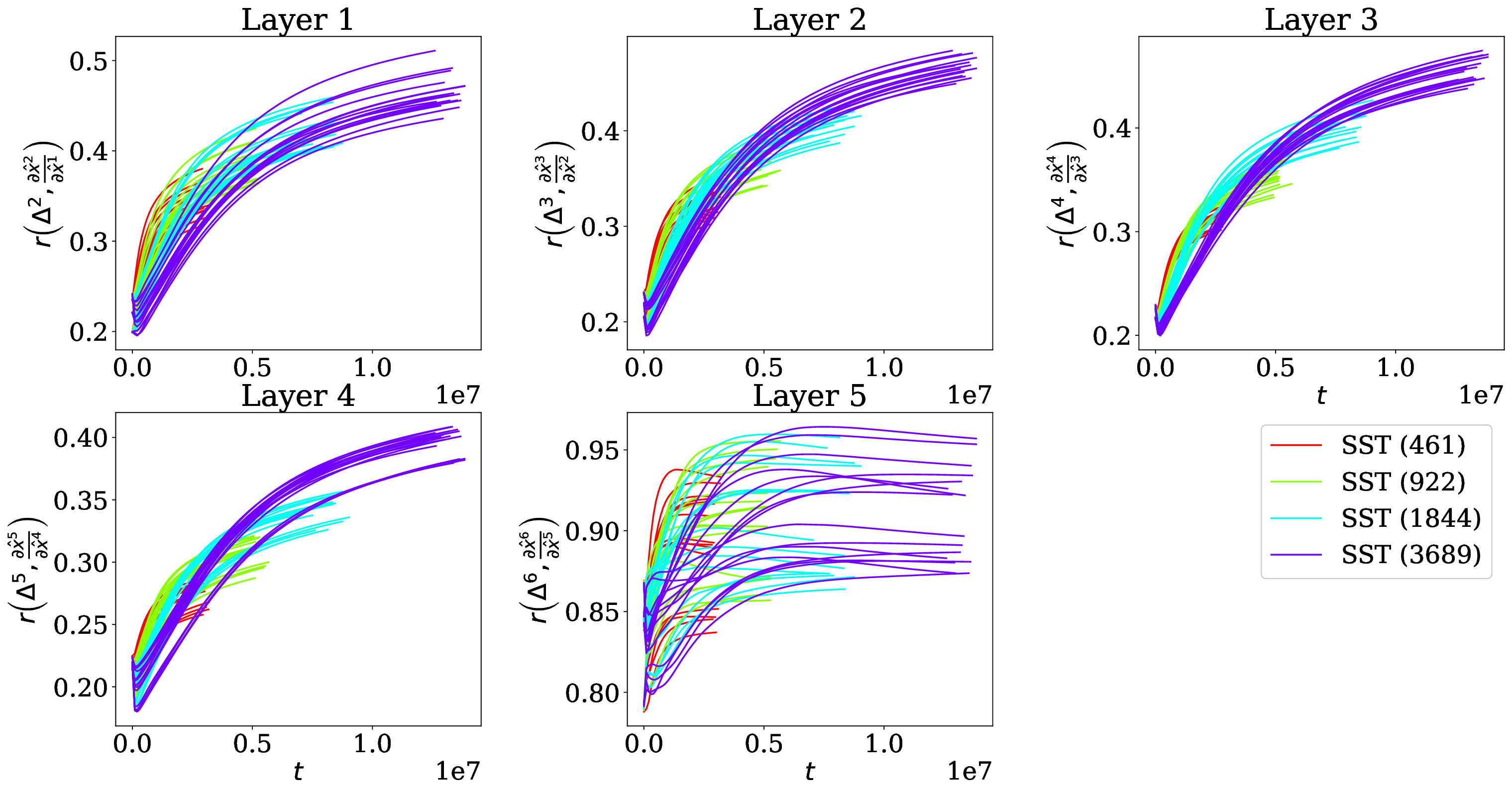}
		\caption{\label{fig:mlp-sst-cross-entropy-delta-jacobian-alignment-ratios}Alignment ratios $r(\Delta^i, \frac{\partial \hat{x}^{i+1}}{\partial \hat{x}^i})$ when trained on SST using cross-entropy for layers $i = 1$ (top left), 2 (top middle), 3 (top right), 4 (bottom left), and 5 (bottom middle)}
	\end{center}
\end{figure}

\end{document}